\documentclass{article}
\PassOptionsToPackage{numbers, compress}{natbib}

\usepackage[nonatbib,final]{neurips_2020}

\usepackage[utf8]{inputenc} 
\usepackage[T1]{fontenc}    
\usepackage{booktabs}       
\usepackage{amsfonts}       
\usepackage{mathtools}
\usepackage{nicefrac}       
\usepackage{microtype}      
\usepackage{graphicx}
\usepackage{amsmath}
\usepackage{pgfplots}
\usepackage{wrapfig}
\usepackage{tikz}
\usepackage{caption}
\usepackage{enumitem}
\usepackage{authblk}
\usepackage[ruled,vlined,noend]{algorithm2e}
\usepackage[labelformat=simple]{subcaption}
\usepackage{adjustbox}
\usepackage{url} 
\usepackage{bbm}
\def\1{\mathbbm{1}}
\usepackage{multirow}
\usepackage{amssymb}
\usepackage{pifont}
\usepackage{booktabs}
\usepackage{color, colortbl}
\usepackage{hhline}
\definecolor{Gray}{gray}{0.9}
\newcolumntype{g}{>{\columncolor{Gray}}c}
\newcommand{\STAB}[1]{\begin{tabular}{@{}c@{}}#1\end{tabular}}
\newcommand{\ie}{\textit{i.e.}}
\newcommand{\eg}{\textit{e.g.}}
\newcommand{\etal}{\textit{et al.}}

\newcommand{\etc}{\textit{etc.}}

\newcommand{\cf}{\textit{cf. }}

\usetikzlibrary{er,positioning,arrows,chains,shapes,patterns}
\newcommand{\norm}[1]{\left\lVert#1\right\rVert}

\DeclareMathOperator{\E}{\mathbb{E}}
\newcommand\independent{\protect\mathpalette{\protect\independenT}{\perp}}
\def\independenT#1#2{\mathrel{\rlap{$#1#2$}\mkern2mu{#1#2}}}

\usepackage{xcolor}
\usepackage[pagebackref=true,breaklinks=true,colorlinks=false,bookmarks=false,linkbordercolor=red]{hyperref}

\hypersetup{%
    pdfborder = {0 0 1}
}

\tikzstyle{line} = [draw, -latex']
\tikzstyle{block} = [rectangle, draw, text width=4.5em, text centered, minimum height=1em]
\title{Interventional Few-Shot Learning}

\author{%
 \textbf{Zhongqi Yue}\textsuperscript{1,3}, \quad \textbf{Hanwang Zhang}\textsuperscript{1}, \quad \textbf{Qianru Sun}\textsuperscript{2}, \quad \textbf{Xian-Sheng Hua}\textsuperscript{3}\\
{\small \textsuperscript{1}Nanyang Technological University,\quad \textsuperscript{2}Singapore Management University,\\ \textsuperscript{3}Damo Academy, Alibaba Group}\\
{\tt\small yuez0003@ntu.edu.sg,\quad hanwangzhang@ntu.edu.sg,\\
\tt\small \quad qianrusun@smu.edu.sg, \quad xiansheng.hxs@alibaba-inc.com}\\}

\begin{document}

\maketitle

\begin{abstract}
We uncover an ever-overlooked deficiency in the prevailing Few-Shot Learning (FSL) methods: the pre-trained knowledge is indeed a confounder that limits the performance. This finding is rooted from our causal assumption: a Structural Causal Model (SCM) for the causalities among the pre-trained knowledge, sample features, and labels. Thanks to it, we propose a novel FSL paradigm: Interventional Few-Shot Learning (IFSL). Specifically, we develop three effective IFSL algorithmic implementations based on the backdoor adjustment, which is essentially a causal intervention towards the SCM of many-shot learning: the upper-bound of FSL in a causal view. It is worth noting that the contribution of IFSL is orthogonal to existing fine-tuning and meta-learning based FSL methods, hence IFSL can improve all of them, achieving a new 1-/5-shot state-of-the-art on \textit{mini}ImageNet, \textit{tiered}ImageNet, and cross-domain CUB. Code is released at \url{https://github.com/yue-zhongqi/ifsl}.
\end{abstract}
\section{Introduction}
\label{sec:1}
Few-Shot Learning (FSL) --- the task of training a model using very few samples --- is nothing short of a panacea for any scenario that requires fast model adaptation to new tasks~\cite{wang2019few}, such as minimizing the need for expensive trials in reinforcement learning~\cite{jamal2019task} and saving computation resource for light-weight neural networks~\cite{howard2017mobilenets,hinton2015distilling}. Although we knew that, more than a decade ago, the crux of FSL is to imitate the human ability of transferring prior knowledge to new tasks~\cite{fei2006one}, not until the recent advances in pre-training techniques, had we yet reached a consensus on ``what \& how to transfer'': a powerful neural network $\Omega$ pre-trained on a large dataset $\mathcal{D}$. In fact, the prior knowledge learned from pre-training prospers today's deep learning era, \eg, $\mathcal{D}$ = ImageNet, $\Omega$ = ResNet in visual recognition~\cite{he2016deep, he2017mask}; $\mathcal{D}$ = Wikipedia, $\Omega$ = BERT in natural language processing~\cite{vaswani2017attention, devlin2018bert}. 

\begin{wrapfigure}{r}{0.5\textwidth}
    \centering
    \includegraphics[width=.9\linewidth]{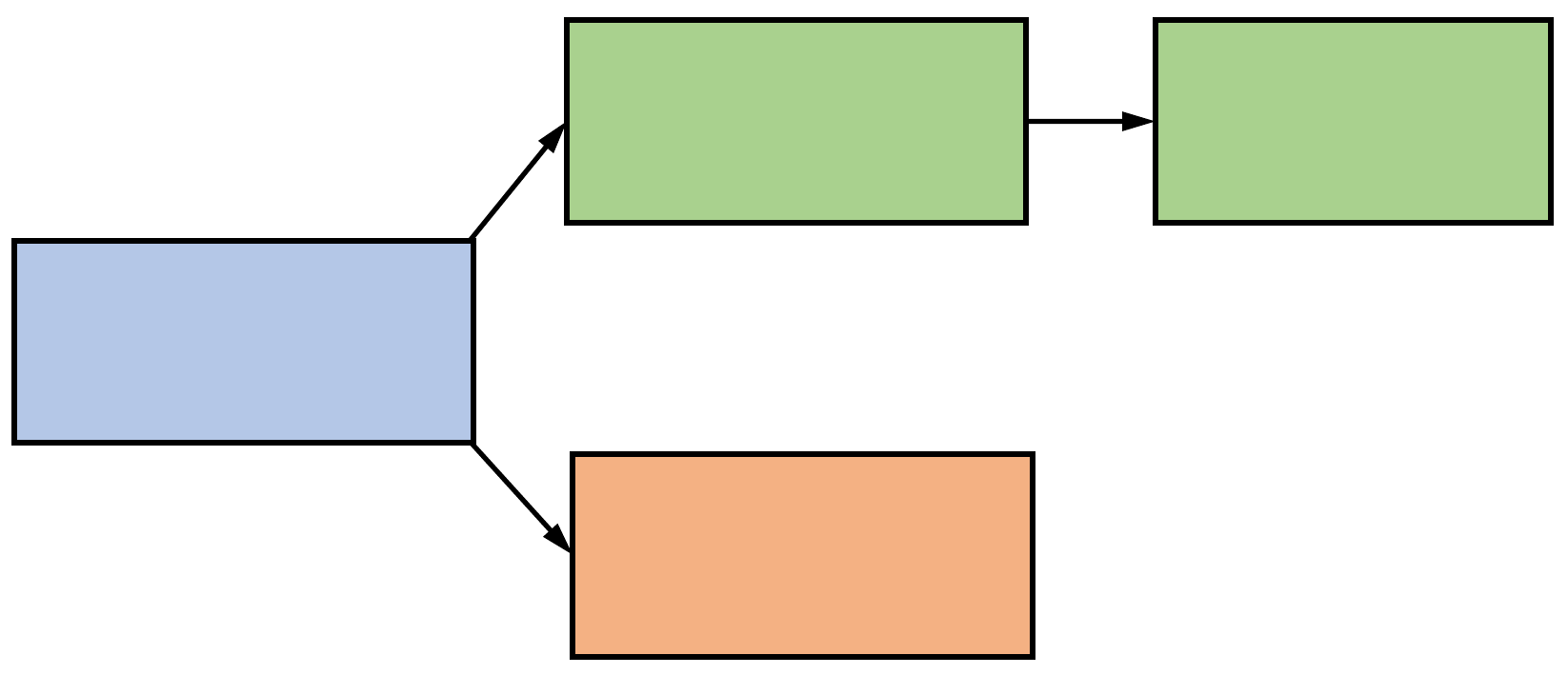}
    
    \begin{tikzpicture}[overlay, remember picture]
        \node [below left,text width=3cm,align=center] at (-0.55,2.07) {\scalebox{.7}{Pre-Training}};
        \node [below left,text width=3cm,align=center] at (-0.55,1.77) {\scalebox{.7}{$\mathcal{D}$}};
        
        \node [below left,text width=3cm,align=center] at (1.67,3) {\scalebox{.7}{Meta-Learning}};
        \node [below left,text width=3cm,align=center] at (1.67,2.7) {\scalebox{.6}{$\{$Fine-tune$(\mathcal{S}_i, \mathcal{Q}_i)\}$}};
        
        \node [below left,text width=3cm,align=center] at (1.7,1.25) {\scalebox{.7}{Fine-Tuning}};
        \node [below left,text width=3cm,align=center] at (1.7,0.95) {\scalebox{.7}{$(\mathcal{S}, \mathcal{Q})$}};
        
        \node [below left,text width=3cm,align=center] at (3.9,3) {\scalebox{.7}{Fine-Tuning}};
        \node [below left,text width=3cm,align=center] at (3.9,2.7) {\scalebox{.7}{$(\mathcal{S}, \mathcal{Q})$}};
        
        \node [below left,text width=3cm,align=center] at (0.88,1.9) {\scalebox{.7}{or}};
        \node [below left,text width=3cm,align=center] at (0.38,2.7) {\scalebox{.7}{$\Omega$}};
        \node [below left,text width=3cm,align=center] at (0.38,1.16) {\scalebox{.7}{$\Omega$}};
        \node [below left,text width=3cm,align=center] at (2.9,3) {\scalebox{.7}{$\Omega_{\phi}$}};
    \end{tikzpicture}
    
    \caption{The relationships among different FSL paradigms (color green and orange). Our goal is to remove the deficiency introduced by Pre-Training.}
    \label{fig:1}
\end{wrapfigure}

In the context of pre-trained knowledge, we denote the original FSL training set as \emph{support} set $\mathcal{S}$ and the test set as \emph{query} set $\mathcal{Q}$, where the classes in $(\mathcal{S},\mathcal{Q})$ are unseen (or new) in $\mathcal{D}$. Then, we can use $\Omega$ as a backbone  (fixed or partially trainable) for extracting sample representations $\mathbf{x}$, and thus FSL can be achieved simply by \emph{fine-tuning} the target model on $\mathcal{S}$ and test it on $\mathcal{Q}$~\cite{chen2019closer,dhillon2019baseline}. However, the fine-tuning only exploits the $\mathcal{D}$'s knowledge on ``what to transfer'', but neglects ``how to transfer''. Fortunately, the latter can be addressed by applying a post-pre-training and pre-fine-tuning strategy: \emph{meta-learning}~\cite{santoro2016meta}. Different from fine-tuning whose goal is the ``model'' trained on $\mathcal{S}$ and tested on $\mathcal{Q}$, meta-learning aims to learn the ``meta-model'' --- a learning behavior --- trained on many learning episodes $\{(\mathcal{S}_i, \mathcal{Q}_i)\}$ sampled from $\mathcal{D}$ and tested on the target task $(\mathcal{S}, \mathcal{Q})$. In particular, the behavior can be parametrized by $\phi$ using model parameter generator~\cite{qiao2018few,gidaris2019generating} or initialization~\cite{finn2017model}. After meta-learning, we denote $\Omega_\phi$ as the new model starting point for the subsequent fine-tuning on target task $(\mathcal{S}, \mathcal{Q})$. Figure~\ref{fig:1} illustrates the relationships among the above discussed FSL paradigms.

\begin{figure}
\centering
\captionsetup{font=footnotesize,labelfont=footnotesize}
\begin{subfigure}{.325\textwidth}
  \centering
  \includegraphics[width=\linewidth]{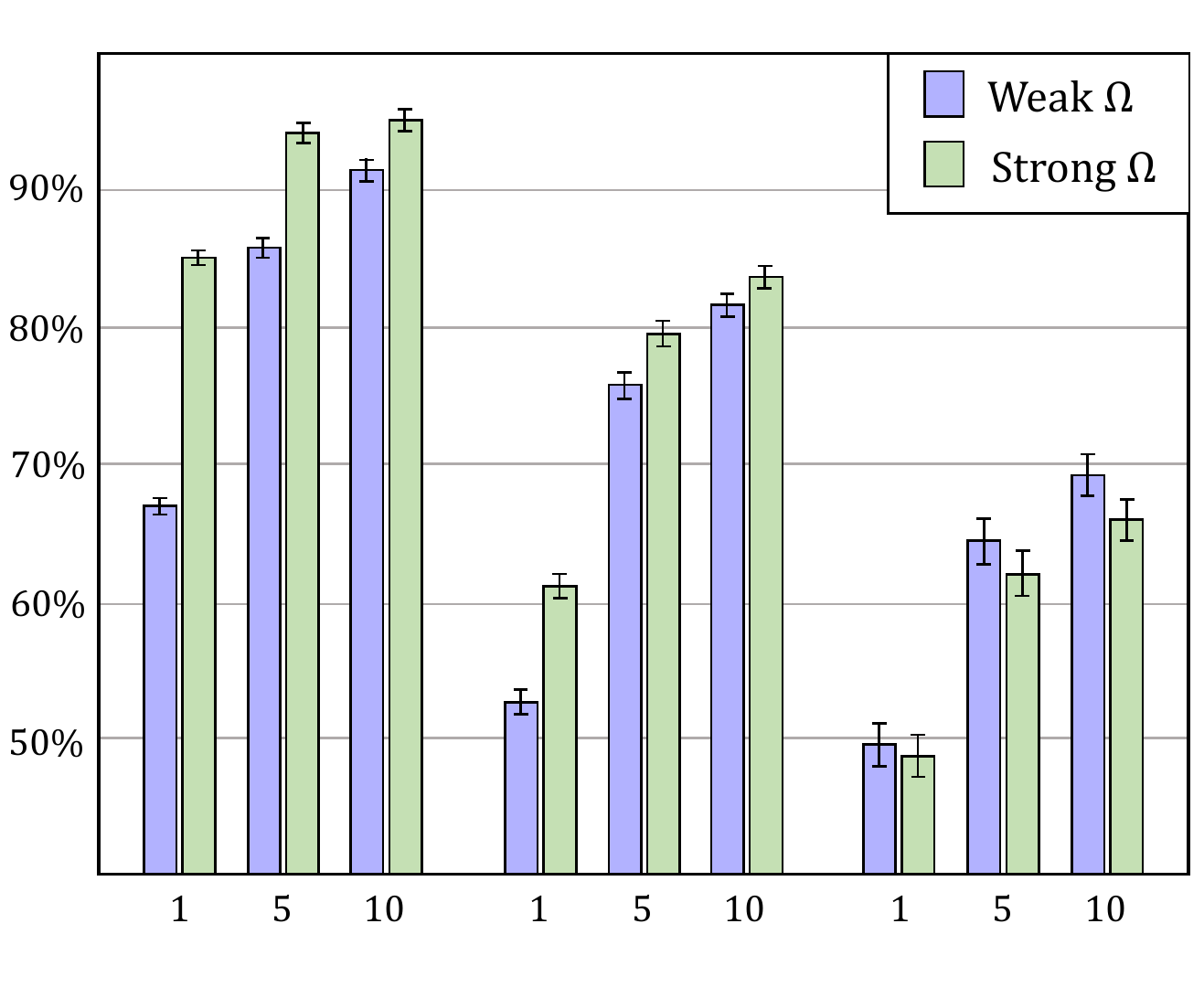}
  \caption{}
  \label{fig:2a}
\end{subfigure}%
\begin{subfigure}{.675\textwidth}
  \centering
  \includegraphics[width=0.95\linewidth]{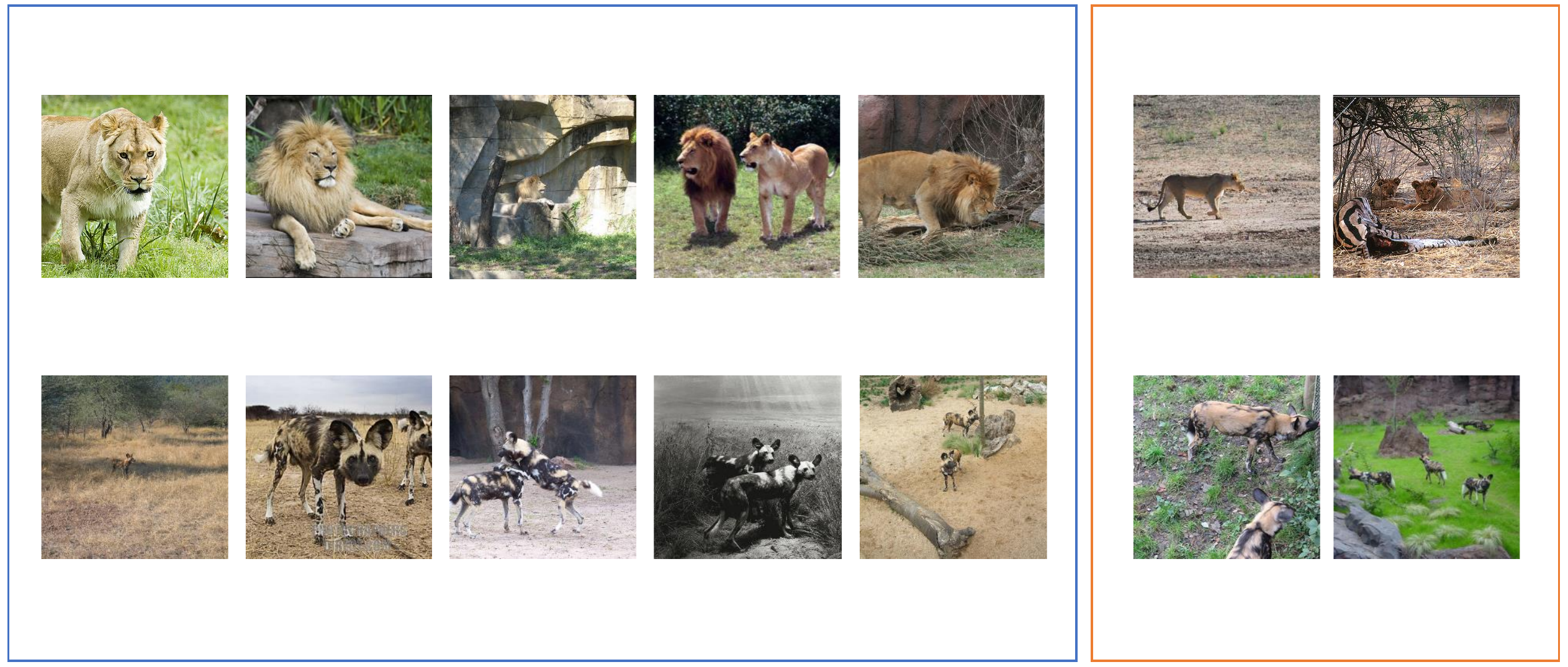}
  
  \caption{}
  \label{fig:2b}
\end{subfigure}
\vspace{-2.5mm}
    \begin{tikzpicture}[overlay, remember picture]
        \node [below left,text width=3cm,align=center] at (-4.8, 4.5) {\scalebox{.55}{Accuracy}};
        \node [below left,text width=3cm,align=center] at (-4.3, 0.9) {\scalebox{.6}{$\mathcal{S} \sim \mathcal{Q}$}};
        \node [below left,text width=3cm,align=center] at (-2.9, 0.9) {\scalebox{.6}{Average}};
        \node [below left,text width=3cm,align=center] at (-1.6, 0.9) {\scalebox{.6}{$\mathcal{S} \not\sim \mathcal{Q}$}};
        
        \node [below left,text width=3cm,align=center] at (2.55, 4.3) {\scalebox{.7}{\textbf{Support Set}}};
        
        \node [below left,text width=3cm,align=center] at (2.56, 2.7) {\scalebox{.7}{\textit{``Lion''}}};
        \node [below left,text width=3cm,align=center] at (2.56, 1.1) {\scalebox{.7}{\textit{``African Hunting Dog''}}};
        
        \node [below left,text width=3cm,align=center] at (7, 4.3) {\scalebox{.7}{\textbf{Query Set}}};
        \node [below left,text width=3cm,align=center] at (7.05, 2.83) {\scalebox{.6}{Classified as \textit{``Dog''}}};
        \node [below left,text width=3cm,align=center] at (7.05, 2.6) {\scalebox{.6}{(due to ``yellow grass'')}};
        
        \node [below left,text width=3cm,align=center] at (7.05, 1.205) {\scalebox{.6}{Classified as \textit{``Lion''}}};
        \node [below left,text width=3cm,align=center] at (7.05, 1) {\scalebox{.6}{(due to ``green grass'')}};
   \end{tikzpicture}
\caption{Quantitative and qualitative evidences of pre-trained knowledge misleading the fine-tune FSL paradigm. (a) \textit{mini}ImageNet fine-tuning accuracy on 1-/5-/10-shot FSL using weak and strong backbones: ResNet-10 and WRN-28-10. $\mathcal{S}\sim \mathcal{Q}$ (or $\mathcal{S}\not\sim \mathcal{Q}$) denotes the pre-trained classifier scores of the query is similar (or dissimilar) to that of the support set. ``Average'' is the mean of both. The dissimilarity is measured using query hardness defined in Section~\ref{sec:5.1}. (b) An example of 5-shot $\mathcal{S}\not\sim \mathcal{Q}$.}
\label{fig:2}
\vspace{-4mm}
\end{figure}

It is arguably a common sense that the stronger the pre-trained $\Omega$ is, the better the downstream model will be. However, we surprisingly find that this may not be always the case in FSL. As shown in Figure~\ref{fig:2a}, we can see a paradox: though stronger $\Omega$ improves the performance on average, it indeed degrades that of samples in $\mathcal{Q}$ dissimilar to $\mathcal{S}$.
To illustrate the ``dissimilar'', we show a 5-shot learning example in Figure~\ref{fig:2b}, where the prior knowledge on ``green grass'' and ``yellow grass'' is misleading. For example, the ``Lion'' samples in $\mathcal{Q}$ have ``yellow grass'', hence they are misclassified as ``Dog'' whose $\mathcal{S}$ has major ``yellow grass''. If we use stronger $\Omega$,  the seen old knowledge (``grass'' \& ``color'') will be more robust than the unseen new knowledge (``Lion'' \& ``Dog''), and thus the old becomes even more misleading. We believe that such a paradox reveals an unknown systematic deficiency in FSL, which has been however hidden for years by our gold-standard ``fair'' accuracy, averaged over all the random $(\mathcal{S}, \mathcal{Q})$ test trials, regardless of the similarity between $\mathcal{S}$ and $\mathcal{Q}$  (\cf Figure~\ref{fig:2a}). Though Figure~\ref{fig:2} only illustrates the fine-tune FSL paradigm, the deficiency is expected in the meta-learning paradigm, as fine-tune is also used in each meta-train episode (Figure~\ref{fig:1}). We will analyze them thoroughly in Section~\ref{sec:5}.

In this paper, we first point out that the cause of the deficiency:  pre-training can do evil in FSL, and then propose a novel FSL paradigm: Interventional Few-Shot Learning (IFSL), to counter the evil. Our theory is based on the assumption of the \emph{causalities} among the pre-trained knowledge, few-shot samples, and class labels. Specifically, our contributions are summarized as follows.

\begin{itemize}[leftmargin=+.2in]
    \item We begin with a Structural Causal Model (SCM) assumption in Section~\ref{sec:2.2}, which shows that the pre-trained knowledge is essentially a \emph{confounder} that causes spurious correlations between the sample features and class labels in support set. As an intuitive example in Figure~\ref{fig:2b}, even though the ``grass'' feature is not the cause of the ``Lion'' label, the prior knowledge on ``grass'' still confounds the classifier to learn a correlation between them.
    \item In Section~\ref{sec:2.3}, we illustrate a causal justification of why the proposed IFSL fundamentally works better: it is essentially a causal approximation to many-shot learning. This motivates us to develop three effective implementations of IFSL using the backdoor adjustment~\cite{pearl2016causal} in Section~\ref{sec:3}.
    \item Thanks to the causal intervention, IFSL is naturally orthogonal to the downstream fine-tuning and meta-learning based FSL methods~\cite{finn2017model,vinyals2016matching,hu1empirical}. In Section~\ref{sec:5.2}, IFSL improves all baselines by a considerable margin, achieving new 1-/5-shot state-of-the-arts: 73.51\%/83.21\% on \textit{mini}ImageNet~\cite{vinyals2016matching}, 83.07\%/88.69\% on \textit{tiered}ImageNet~\cite{ren2018meta}, and 50.71\%/64.43\% on cross-domain CUB~\cite{WelinderEtal2010}.
    \item We further diagnose the detailed performances of FSL methods across different similarities between $\mathcal{S}$ and $\mathcal{Q}$. We find that IFSL outperforms all baselines in every inch.
\end{itemize}
\vspace{-2mm}

\section{Problem Formulations}
\label{sec:2}
\subsection{Few-Shot Learning}
\label{sec:2.1}
We are  interested in a prototypical FSL: train a $K$-way classifier on an $N$-shot support set $\mathcal{S}$, where $N$ is a small number of training samples per class (\eg, $N$=1 or 5); then test the classifier on a query set $\mathcal{Q}$.  As illustrated in Figure~\ref{fig:1}, we have the following two paradigms to train the classifier $P(y|\mathbf{x};\theta)$, predicting the class $y\in\{1, ..., K\}$ of a sample $\mathbf{x}$:


\noindent\textbf{Fine-Tuning}.
We consider the prior knowledge as the sample feature representation $\mathbf{x}$, encoded by the pre-trained network $\Omega$ on dataset $\mathcal{D}$. In particular, we refer $\mathbf{x}$ to the output of the frozen sub-part of $\Omega$ and the rest trainable sub-part of $\Omega$ (if any) can be absorbed into $\theta$. We train the classifier $P(y|\mathbf{x};\theta)$ on the support set $\mathcal{S}$, and then evaluate it on the query set $\mathcal{Q}$ in a standard supervised way.

\noindent\textbf{Meta-Learning}. 
Yet, $\Omega$ only carries prior knowledge in a way of ``representation''. If the dataset $\mathcal{D}$ can be re-organized as the training episodes $\{(\mathcal{S}_i, \mathcal{Q}_i)\}$, each of which can be treated as a ``sandbox'' that has the same $N$-shot-$K$-way setting as the target $(\mathcal{S}, \mathcal{Q})$. Then, we can model the ``learning behavior'' from $\mathcal{D}$ parameterized as $\phi$, which can be learned by the above fine-tuning paradigm for each $(\mathcal{S}_i, \mathcal{Q}_i)$. Formally, we denote $P_\phi(y|\mathbf{x};\theta)$ as the enhanced classifier equipped with the learned behavior. For example, $\phi$ can be the classifier weight generator~\cite{gidaris2019generating}, distance kernel function in $k$-NN~\cite{vinyals2016matching}, or even $\theta$'s initialization~\cite{finn2017model}. Considering $L_\phi(\mathcal{S}_i, \mathcal{Q}_i;\theta)$ as the loss function of $P_\phi(y|\mathbf{x};\theta)$ trained on $\mathcal{S}_i$ and tested on $\mathcal{Q}_i$, we can have $\phi\leftarrow\arg\min_{(\phi,\theta)}\mathbb{E}_i \left[ L_\phi(\mathcal{S}_i, \mathcal{Q}_i;\theta)\right]$, and then we fix the optimized $\phi$ and fine-tune for $\theta$ on $\mathcal{S}$ and test on $\mathcal{Q}$. Please refer to Appendix 5 for the details of various fine-tuning and meta-learning settings.

\subsection{Structural Causal Model}
\label{sec:2.2}

\begin{wrapfigure}[18]{r}{0.6\textwidth}
\centering
\captionsetup{font=footnotesize,labelfont=footnotesize}
\begin{subfigure}[t]{.2\textwidth}
  \centering
  \captionsetup{width=\linewidth}
  \includegraphics[width=0.95\linewidth]{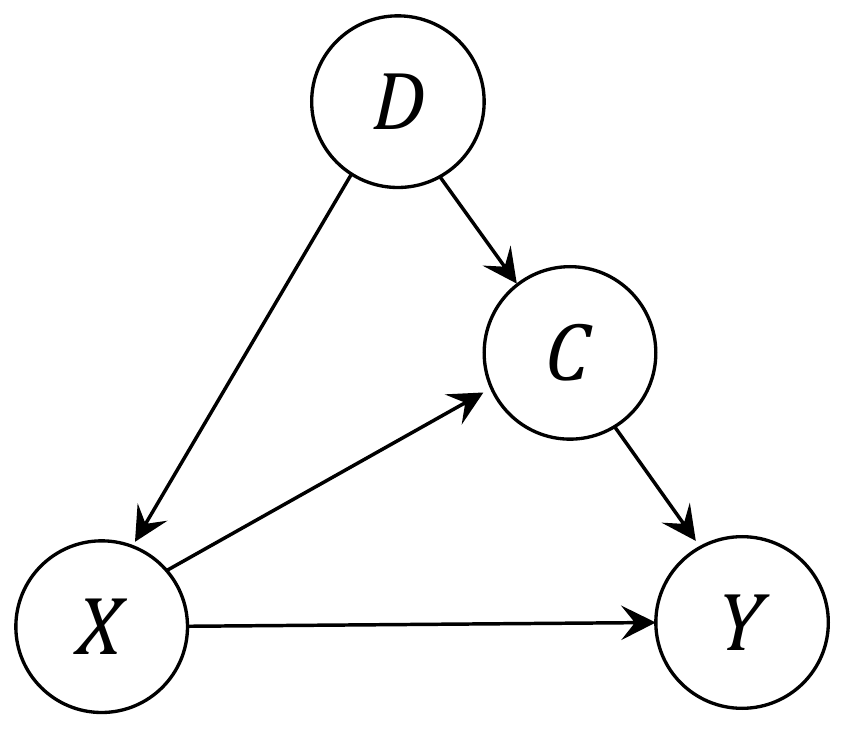}
  \caption{}
  \label{fig:3a}
\end{subfigure}%
\begin{subfigure}[t]{.2\textwidth}
  \centering
  \captionsetup{width=\linewidth}
  \includegraphics[width=0.95\linewidth]{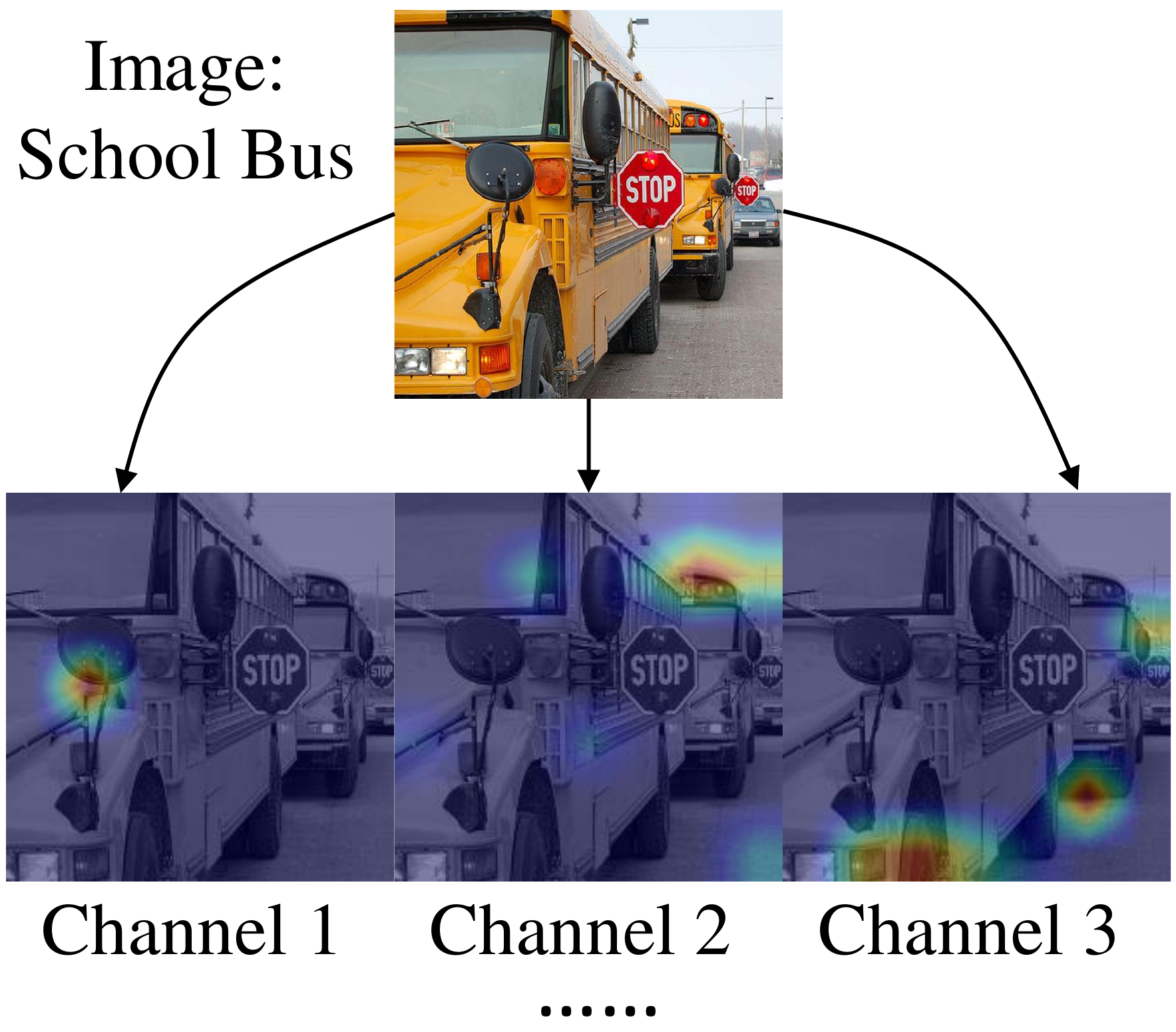}
  \caption{}
  \label{fig:3b}
\end{subfigure}%
\begin{subfigure}[t]{.2\textwidth}
  \centering
  \captionsetup{width=\linewidth}
  \includegraphics[width=0.95\linewidth]{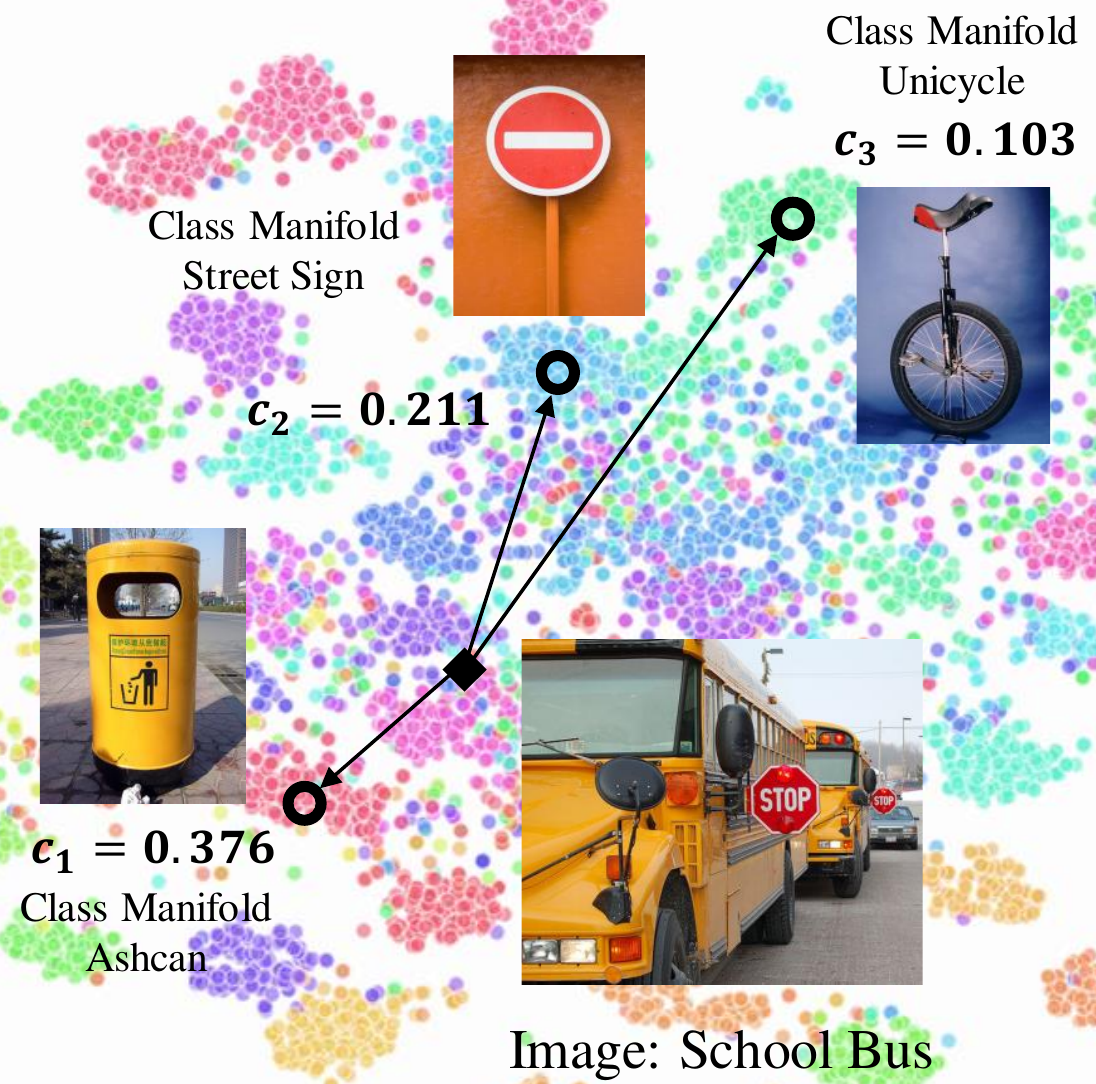}
  \caption{}
  \label{fig:3c}
\end{subfigure}
\vspace{-1mm}
\caption{(a) Causal Graph for FSL; (b) Feature-wise illustration of $D\rightarrow C$: Feature channels of pre-trained network(\eg $1\ldots 512$ for ResNet-10). $X\rightarrow C$: Per-channel response to an image (``school bus'') visualized by CAM\cite{zhou2016learning}; (c) Class-wise illustration for $D\rightarrow C$: features are clustered according to the pre-training semantic classes (colored t-SNE plot\cite{maaten2008visualizing}). $X\rightarrow C$: An image (``school bus'') can be represented in terms of the similarities among the base classes (``ashcan'', ``unicycle'', ``sign'').}
\label{fig:3}
\end{wrapfigure}

From the above discussion, we can see that $(\phi,\theta)$ in meta-learning and  $\theta$ in fine-tuning are both dependent on the pre-training. Such ``dependency'' can be formalized with a Structural Causal Model (SCM)~\cite{pearl2016causal} proposed in Figure~\ref{fig:3a}, where the nodes denote the abstract data variables and the directed edges denote the (functional) causality, \eg, $X\rightarrow Y$ denotes that $X$ is the cause and $Y$ is the effect. Now we introduce the graph and the rationale behind its construction at a high-level. Please see Section~\ref{sec:3} for the detailed functional implementations.


\noindent\textbf{$\boldsymbol{D \to X}$}. We denote $X$ as the feature representation and $D$ as the pre-trained knowledge, \eg, the dataset $\mathcal{D}$ and its induced model $\Omega$. This link assumes that the feature $X$ is extracted by using $\Omega$.

\noindent\textbf{$\boldsymbol{D \rightarrow C\leftarrow X}$}. We denote $C$ as the transformed representation of $X$ in the low-dimensional manifold, whose base is inherited from $D$. This assumption can be rationalized as follows. 1) $D\rightarrow C$:
a set of data points are usually embedded in a low-dimensional manifold. This finding can be dated back to the long history of dimensionality reduction~\cite{tenenbaum2000global,roweis2000nonlinear}. Nowadays, there are theoretical~\cite{arora2019implicit,besserve2018counterfactuals} and empirical~\cite{zhou2016learning,zeiler2014visualizing} evidences showing that disentangled semantic manifolds emerge during training deep networks. 2) $X\rightarrow C$: features can be represented using (or projected onto) the manifold base linearly~\cite{turk1991face,candes2011robust} or non-linearly~\cite{bengio2013representation}. In particular, as later discussed in Section~\ref{sec:3}, we explicitly implement the base as feature dimensions (Figure~\ref{fig:3b}) and class-specific mean features (Figure~\ref{fig:3c}).

\noindent\textbf{$\boldsymbol{X\rightarrow Y \leftarrow C}$}. We denote $Y$ as the classification effect (\eg, logits), which is determined by  $X$ via two ways: 1) the direct $X\rightarrow Y$ and 2) the mediation $X\rightarrow C\rightarrow Y$. In particular, the first way can be removed if $X$ can be fully represented by $C$ (\eg, feature-wise adjustment in Section~\ref{sec:3}). The second way is inevitable even if the classifier does not take $C$ as an explicit input, because any $X$ can be inherently represented by $C$. To illustrate, suppose that $X$ is a linear combination of two base vectors plus a noise residual: $\mathbf{x} = c_1\mathbf{b}_1+c_2\mathbf{b}_2+\mathbf{e}$, any classifier $f(\mathbf{x})$ = $f(c_1\mathbf{b}_1+c_2\mathbf{b}_2+\mathbf{e})$ will implicitly exploit the $C$ representation in terms of $\mathbf{b}_1$ and $\mathbf{b}_2$. In fact, this assumption also fundamentally validates unsupervised representation learning~\cite{bengio2012deep}. To see this, if $C\not\rightarrow Y$ in Figure~\ref{fig:3a}, uncovering the latent knowledge representation from $P(Y|X)$ would be impossible, because the only path
left that transfers knowledge from $D$ to $Y$: $D\rightarrow X\rightarrow Y$, is cut off by conditioning on $X$: $D\not\rightarrow X\rightarrow Y$.


An ideal FSL model should capture the true causality between $X$ and $Y$ to generalize to unseen samples. For example, as illustrated in Figure~\ref{fig:2b}, we expect that the ``Lion'' prediction is caused by the ``lion'' feature \emph{per se}, but not the background ``grass''. However, from the SCM in Figure~\ref{fig:3a}, the conventional correlation $P(Y|X)$ fails to do so, because the increased likelihood of $Y$ given $X$ is not only due to ``X causes Y'' via $X\rightarrow Y$ and $X\rightarrow C\rightarrow Y$,  but also the spurious correlation via 1) $D\rightarrow X$, \eg, 
the ``grass'' knowledge generates the ``grass'' feature, and 2) $D\rightarrow C\rightarrow Y$, \eg, the ``grass'' knowledge generates the ``grass'' semantic, which provides useful context for ``Lion'' label. Therefore, to pursue the true causality between $X$ and $Y$, we need to use the \emph{causal intervention} $P(Y|do(X))$~\cite{pearl2009causality} instead of the likelihood $P(Y|X)$ for the FSL objective.

\vspace{-1mm}

\subsection{Causal Intervention via Backdoor Adjustment}
\label{sec:2.3}
By now, an astute reader may notice that the causal graph in Figure~\ref{fig:3a} is also valid for Many-Shot Learning (MSL), \ie, conventional learning based on pre-training. Compared to FSL, the $P(Y|X)$ estimation of MSL is much more robust. For example, on \textit{mini}ImageNet, a 5-way-550-shot fine-tuned classifier can achieve 95\% accuracy, while a 5-way-5-shot one only obtains 79\%. We used to blame FSL for insufficient data by the law of large numbers in point estimation~\cite{Dekking2005}. However, it does not answer why MSL converges to the true causal effects as the number of samples increases infinitely. In other words, why $P(Y|do(X)) \approx P(Y|X)$ in MSL while $P(Y|do(X)) \not\approx P(Y|X)$ in FSL?

To answer the question,  we need to incorporate the endogenous feature sampling $\mathbf{x}\sim P(X|I)$ into the estimation of $P(Y|X)$, where $I$ denotes the sample ID. We have $P(Y|X =\mathbf{x}_i) \coloneqq \mathbb{E}_{\mathbf{x}\sim P(X|I)} P(Y|X = \mathbf{x}, I = i) = P(Y|I)$, \ie, we can use $P(Y|I)$ to estimate $P(Y|X)$. In Figure~\ref{fig:4a}, the causal relation between $I$ and $X$ is purely $I\to X$, \ie, $X \to I$ does not exist, because tracing the $X$'s ID  out of many-shot samples is like to find a needle in a haystack, given the nature that a DNN feature is an abstract and diversity-reduced representation of many samples~\cite{goodfellow2016deep}. However, as shown in Figure~\ref{fig:4b}, $X\rightarrow I$ persists in FSL, because it is much easier for a model to ``guess'' the correspondence, \eg, the 1-shot extreme case that has a trivial 1-to-1 mapping for $X\leftrightarrow I$.  Therefore, as we formally show in Appendix 1, the key causal difference between MSL and FSL is: MSL essentially makes $I$ an \emph{instrumental variable}~\cite{angrist2001instrumental} that achieves $P(Y|X) \coloneqq P(Y|I) \approx P(Y|do(X))$. Intuitively, we can see that all the causalities between $I$ and $D$ in MSL are all blocked by colliders\footnote[1]{In causal graph, the junction $A\rightarrow B\leftarrow C$ is called a ``collider'', making $A$ and $C$ independent even though $A$ and $C$ are linked via $B$~\cite{pearl2016causal}. For example, $A$ = ``Quality'', $C$ = ``Luck'', and $B$ = ``Paper Acceptance''.}, making $I$ and $D$ independent. So, the feature $X$ is essentially ``intervened'' by $I$, no longer dictated by $D$, \eg, neither ``yellow grass'' nor ``green grass'' dominates ``Lion'' in Figure~\ref{fig:2b}, mimicking the casual intervention by controlling the use of pre-trained knowledge. 
\vspace{-2mm}
\begin{wrapfigure}{r}{0.6\textwidth}
\centering
\captionsetup{font=footnotesize,labelfont=footnotesize}
\begin{subfigure}[t]{.2\textwidth}
  \centering
  \captionsetup{width=\linewidth}
  \includegraphics[width=\linewidth]{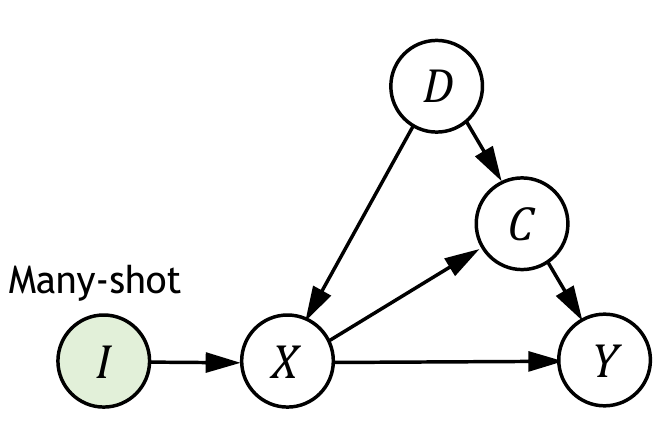}
  \caption{}
  \label{fig:4a}
\end{subfigure}%
\begin{subfigure}[t]{.2\textwidth}
  \centering
  \captionsetup{width=\linewidth}
  \includegraphics[width=\linewidth]{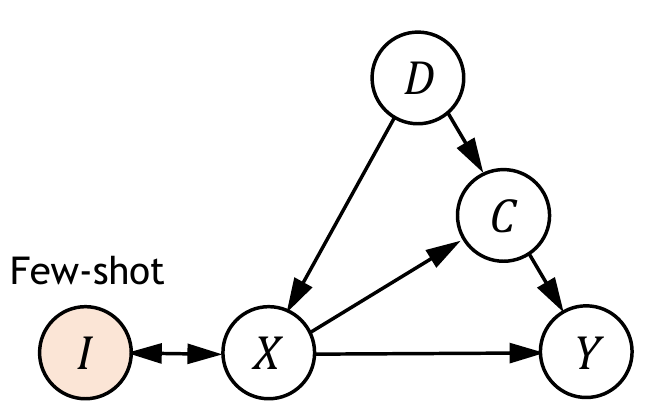}
  \caption{}
  \label{fig:4b}
\end{subfigure}
\begin{subfigure}[t]{.12\textwidth}
  \centering
  \captionsetup{width=\linewidth}
  \includegraphics[width=\linewidth]{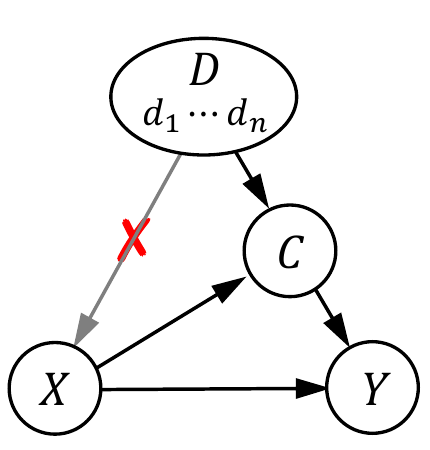}
  \caption{}
  \label{fig:4c}
\end{subfigure}
\caption{Causal graphs with sampling process. (a) Many-Shot Learning, where $P(Y|X) \approx P(Y|do(X))$; (b) Few-Shot Learning where $P(Y|X)\not\approx P(Y|do(X)$; (c) Interventional Few-Shot Learning where we directly model $P(Y|do(X))$.}
\label{fig:4}
\end{wrapfigure}

In this paper, we propose to use the backdoor adjustment~\cite{pearl2016causal} to achieve $P(Y|do(X))$ without the need for many-shot, which certainly undermines the definition of FSL. The backdoor adjustment assumes that we can observe and stratify the confounder, \ie, $D = \{d\}$, where each $d$ is a stratification of the pre-trained knowledge. Formally, as shown in Appendix 2, the backdoor adjustment for the graph in Figure~\ref{fig:3a} is:
\vspace{-1mm}
\begin{equation}
    P\left(Y|do(X=\boldsymbol{x})\right)=\sum_{d} P\left(Y|X=\boldsymbol{x}, D=d, C = g(\mathbf{x}, d)\right)P(D=d),
    \label{eq:backdoor}
\end{equation}

where $g$ is a function defined later. However, it is not trivial to instantiate $d$, especially when $D$ is a 3rd-party delivered pre-trained network where the dataset is unobserved~\cite{Detectron2018}. Next, we will offer three practical implementations of Eq.~\eqref{eq:backdoor} for Interventional FSL.

\section{Interventional Few-Shot Learning}
\label{sec:3}
Our implementation idea is inspired from the two inherent properties of any pre-trained DNN. First, each feature dimension carries a semantic meaning, \eg, every channel in convolutional neural network is well-known to encode visual concepts~\cite{zhou2016learning,zeiler2014visualizing}. So, each feature dimension represents a piece of knowledge. Second, most prevailing pre-trained models use a classification task as the objective, such as the 1,000-way classifier of ResNet~\cite{he2016deep} and the token predictor of BERT~\cite{devlin2018bert}. Therefore, the classifier can be considered as the distilled knowledge, which has been already widely adopted in literature~\cite{hinton2015distilling}. Next, we will detail the proposed Interventional FSL (IFSL)
by providing three \emph{different} implementations\footnote[2]{We assume that the combinations of the feature dimensions or classes are linear, otherwise the adjustment requires prohibitive $\mathcal{O}(2^n)$ sampling. We will relax this assumption in future work.} for $g(\mathbf{x},d)$,  $P(Y|X, D, C)$, and $P(D)$ in Eq.~\eqref{eq:backdoor}. In particular, the exact forms of $P(Y|\cdot)$ across different classifiers are given in Appendix 5.


\noindent\textbf{Feature-wise Adjustment}.
Suppose that $\mathcal{F}$ is the index set of the feature dimensions of $\mathbf{x}$, \eg, from the last-layer of the pre-trained network $\Omega$. We divide $\mathcal{F}$ into $n$ equal-size disjoint subsets, 
\eg, the output feature dimension of ResNet-10 is 512, if $n = 8$, the $i$-th set will be a feature dimension index set of size 512/8 = 64, \ie, $\mathcal{F}_i = \{64(i-1)+1, ..., 64i\}$. The stratum set of pre-trained knowledge is defined as $D:=\{d_1,\ldots,d_n\}$, where each $d_i=\mathcal{F}_i$.

\noindent\textbf{(\romannumeral 1)} $g(\mathbf{x}, d_i) \coloneqq  \{k|k\in \mathcal{F}_i\cap \mathcal{I}_{t}\}$, where $\mathcal{I}_{t}$ is an index set whose corresponding absolute values in $\mathbf{x}$ are larger than the threshold $t$. The reason is simple: if a feature dimension is inactive in $\mathbf{x}$, its corresponding adjustment can be omitted. We set $t$=1e-3 in this paper.

\noindent\textbf{(\romannumeral 2) $P(Y|X, D, C) = P(Y|[\mathbf{x}]_{c})$}, where $c = g(\mathbf{x}, d_i)$ is implemented as the index set defined above, 
$[\mathbf{x}]_c=\{x_k\}_{k\in c}$ is a feature selector which selects the dimensions of $\mathbf{x}$ according to the index set $c$. The classifier takes the adjusted feature $[\mathbf{x}]_{c}$ as input. Note that $d$ is already absorbed in $c$, so $[\mathbf{x}]_{c}$ is essentially a function of $(X, D, C)$.

\noindent\textbf{(\romannumeral 3)} $P(d_i) = 1/n$, where we assume a uniform prior for the adjusted features.

\noindent\textbf{(\romannumeral 4)} The overall feature-wise adjustment is:
\begin{equation}
    P(Y|do(X=\boldsymbol{x}))=\frac{1}{n}\sum_{i=1}^{n} P(Y|[\mathbf{x}]_c),~~~\textrm{where}~~c = {\{k|k\in \mathcal{F}_i~\cap ~\mathcal{I}_{t}\}}.
    \label{eq:featurewise_adjustment}
\end{equation}
It is worth noting that the feature-wise adjustment is always applicable, as we can always have the feature representation $\mathbf{x}$ from the pre-trained network. Interestingly, our feature-wise adjustment sheds some light on the theoretical justifications for the multi-head trick in transformers~\cite{vaswani2017attention}. We will explore this in future work.

\noindent\textbf{Class-wise Adjustment}. Suppose that there are $m$ pre-training classes, denoted as $\mathcal{A} = \{a_1,\ldots a_{m}\}$. In class-wise adjustment, each stratum of pre-trained knowledge is defined as a pre-training class, \ie, $D:=\{d_1,\ldots,d_m\}$ and each $d_i=a_i$.

\noindent\textbf{(\romannumeral 1)} $g(\mathbf{x},d_i) \coloneqq  P(a_i|\mathbf{x})\mathbf{\bar x}_i$, where $P(a_i|\mathbf{x})$ is the pre-trained classifier's probability output that $\mathbf{x}$ belongs to class $a_i$, and $\mathbf{\bar x}_i$ is the mean feature of pre-training samples from class $a_i$. Note that unlike feature-wise adjustment where $c$ is an index set, here $c=g(\mathbf{x},d_i)$ is implemented as a real vector.

\noindent\textbf{(\romannumeral 2)} $P(Y|X, D, C) = P(Y|\mathbf{x} \oplus g(\mathbf{x},d_i))$, where $\oplus$ denotes vector concatenation.

\noindent\textbf{(\romannumeral 3)} $P(d_i) = 1/m$, where we assume a uniform prior of each class.

\noindent\textbf{(\romannumeral 4)} The overall class-wise adjustment is:

\begin{equation}
    P(Y|do(X=\mathbf{x})) = \frac{1}{m} \sum_{i=1}^{m} P(Y|\mathbf{x} \oplus  P\left(a_i|\mathbf{x})\mathbf{\bar x}_i\right) \approx P(Y|\mathbf{x} \oplus \frac{1}{m} \sum_{i=1}^{m} P\left(a_i|\mathbf{x})\mathbf{\bar x}_i\right),
    \label{eq:classwise_adjustment}
\end{equation}

where we adopt the Normalized Weighted Geometric Mean (NWGM)~\cite{xu2015show,xu2020deconfounded} approximation to move the outer sum $\sum P$ into the inner $P(\sum)$. This greatly reduces the network forward-pass consumption as $m$ is usually large in pre-training dataset. Please refer to Appendix 3 for the detailed derivation.


\noindent\textbf{Combined Adjustment}. We can combine feature-wise and class-wise adjustment to make the stratification in backdoor adjustment much more fine-grained. Our combination is simple: applying feature-wise adjustment after class-wise adjustment. Thus, we have:
\begin{equation}
    P(Y|do(X=\mathbf{x})) \approx \frac{1}{n} \sum_{i=1}^n P(Y|[\mathbf{x}]_{c} \oplus \frac{1}{m} \sum_{j=1}^{m} [P(a_j|\mathbf{x})\mathbf{\bar x}_j]_{c}),~ \textrm{where}~c= {\{k|k\in \mathcal{F}_i~\cap ~\mathcal{I}_{t}\}}.
    \label{eq:combined_adjustment}
\end{equation}


\section{Related Work}
\label{sec:4}

\noindent\textbf{Few-Shot Learning}. FSL has a wide spectrum of methods, including fine-tuning~\cite{chen2019closer,dhillon2019baseline}, optimizing model initialization~\cite{finn2017model,nichol2018reptile}, generating model parameters~\cite{rusu2018meta, li2019lgm}, learning a feature space for a better separation of sample categories~\cite{vinyals2016matching,zhang2020deepemd}, feature transfer~\cite{sun2019meta,oreshkin2018tadam}, and transductive learning that additionally uses query set data~\cite{dhillon2019baseline,hu1empirical,hou2019cross}.
Thanks to them, the classification accuracy has been drastically increased~\cite{hu1empirical,zhang2020deepemd,ye2020fewshot,yaoyao2020ensemble}. 
However, accuracy as a single number cannot explain the paradoxical phenomenon in Figure~\ref{fig:2}.
Our work offers an answer from a causal standpoint by showing that pre-training is a confounder. We not only further improve the accuracy of various FSL methods, but also explain the reason behind the improvements. In fact, the perspective offered by our work can benefit all the tasks that involve pre-training---any downstream task can be seen as FSL compared to the large-scale pre-training data.

\noindent\textbf{Negative Transfer}. The above phenomenon is also known as the negative transfer, where learning in source domain contributes negatively to the performance in target domain~\cite{pan2009survey}.
Many research works have being focused on when and how to conduct this transfer learning~\cite{huh2016makes,azizpour2015factors,zhang2020impact}.
Yosinski \etal~\cite{yosinski2014transferable} split ImageNet according to man-made objects and natural objects as a test bed for feature transferability. They resemble the $\mathcal{S}\not\sim\mathcal{Q}$ settings used in Figure~\ref{fig:2a}. 
Other work also revealed that using
deeper backbone might lead to degraded performance when the domain gap between training and 
test is large~\cite{kornblith2019better}. 
Some similar findings are reported in the few-shot setting~\cite{ramalho2019empirical} 
and NLP tasks~\cite{tanti2019transfer}. Unfortunately, they didn't provide a theoretical explanation why it happens.

\noindent\textbf{Causal Inference}. Our work aims to deal with the pre-training confounder in FSL based on causal inference~\cite{pearl2009causality}. Causal inference was recently introduced to machine learning~\cite{magliacane2018domain,bengio2019meta} and has been applied to various fields in computer vision. ~\cite{xu2020deconfounded} proposes a retrospective for image captioning and other applications include image classification~\cite{chalupka2014visual,lopez2017discovering}, imitation learning~\cite{de2019causal}, long-tailed recognition~\cite{kaihua2020long} and semantic segmentation~\cite{dong2020causal}. We are the first to approach FSL from a causal perspective. 
We would like to highlight that data-augmentation based FSL can also be considered as approximated intervention. These methods learn to generate additional support samples with image deformation~\cite{chen2019image,zhang2019few} or generative models~\cite{antoniou2017data,zhang2018metagan}. This can be view as physical interventions on the image features. Regarding the causal relation between image $X$ and label $Y$, some works adopted anti-causal learning~\cite{mingming2016domain}, \ie, $Y\to X$, where the assumption is that labels $Y$ are disentangled enough to be treated as Independent Mechanism (IM)~\cite{giambattista2018learning,raphael2019robustly}, which generates observed images $X$ through $Y\to X$. However, our work targets at the more general case where labels can be entangled (\eg ``lion'' and ``dog'' share the semantic ``soft fur'') and the IM assumption may not hold.
Therefore, we use causal prediction $X\to Y$ as it is essentially a reasoning process, where the IM is captured by $D$, which is engineered to be disentangled through CNN (\eg, the conv-operations are applied independently). In this way, $D$ generates visual features through $D\to X$ and emulates human's naming process through $D\to Y$ (\eg, ``fur'', ``four-legged''$\to$ ``meerkat''). In fact, 
the causal direction $X\to Y$ (NOT anti-causal $Y\to X$) has been empirically justified in complex CV tasks~\cite{qi2019two,tan2019visual,kaihua2020long,tang2020unbiased}.
\section{Experiments}
\label{sec:5}
\subsection{Datasets and Settings}
\label{sec:5.1}
\noindent\textbf{Datasets}. We conducted experiments on benchmark datasets in FSL literature: 1) \textbf{\textit{mini}ImageNet}~\cite{vinyals2016matching} containing 600 images per class over 100 classes. We followed the split proposed in~\cite{ravi2016optimization}: 64/16/20 classes for train/val/test.
%
2) \textbf{\textit{tiered}ImageNet}~\cite{ren2018meta} is much larger compared to \textit{mini}ImageNet with 608 classes and each class around 1,300 samples.
These classes were grouped into 34 higher-level concepts and then partitioned into 20/6/8 disjoint sets for train/val/test to achieve larger domain difference between training and testing.
3) Caltech-UCSD Birds-200-2011 (\textbf{CUB})~\cite{WelinderEtal2010} for cross-domain evaluation. It contains 200 classes and each class has around 60 samples. 
The models used for CUB test were trained on the \textit{mini}ImageNet.
Training and evaluation settings on \textit{mini}ImageNet and \textit{tiered}ImageNet are included in Appendix 5.

\noindent\textbf{Implementation Details}. 
We pre-trained the 10-layer ResNet (ResNet-10)~\cite{he2016deep} and the WideResNet (WRN-28-10)~\cite{zagoruyko2016wide} as our backbones.
Our proposed IFSL supports both fine-tuning and meta-learning.
For fine-tuning, we applied average pooling on the last residual block and used the pooled features to train classifiers.
%
For meta-learning, we deployed 5 representative methods that cover a large spectrum of meta-learning based FSL: 1) model initialization: MAML~\cite{finn2017model}, 2) weight generator: LEO~\cite{rusu2018meta}, transductive learning: SIB~\cite{hu1empirical}, 4) metric learning: MatchingNet (MN)~\cite{vinyals2016matching}, and 5) feature transfer: MTL~\cite{sun2019meta}.
%
For both fine-tuning and meta-learning, our IFSL aims to the learn classifier $P(Y|do(X))$ instead of the conventional $P(Y|X)$.
Detailed implementations are given in Appendix 5.

\noindent\textbf{Evaluation Metrics}. Our evaluation is based on the following metrics: 
1) Conventional accuracy (\textbf{Acc}) is the average classification accuracy commonly used in FSL~\cite{finn2017model,vinyals2016matching, sun2019meta}.
2) \textbf{Hardness-specific Acc}.
For each query, we define a hardness that measures its semantic dissimilarity to the support set, and accuracy is then computed at different levels of query hardness. 
Specifically, query hardness is 
computed
by $h = \log \left((1-s)/s\right)$ and $s = exp{\langle \mathbf{r}^{+},\mathbf{p}_{c=gt}^+\rangle}/{\sum\nolimits_c exp\langle\mathbf{r}^+,\mathbf{p}_c^+\rangle}$, where $\langle\cdot\rangle$ is the cosine similarity, $(\cdot)^+$ represents the ReLU activation function, $\mathbf{r}$ denotes the $\Omega$ prediction logits of query, $\mathbf{p}_c$ denotes the average prediction logits of class $c$ in the support set
and $gt$ is the ground-truth of query.
Using \textbf{Hardness-specific Acc} is similar to 
evaluating the hardness of FSL tasks~\cite{dhillon2019baseline}, while
ours is query-sample-specific and hence is more fine-grained.
Later, we will show its effectiveness to
unveil the spurious effects in FSL.
3) Feature localization accuracy (\textbf{CAM-Acc}) quantifies if a model ``pays attention'' to the actual object when making prediction. It is defined as the percentage of pixels inside the object bounding box by using Grad-CAM~\cite{selvaraju2017grad} score larger than $0.9$. Compared to \textbf{Acc} that shows if the prediction is correct, \textbf{CAM-Acc} reveals whether the prediction is based on the correct visual 
cues.

\begin{figure}[h!]
\centering
\captionsetup{font=footnotesize,labelfont=footnotesize,skip=5pt}
\begin{minipage}[t][6cm][t]{.7\textwidth}
    \raggedright
    \fontsize{5.4}{6.4}\selectfont
    \renewcommand\arraystretch{1.2}
    \captionsetup{justification=raggedright,singlelinecheck=false,width=\linewidth}
    \captionof{table}{Acc (\%) averaged over 2000 5-way FSL tasks before and after applying IFSL. We obtained the results by using official code and our backbones for a fair comparison across methods. We also implemented SIB in both transductive and inductive setting to facilitate fair comparison. For IFSL, we reported results of combined adjustment as it almost always outperformed feature-wise and class-wise adjustment. See Appendix 6 for Acc and 95\% confidence intervals on all 3 types of adjustment.}
    \setlength\tabcolsep{1.8pt}
    \begin{tabular}{@{\hskip0pt}c c g g g g g g g g g}
    \hhline{-|-|-|-|-|-|-|-|-|-|-|}
    \hhline{-|-|-|-|-|-|-|-|-|-|-|}
    \rowcolor{white}
    & & & \multicolumn{4}{c}{\textbf{ResNet-10}} & \multicolumn{4}{c}{\textbf{WRN-28-10}} \\
    
    \rowcolor{white}
    & & & \multicolumn{2}{c}{\textit{mini}ImageNet} & \multicolumn{2}{c}{\textit{tiered}ImageNet} & \multicolumn{2}{c}{\textit{mini}ImageNet} & \multicolumn{2}{c}{\textit{tiered}ImageNet} \\ \cmidrule(lr){4-5} \cmidrule(lr){6-7} \cmidrule(lr){8-9} \cmidrule(lr){10-11}
    
    \rowcolor{white}
    \multicolumn{3}{c}{\multirow{-3}{*}[0.3em]{\textbf{Method}}} & $5$-shot & $1$-shot & $5$-shot & $1$-shot & $5$-shot & $1$-shot & $5$-shot & $1$-shot\\
    \hline

    \rowcolor{white}
     & \multicolumn{1}{c}{\multirow{2}{*}{\shortstack{Linear}}} & & 76.38 & 56.26 & 81.01 & 61.39 & 79.79 & 60.69 & 85.37 & 67.27\\
    & \multicolumn{1}{c}{} & +IFSL\textcolor{red}{\scalebox{.8}{+2.19}} & 77.97\textcolor{red}{\scalebox{.8}{+1.59}}& 60.13\textcolor{red}{\scalebox{.8}{+3.87}} & 82.08\textcolor{red}{\scalebox{.8}{+1.07}}& 64.29\textcolor{red}{\scalebox{.8}{+2.9}}& 80.97\textcolor{red}{\scalebox{.8}{+1.18}} & 64.12\textcolor{red}{\scalebox{.8}{+3.43}} & 86.19\textcolor{red}{\scalebox{.8}{+0.82}} & 69.96\textcolor{red}{\scalebox{.8}{+2.69}}\\ \hhline{~;-;-|-|-|-|-|-|-|-|-|}

    \rowcolor{white}
    &\multicolumn{1}{c}{\multirow{2}{*}{\shortstack{Cosine}}} & & 76.68 & 56.40 & 81.13 & 62.08 & 79.72 & 60.83 & 85.41 & 67.30 \\
    & & +IFSL\textcolor{red}{\scalebox{.8}{+1.77}} & 77.63\textcolor{red}{\scalebox{.8}{+0.95}} & 59.84\textcolor{red}{\scalebox{.8}{+3.44}} & 81.75\textcolor{red}{\scalebox{.8}{+0.62}} & 64.47\textcolor{red}{\scalebox{.8}{+2.39}} & 80.74\textcolor{red}{\scalebox{.8}{+1.02}} & 63.76\textcolor{red}{\scalebox{.8}{+2.93}} & 86.13\textcolor{red}{\scalebox{.8}{+0.72}} & 69.36\textcolor{red}{\scalebox{.8}{+2.06}}\\ \hhline{~;-;-|-|-|-|-|-|-|-|-|}

    \rowcolor{white}
    & \multicolumn{1}{c}{\multirow{2}{*}{\shortstack{$k$-NN}}} & & 76.63 & 55.92 & 80.85 & 61.16 & 79.60 & 60.34 & 84.67 & 67.25\\
    \multirow{-6}{*}{\STAB{\rotatebox[origin=c]{90}{\textbf{Fine-Tuning}}}} & & +IFSL\textcolor{red}{\scalebox{.8}{+3.13}} & 78.42\textcolor{red}{\scalebox{.8}{+1.79}} & 62.31\textcolor{red}{\scalebox{.8}{+6.36}} & 81.98\textcolor{red}{\scalebox{.8}{+1.13}} & 65.71\textcolor{red}{\scalebox{.8}{+4.55}} & 81.08\textcolor{red}{\scalebox{.8}{+1.48}} & 64.98\textcolor{red}{\scalebox{.8}{+4.64}} & 86.06\textcolor{red}{\scalebox{.8}{+1.39}} & 70.94\textcolor{red}{\scalebox{.8}{+3.69}}\\
    
    \hline
    
    \rowcolor{white}
    & \multicolumn{1}{c}{\multirow{2}{*}{\shortstack{MAML~\cite{finn2017model}}}} & & 70.85 & 56.59 & 74.02 & 59.17 & 73.92 & 58.02 & 77.20 & 61.40\\
    & & +IFSL\textcolor{red}{\scalebox{.8}{+5.55}} & 76.37\textcolor{red}{\scalebox{.8}{+5.52}} & 59.36\textcolor{red}{\scalebox{.8}{+2.77}} & 81.04\textcolor{red}{\scalebox{.8}{+7.02}} & 63.88\textcolor{red}{\scalebox{.8}{+4.71}} & 79.25\textcolor{red}{\scalebox{.8}{+5.33}} & 62.84\textcolor{red}{\scalebox{.8}{+4.82}} & 85.10\textcolor{red}{\scalebox{.8}{+7.90}} & 67.70\textcolor{red}{\scalebox{.8}{+6.30}}\\ \hhline{~;-;-|-|-|-|-|-|-|-|-|}

    \rowcolor{white}
    &\multicolumn{1}{c}{\multirow{2}{*}{\shortstack{LEO~\cite{rusu2018meta}}}} & & 74.49 & 58.48 & 80.25 & 65.25 & 75.86 & 59.77 & 82.15 & 68.90\\
    & & +IFSL\textcolor{red}{\scalebox{.8}{+1.94}} & 76.91\textcolor{red}{\scalebox{.8}{+2.42}} & 61.09\textcolor{red}{\scalebox{.8}{+2.61}} & 81.43\textcolor{red}{\scalebox{.8}{+1.18}} & 66.03\textcolor{red}{\scalebox{.8}{+0.78}} & 77.72\textcolor{red}{\scalebox{.8}{+1.86}} & 62.19\textcolor{red}{\scalebox{.8}{+2.42}} & 85.04\textcolor{red}{\scalebox{.8}{+2.89}} & 70.28\textcolor{red}{\scalebox{.8}{+1.38}}\\ \hhline{~;-;-|-|-|-|-|-|-|-|-|}

    \rowcolor{white}
    &\multicolumn{1}{c}{\multirow{2}{*}{\shortstack{MTL~\cite{sun2019meta}}}} & & 75.65 & 58.49 & 81.14 & 64.29 & 77.30 & 62.99 & 83.23 & 70.08\\
    & & +IFSL\textcolor{red}{\scalebox{.8}{+2.02}} & 78.03\textcolor{red}{\scalebox{.8}{+2.38}} & 61.17\textcolor{red}{\scalebox{.8}{+2.68}} & 82.35\textcolor{red}{\scalebox{.8}{+1.21}} & 65.72\textcolor{red}{\scalebox{.8}{+1.43}} & 80.20\textcolor{red}{\scalebox{.8}{+2.9}} & 64.40\textcolor{red}{\scalebox{.8}{+1.41}} & 86.02\textcolor{red}{\scalebox{.8}{+2.79}} & 71.45\textcolor{red}{\scalebox{.8}{+1.37}}\\ \hhline{~;-;-|-|-|-|-|-|-|-|-|}
    
    \rowcolor{white}
    &\multicolumn{1}{c}{\multirow{2}{*}{\shortstack{MN~\cite{vinyals2016matching}}}} & & 75.21 & 61.05 & 79.92 & 66.01 & 77.15 & 63.45 & 82.43 & 70.38\\
    & & +IFSL\textcolor{red}{\scalebox{.8}{+1.34}} & 76.73\textcolor{red}{\scalebox{.8}{+1.52}} & 62.64\textcolor{red}{\scalebox{.8}{+1.59}} & 80.79\textcolor{red}{\scalebox{.8}{+0.87}} & 67.30\textcolor{red}{\scalebox{.8}{+1.29}} & 78.55\textcolor{red}{\scalebox{.8}{+1.40}} & 64.89\textcolor{red}{\scalebox{.8}{+1.44}} & 84.03\textcolor{red}{\scalebox{.8}{+1.60}} & 71.41\textcolor{red}{\scalebox{.8}{+1.03}}\\ \hhline{~;-;-|-|-|-|-|-|-|-|-|}
    
    \rowcolor{white}
    &\multicolumn{1}{c}{\multirow{2}{*}{\shortstack{SIB~\cite{hu1empirical}\\(transductive)}}} & & 78.88 & 67.10 & 85.09 & 77.64 & 81.73 & 71.31 & 88.19 & 81.97\\
    & & +IFSL\textcolor{red}{\scalebox{.8}{+1.15}} & 80.32\textcolor{red}{\scalebox{.8}{+1.44}} & 68.85\textcolor{red}{\scalebox{.8}{+1.75}} & 85.43\textcolor{red}{\scalebox{.8}{+0.34}} & 78.03\textcolor{red}{\scalebox{.8}{+0.39}} & 83.21\textcolor{red}{\scalebox{.8}{+1.48}} & 73.51\textcolor{red}{\scalebox{.8}{+2.20}} & 88.69\textcolor{red}{\scalebox{.8}{+0.50}} & 83.07\textcolor{red}{\scalebox{.8}{+1.10}}\\ \hhline{~;-;-|-|-|-|-|-|-|-|-|}
    
    \rowcolor{white}
    &\multicolumn{1}{c}{\multirow{2}{*}{\shortstack{SIB~\cite{hu1empirical}\\(inductive)}}} & & 75.64 & 57.20 & 81.69 & 65.51 & 78.17 & 60.12 & 84.96 & 69.20\\
    \parbox[t]{2mm}{\multirow{-12}{*}{\rotatebox[origin=c]{90}{\textbf{Meta-Learning}}}} & & +IFSL\textcolor{red}{\scalebox{.8}{+2.05}} & 77.68\textcolor{red}{\scalebox{.8}{+2.04}} & 60.33\textcolor{red}{\scalebox{.8}{+3.13}} & 82.75\textcolor{red}{\scalebox{.8}{+1.06}} & 67.34\textcolor{red}{\scalebox{.8}{+1.83}} & 80.05\textcolor{red}{\scalebox{.8}{+1.88}} & 63.14\textcolor{red}{\scalebox{.8}{+3.02}} & 86.14\textcolor{red}{\scalebox{.8}{+1.18}} & 71.45\textcolor{red}{\scalebox{.8}{+2.25}}\\
    \hhline{-|-|-|-|-|-|-|-|-|-|-|}
    \hhline{-|-|-|-|-|-|-|-|-|-|-|}
\end{tabular}
    \label{tab:1}
\end{minipage}%
\begin{minipage}[t][8.5cm][b]{.3\textwidth}
    \centering
    \begin{subfigure}[b]{0.85\linewidth}
        \centering
        \addtocounter{subfigure}{-3}
        \addtocounter{figure}{1}
        \includegraphics[width=1\linewidth]{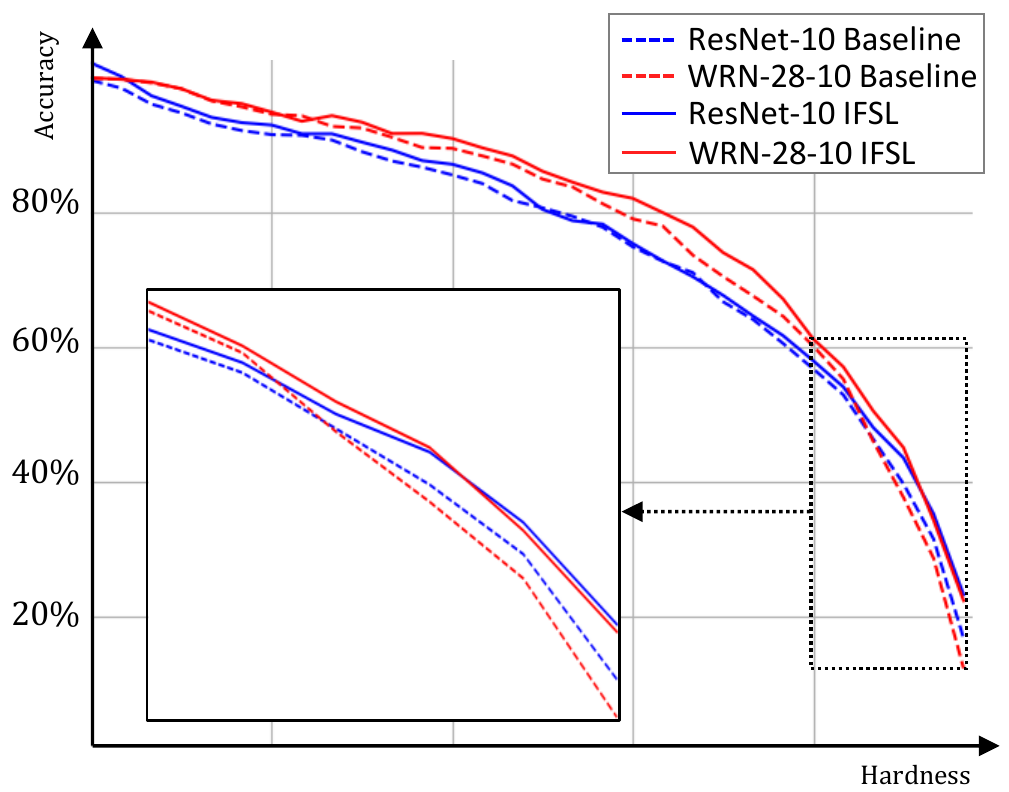}
        \caption{Fine-Tuning (Linear)}
        \label{fig:5a} 
    \end{subfigure}%

    \begin{subfigure}[b]{0.85\linewidth}
        \centering
        \includegraphics[width=1\linewidth]{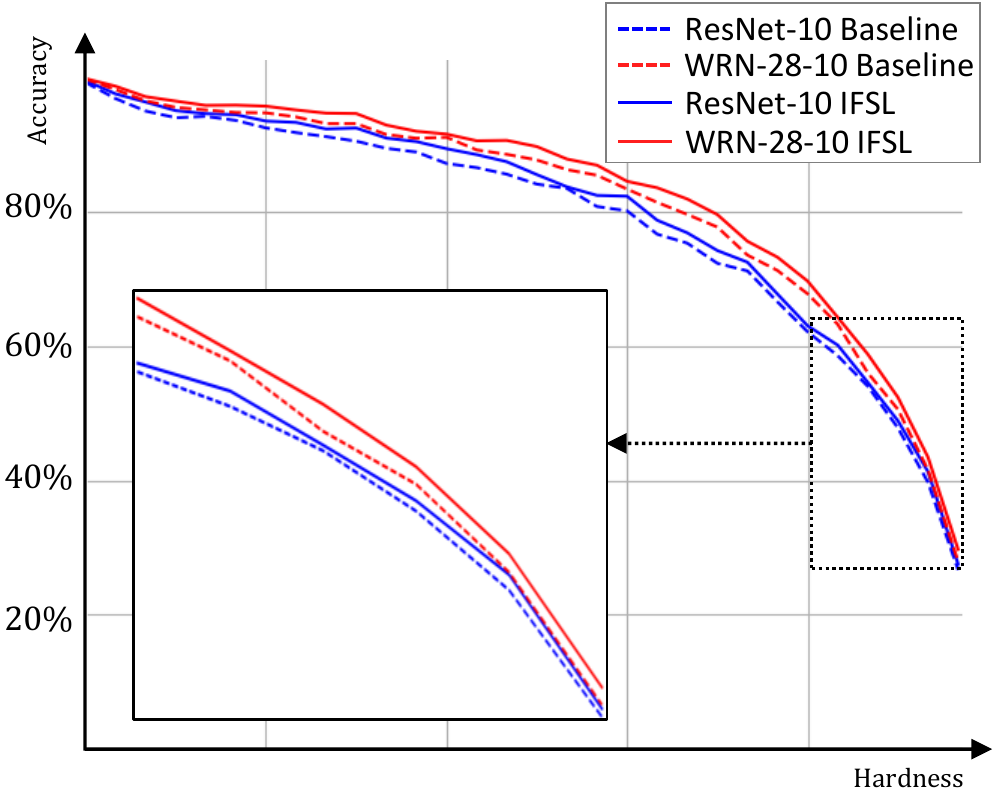}
        \caption{Meta-Learning (SIB)}
        \label{fig:5b}
    \end{subfigure}
    \captionsetup{width=0.9\linewidth}
    \addtocounter{figure}{-1}
    \captionof{figure}{Accuracy across query hardness on 5-shot fine-tuning and meta-learning. Additional results are shown in Appendix 6.}
    \label{fig:5}
\end{minipage}
\vspace{-2mm}
\end{figure}

\begin{figure}[h!]
\centering
\captionsetup{font=footnotesize,labelfont=footnotesize,skip=5pt}
\begin{minipage}[t][5.2cm][t]{.52\textwidth}
    \centering
    \fontsize{6.3}{7.56}\selectfont
    \renewcommand\arraystretch{1.1}
    \captionof{table}{Comparison with state-of-the-arts of 5-way 1-/5-shot Acc (\%) on \textit{mini}ImageNet and \textit{tiered}ImageNet.}
    \setlength\tabcolsep{4pt}
    \begin{tabular}{c c c c c c}
    \toprule
    \multicolumn{1}{c}{\multirow{2}{*}{\textbf{Method}}} & \multicolumn{1}{c}{\multirow{2}{*}{\textbf{Backbone}}} & \multicolumn{2}{c}{\textbf{\textit{mini}ImageNet}} & \multicolumn{2}{c}{\textbf{\textit{tiered}ImageNet}}\\
    & & $5$-shot & $1$-shot & $5$-shot & $1$-shot\\
    \midrule
    Baseline++~\cite{chen2019closer} & ResNet-10 & 75.90 & 53.97& - & -\\
    IdeMe-Net$^\dagger$~\cite{chen2019image} & ResNet-10 & 73.78 & 57.61 & 80.34 & 60.32\\
    TRAML~\cite{li2020boosting} & ResNet-12 & 79.54 & 67.10 & - & - \\
    DeepEMD~\cite{zhang2020deepemd} & ResNet-12 & 82.41 & 65.91 & 86.03 & 71.16 \\
    CTM~\cite{li2019finding} & ResNet-18 & 80.51 & 64.12 & 84.28 & 68.41\\
    FEAT~\cite{ye2020fewshot} & WRN-28-10 & 81.80 & 66.69 & 84.38 & 70.41 \\
    Tran. Baseline~\cite{dhillon2019baseline} & WRN-28-10 & 78.40 & 65.73 & 85.50 & 73.34 \\
    wDAE-GNN~\cite{gidaris2019generating} & WRN-28-10 & 78.85 & 62.96 & 83.09 & 68.18 \\
    SIB$^\dagger$~\cite{hu1empirical} & WRN-28-10 & 81.73 & 71.31 & 88.19 & 81.97\\
    \midrule
    \textbf{SIB+IFSL (ours)} & WRN-28-10 & \textbf{83.21} & \textbf{73.51} & \textbf{88.69} & \textbf{83.07}\\
    \bottomrule
    \multicolumn{3}{c}{$^\dagger$Using our pre-trained backbone.} & &
\end{tabular}
    \label{tab:2}
\end{minipage}%
\begin{minipage}[t][5.2cm][t]{.48\textwidth}
    \centering
    \fontsize{6.3}{7.56}\selectfont
    \renewcommand\arraystretch{1.32}
    \captionsetup{width=0.7\linewidth}
    \captionof{table}{Results of cross-domain evaluation: \textit{mini}ImageNet $\rightarrow$ CUB. The whole report is in Appendix 6.}
    
    

\begin{tabular}{c c g@{\hskip6pt} g@{\hskip6pt} g@{\hskip6pt}}
    \toprule
    
    \rowcolor{white}
    \textbf{Backbone} & \multicolumn{2}{c}{\textbf{Method}} & \textbf{5-shot} & \textbf{1-shot} \\
    \hline
    
    \rowcolor{white}
    & \multicolumn{1}{c}{\multirow{2}{*}{Linear}} &  & 58.84 & 42.25\\
    & & +IFSL & 60.65 & 45.14\\ \hhline{~;-;-|-|-}
    
    \rowcolor{white}
    & \multicolumn{1}{c}{\multirow{2}{*}{SIB}} & & 60.60 & 45.87\\
    \multicolumn{1}{c}{\multirow{-4}{*}{ResNet-10}} & & +IFSL & 62.07 & 47.07\\
    
    \hline
    
    \rowcolor{white}
     & \multicolumn{1}{c}{\multirow{2}{*}{Linear}} &  & 62.12 & 42.89\\
    & & +IFSL & 64.15 & 45.64\\ \hhline{~;-;-|-|-}
    
    \rowcolor{white}
    & \multicolumn{1}{c}{\multirow{2}{*}{SIB}} & & 62.59 & 49.16\\
    \multicolumn{1}{c}{\multirow{-4}{*}{WRN-28-10}} & & +IFSL & 64.43 & 50.71\\
    \hhline{-;-;-|-|-}
    \hhline{-;-;-|-|-}
\end{tabular}

    \label{tab:3}
\end{minipage}
\end{figure}

\subsection{Results and Analysis}
\label{sec:5.2}

\noindent\textbf{Conventional Acc}.
1) From Table~\ref{tab:1}, we observe that IFSL consistently improves fine-tuning and meta-learning in all settings, which suggests that IFSL is agnostic to methods, datasets, and backbones.
2) In particular, the improvements are typically larger on 1-shot than 5-shot. For example, in fine-tuning, the average performance gain is 1.15\% on 5-shot and 3.58\% on 1-shot. 
The results support our analysis in Section~\ref{sec:2.3} that FSL models are more prone to bias in lower-shot settings. 
3)
Regarding the average improvements on fine-tuning vs. meta-learning (\eg $k$-NN and MN), we observe  that IFSL improves more on fine-tuning in most cases. We conjecture that this is because meta-learning is an implicit form of intervention, where randomly sampled meta-training episodes effectively stratify the pre-trained knowledge. This suggests that meta-learning is fundamentally superior over fine-tuning due to increased robustness against confounders. We will investigate this potential theory in future work.
4) Additionally we see that the improvements on \textit{mini}ImageNet are usually larger than that on \textit{tiered}ImageNet. A possible reason is the much larger training set for \textit{tiered}ImageNet: it substantially increases the breadth of the pre-trained knowledge and the resulting models explain query samples much better.
5) According to Table~\ref{tab:1} and Table~\ref{tab:2}, it is clear that our $k$-NN+IFSL outperforms IdeMe-Net~\cite{chen2019image} using the same pre-trained ResNet-10. 
This shows that using data augmentation --- a method of physical data intervention as in IdeMe-Net~\cite{chen2019image} is inferior to our causal intervention in IFSL.
6) 
Overall, our IFSL achieves the new state-of-the-art on both datasets. Note that IFSL is flexible to be plugged into different baselines.


\begin{figure}[t!]
 \centering
 \captionsetup{font=footnotesize,labelfont=footnotesize}
 \includegraphics[scale=0.5]{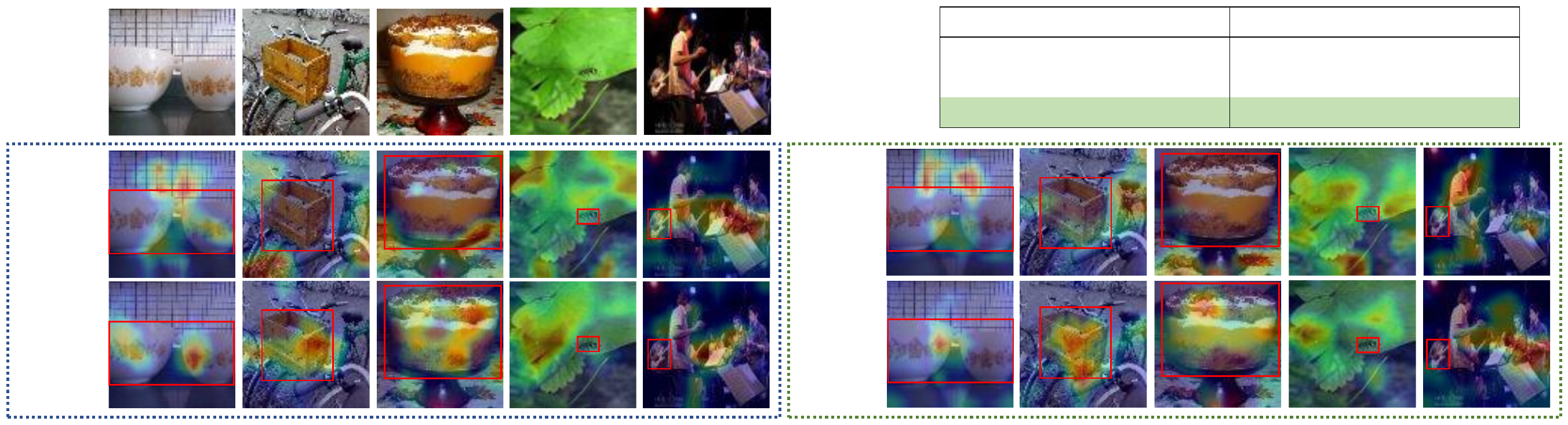}
 \vspace{-2mm}
 \begin{tikzpicture}[overlay, remember picture]
    \node [below left,text width=3cm,align=center] at (-10.6,4) {\scalebox{.5}{\textit{``mixing bowl''}}};
    \node [below left,text width=3cm,align=center] at (-9.45,4) {\scalebox{.5}{\textit{``crate''}}};
    \node [below left,text width=3cm,align=center] at (-8.27,4) {\scalebox{.5}{\textit{``trifle''}}};
    \node [below left,text width=3cm,align=center] at (-7.1,4) {\scalebox{.5}{\textcolor{red}{\textit{``ant''}}}};
    \node [below left,text width=3cm,align=center] at (-5.9,4) {\scalebox{.5}{\textcolor{red}{\textit{``electric guitar''}}}};
    
    \node [below left,text width=3cm,align=center] at (-11.7,3.35) {\scalebox{.6}{{Query}}};
    \node [below left,text width=3cm,align=center] at (-11.7,2.1) {\scalebox{.6}{{Linear}}};
    \node [below left,text width=3cm,align=center] at (-11.7,0.9) {\scalebox{.6}{{+IFSL}}};
    
    \node [below left,text width=3cm,align=center] at (-4.85,2.1) {\scalebox{.6}{{MAML}}};
    \node [below left,text width=3cm,align=center] at (-4.85,0.9) {\scalebox{.6}{{+IFSL}}};
    
    \node [below left,text width=3cm,align=center] at (-4.75,3.35) {\scalebox{.6}{{CAM-Acc}}};
    
    \node [below left,text width=3cm,align=center] at (-2.75,3.72) {\scalebox{.5}{{5-shot}}};
    \node [below left,text width=3cm,align=center] at (-0.16,3.72) {\scalebox{.5}{{1-shot}}};
    
    \node [below left,text width=3cm,align=center] at (-2.75,3.46) {\scalebox{.5}{Linear}};
    \node [below left,text width=3cm,align=center] at (-1.95,3.458) {\scalebox{.5}{MAML}};
    \node [below left,text width=3cm,align=center] at (-0.16,3.46) {\scalebox{.5}{Linear}};
    \node [below left,text width=3cm,align=center] at (0.64,3.458) {\scalebox{.5}{MAML}};
    
    \node [below left,text width=3cm,align=center] at (-2.75,3.21) {\scalebox{.5}{29.02}};
    \node [below left,text width=3cm,align=center] at (-1.95,3.21) {\scalebox{.5}{29.43}};
    \node [below left,text width=3cm,align=center] at (-0.16,3.21) {\scalebox{.5}{25.22}};
    \node [below left,text width=3cm,align=center] at (0.64,3.21) {\scalebox{.5}{27.39}};
    
    \node [below left,text width=3cm,align=center] at (-3.45,2.95) {\scalebox{.5}{{+IFSL}}};
    \node [below left,text width=3cm,align=center] at (-2.75,2.95) {\scalebox{.5}{29.85}};
    \node [below left,text width=3cm,align=center] at (-1.95,2.95) {\scalebox{.5}{30.06}};
    \node [below left,text width=3cm,align=center] at (-0.9,2.95) {\scalebox{.5}{{+IFSL}}};
    \node [below left,text width=3cm,align=center] at (-0.16,2.95) {\scalebox{.5}{26.67}};
    \node [below left,text width=3cm,align=center] at (0.64,2.95) {\scalebox{.5}{28.42}};
 \end{tikzpicture}
 \caption{Some \textit{mini}ImageNet visualizations of Grad-Cam~\cite{selvaraju2017grad} activation of query images and the CAM-Acc (\%) table of using linear classifier and MAML. Categories with red text represent failed cases. The complete results on CAM-Acc are shown in Appendix 6, where IFSL achieves similar or better results in all settings.}
 \label{fig:6}
 \vspace{-5mm}
\end{figure}

\noindent\textbf{Hardness-specific Acc.}
1) Figure~\ref{fig:5a} shows the plot of Hardness-specific Acc of fine-tuning. We notice that
when query becomes harder, ResNet-10 (blue curves) 
becomes superior to WRN-28-10 (red curves).
This tendency is 
consistent with Figure~\ref{fig:2a} illustrating the effect of the confounding bias caused by pre-training.
2) 
Intriguingly, in Figure~\ref{fig:5b}, we notice that this tendency is reversed for meta-learning, \ie,
deeper backbone always performs better.
The improved performance of deeper backbone on hard queries suggests that meta-learning should have some functions to remove the confounding bias. This evidence will inspire us to provide a causal view of meta-learning in future work.
3) 
%
Overall, Figure~\ref{fig:5} shows that using IFSL futher improves fine-tuning and meta-learning consistently across all hardness, validating the effectiveness of the proposed causal intervention.

\noindent\textbf{CAM-Acc \& Visualization.}
In Figure~\ref{fig:6}, 
we compare +IFSL to baseline linear classifier
on the left and to baseline MAML~\cite{finn2017model} on the right, and summarize CAM-Acc results
in the upper-right table.
From the visualization, we see that using IFSL let the model pay more attention to the objects.
However, notice that all models failed in the categories colored as red. A possible reason behind the failures is the extremely small size of the object --- models have to resort to context for prediction.
From the numbers, we can see our improvements for 1-shot are larger than that for 5-shot, consistent with our findings using other evaluation metrics. These results suggest that IFSL helps models use the correct visual semantics for prediction by removing the confounding bias.


\noindent\textbf{Cross-Domain Generalization Ability.}
In Table~\ref{tab:3}, we show the testing results on  CUB using the models trained on the \textit{mini}ImageNet. The setting is challenging due to the big domain gap between the two datasets. We chose linear classifier as it outperforms cosine and $k$-NN in cross-domain setting and compared with transductive method --- SIB.
%
The results clearly show that IFSL works well in this setting and brings consistent improvements, with the average 1.94\% of Acc.
%
In addition, we can see that 
applying IFSL brings larger improvements to the inductive linear classifier than to the transductive SIB. It is possibly because transductive methods involve unlabeled query data and performs better than inductive methods with the additional information. Nonetheless we observe that IFSL can further improve SIB in cross-domain (Table~\ref{tab:3}) and single-domain (Table~\ref{tab:1}) generalization.
\vspace{-2mm}
\section{Conclusions}
\vspace{-2mm}
We presented a novel casual framework: Interventional Few-Shot Learning (IFSL), to address an overlooked deficiency in recent FSL methods: the pre-training is a confounder hurting the performance. Specifically, we proposed a structural causal model of the causalities in the process of FSL and then developed three practical implementations based on the backdoor adjustment. 
To better illustrate the deficiency, we diagnosed the classification accuracy comprehensively across query hardness, and showed that IFSL improves all the baselines across all the hardness. It is worth highlighting that the contribution of IFSL is not only about improving the performance of FSL, but also offering a causal explanation why IFSL works well: it is a causal approximation to  many-shot learning. We believe that IFSL may shed light on exploring the new boundary of FSL, even though FSL is well-known to be ill-posed due to insufficient data. To upgrade IFSL, we will seek other observational intervention algorithms for better performance, and devise counterfactual reasoning for more general few-shot settings such as domain transfer.

\section{Acknowledgements}

The authors would like to thank all the anonymous reviewers for their constructive comments and suggestions. This research is partly supported by the Alibaba-NTU Singapore Joint Research Institute, Nanyang Technological University (NTU), Singapore; the Singapore Ministry of Education (MOE) Academic Research Fund (AcRF) Tier 1 and Tier 2 grant; and Alibaba Innovative Research (AIR) programme. We also want to thank Alibaba City Brain Group for the donations of GPUs.
\section{Broader Impact}

The proposed method aims to improve the Few-Shot Learning task. Advancements in FSL helps the deployment of machine learning models in areas where labelled data is difficult or expensive to obtain and it is closely related to social well-beings: few-shot drug discovery or medical imaging analysis in medical applications, cold-start item recommendation in e-commerce, few-shot reinforcement learning for industrial robots, \etc. Our method is based on causal inference and the analysis is rooted on causation rather than correlation. The marriage between causality and machine learning can produce more robust, transparent and explainable models, broadening the applicability of ML models and promoting fairness in artificial intelligence.

\bibliography{references/references}

\begin{thebibliography}{10}

\bibitem{angrist2001instrumental}
Joshua~D Angrist and Alan~B Krueger.
\newblock Instrumental variables and the search for identification: From supply
  and demand to natural experiments.
\newblock {\em Journal of Economic Perspectives}, 2001.

\bibitem{antoniou2017data}
Antreas Antoniou, Amos Storkey, and Harrison Edwards.
\newblock Data augmentation generative adversarial networks.
\newblock In {\em Proceedings of the International Conference on Learning
  Representations Workshops}, 2018.

\bibitem{arora2019implicit}
Sanjeev Arora, Nadav Cohen, Wei Hu, and Yuping Luo.
\newblock Implicit regularization in deep matrix factorization.
\newblock In {\em Advances in Neural Information Processing Systems}, 2019.

\bibitem{azizpour2015factors}
Hossein Azizpour, Ali~Sharif Razavian, Josephine Sullivan, Atsuto Maki, and
  Stefan Carlsson.
\newblock Factors of transferability for a generic convnet representation.
\newblock {\em IEEE transactions on Pattern Analysis and Machine Intelligence},
  2015.

\bibitem{baldi2014dropout}
Pierre Baldi and Peter Sadowski.
\newblock The dropout learning algorithm.
\newblock {\em Artificial intelligence}, 2014.

\bibitem{balke1997bounds}
Alexander Balke and Judea Pearl.
\newblock Bounds on treatment effects from studies with imperfect compliance.
\newblock {\em Journal of the American Statistical Association}, 1997.

\bibitem{bengio2012deep}
Yoshua Bengio.
\newblock Deep learning of representations for unsupervised and transfer
  learning.
\newblock In {\em Proceedings of ICML Workshop on Unsupervised and Transfer
  Learning}, 2012.

\bibitem{bengio2013representation}
Yoshua Bengio, Aaron Courville, and Pascal Vincent.
\newblock Representation learning: A review and new perspectives.
\newblock {\em IEEE Transactions on Pattern Analysis and Machine Intelligence},
  2013.

\bibitem{bengio2019meta}
Yoshua Bengio, Tristan Deleu, Nasim Rahaman, Rosemary Ke, S{\'e}bastien
  Lachapelle, Olexa Bilaniuk, Anirudh Goyal, and Christopher Pal.
\newblock A meta-transfer objective for learning to disentangle causal
  mechanisms.
\newblock In {\em International Conference on Learning Representations}, 2019.

\bibitem{besserve2018counterfactuals}
Michel Besserve, R{\'e}my Sun, and Bernhard Sch{\"o}lkopf.
\newblock Counterfactuals uncover the modular structure of deep generative
  models.
\newblock {\em arXiv preprint arXiv:1812.03253}, 2018.

\bibitem{candes2011robust}
Emmanuel~J Cand{\`e}s, Xiaodong Li, Yi~Ma, and John Wright.
\newblock Robust principal component analysis?
\newblock {\em Journal of the ACM (JACM)}, 2011.

\bibitem{chalupka2014visual}
Krzysztof Chalupka, Pietro Perona, and Frederick Eberhardt.
\newblock Visual causal feature learning.
\newblock In {\em Uncertainty in Artificial Intelligence}, 2015.

\bibitem{chen2019closer}
Wei-Yu Chen, Yen-Cheng Liu, Zsolt Kira, Yu-Chiang~Frank Wang, and Jia-Bin
  Huang.
\newblock A closer look at few-shot classification.
\newblock In {\em International Conference on Learning Representations}, 2019.

\bibitem{chen2019image}
Zitian Chen, Yanwei Fu, Yu-Xiong Wang, Lin Ma, Wei Liu, and Martial Hebert.
\newblock Image deformation meta-networks for one-shot learning.
\newblock In {\em Proceedings of the IEEE Conference on Computer Vision and
  Pattern Recognition}, 2019.

\bibitem{de2019causal}
Pim de~Haan, Dinesh Jayaraman, and Sergey Levine.
\newblock Causal confusion in imitation learning.
\newblock In {\em Advances in Neural Information Processing Systems}, 2019.

\bibitem{Dekking2005}
F.M. Dekking, C.~Kraaikamp, H.P. Lopuha{\"a}, and L.E. Meester.
\newblock {\em A Modern Introduction to Probability and Statistics:
  Understanding Why and How}.
\newblock Springer Texts in Statistics. Springer, 2005.

\bibitem{devlin2018bert}
Jacob Devlin, Ming-Wei Chang, Kenton Lee, and Kristina Toutanova.
\newblock Bert: Pre-training of deep bidirectional transformers for language
  understanding.
\newblock {\em Proceedings of the 2019 Conference of the North American Chapter
  of the Association for Computational Linguistics: Human Language
  Technologies}, 2019.

\bibitem{dhillon2019baseline}
Guneet~S Dhillon, Pratik Chaudhari, Avinash Ravichandran, and Stefano Soatto.
\newblock A baseline for few-shot image classification.
\newblock In {\em International Conference on Learning Representations}, 2020.

\bibitem{fei2006one}
Li~Fei-Fei, Rob Fergus, and Pietro Perona.
\newblock One-shot learning of object categories.
\newblock {\em IEEE Transactions on Pattern Analysis and Machine Intelligence},
  2006.

\bibitem{finn2017model}
Chelsea Finn, Pieter Abbeel, and Sergey Levine.
\newblock Model-agnostic meta-learning for fast adaptation of deep networks.
\newblock In {\em International Conference on Machine Learning}, 2017.

\bibitem{gidaris2019generating}
Spyros Gidaris and Nikos Komodakis.
\newblock Generating classification weights with gnn denoising autoencoders for
  few-shot learning.
\newblock In {\em Proceedings of the IEEE Conference on Computer Vision and
  Pattern Recognition}, 2019.

\bibitem{Detectron2018}
Ross Girshick, Ilija Radosavovic, Georgia Gkioxari, Piotr Doll\'{a}r, and
  Kaiming He.
\newblock Detectron.
\newblock \url{https://github.com/facebookresearch/detectron}, 2018.

\bibitem{goodfellow2016deep}
Ian Goodfellow, Yoshua Bengio, and Aaron Courville.
\newblock {\em Deep Learning}.
\newblock The MIT Press, 2016.

\bibitem{he2017mask}
Kaiming He, Georgia Gkioxari, Piotr Doll{\'a}r, and Ross Girshick.
\newblock Mask r-cnn.
\newblock In {\em Proceedings of the IEEE International Conference on Computer
  Vision}, 2017.

\bibitem{he2016deep}
Kaiming He, Xiangyu Zhang, Shaoqing Ren, and Jian Sun.
\newblock Deep residual learning for image recognition.
\newblock In {\em Proceedings of the IEEE Conference on Computer Vision and
  Pattern Recognition}, 2016.

\bibitem{hinton2015distilling}
Geoffrey Hinton, Oriol Vinyals, and Jeff Dean.
\newblock Distilling the knowledge in a neural network.
\newblock In {\em Advances in Neural Information Processing Systems Deep
  Learning Workshop}, 2014.

\bibitem{hou2019cross}
Ruibing Hou, Hong Chang, MA~Bingpeng, Shiguang Shan, and Xilin Chen.
\newblock Cross attention network for few-shot classification.
\newblock In {\em Advances in Neural Information Processing Systems}, 2019.

\bibitem{howard2017mobilenets}
Andrew~G Howard, Menglong Zhu, Bo~Chen, Dmitry Kalenichenko, Weijun Wang,
  Tobias Weyand, Marco Andreetto, and Hartwig Adam.
\newblock Mobilenets: Efficient convolutional neural networks for mobile vision
  applications.
\newblock {\em arXiv preprint arXiv:1704.04861}, 2017.

\bibitem{hu1empirical}
Shell~Xu Hu, Pablo Moreno, Yang Xiao, Xi~Shen, Guillaume Obozinski, Neil
  Lawrence, and Andreas Damianou.
\newblock Empirical bayes transductive meta-learning with synthetic gradients.
\newblock In {\em International Conference on Learning Representations}, 2020.

\bibitem{huh2016makes}
Minyoung Huh, Pulkit Agrawal, and Alexei~A Efros.
\newblock What makes imagenet good for transfer learning?
\newblock {\em arXiv preprint arXiv:1608.08614}, 2016.

\bibitem{jamal2019task}
Muhammad~Abdullah Jamal and Guo-Jun Qi.
\newblock Task agnostic meta-learning for few-shot learning.
\newblock In {\em Proceedings of the IEEE Conference on Computer Vision and
  Pattern Recognition}, 2019.

\bibitem{qi2019two}
Qi~Jiaxin, Niu Yulei, Huang Jianqiang, and Zhang Hanwang.
\newblock Two causal principles for improving visual dialog.
\newblock In {\em Proceedings of the IEEE Conference on Computer Vision and
  Pattern Recognition}, 2020.

\bibitem{kornblith2019better}
Simon Kornblith, Jonathon Shlens, and Quoc~V Le.
\newblock Do better imagenet models transfer better?
\newblock In {\em Proceedings of the IEEE conference on Computer Vision and
  Pattern Recognition}, 2019.

\bibitem{li2020boosting}
Aoxue Li, Weiran Huang, Xu~Lan, Jiashi Feng, Zhenguo Li, and Liwei Wang.
\newblock Boosting few-shot learning with adaptive margin loss.
\newblock In {\em Proceedings of the IEEE Conference on Computer Vision and
  Pattern Recognition}, 2020.

\bibitem{li2019finding}
Hongyang Li, David Eigen, Samuel Dodge, Matthew Zeiler, and Xiaogang Wang.
\newblock Finding task-relevant features for few-shot learning by category
  traversal.
\newblock In {\em Proceedings of the IEEE Conference on Computer Vision and
  Pattern Recognition}, 2019.

\bibitem{li2019lgm}
Huaiyu Li, Weiming Dong, Xing Mei, Chongyang Ma, Feiyue Huang, and Bao-Gang Hu.
\newblock Lgm-net: Learning to generate matching networks for few-shot
  learning.
\newblock In {\em International Conference on Machine Learning}, 2019.

\bibitem{yaoyao2020ensemble}
Yaoyao Liu, Bernt Schiele, and Qianru Sun.
\newblock An ensemble of epoch-wise empirical bayes for few-shot learning.
\newblock In {\em European Conference on Computer Vision}, 2020.

\bibitem{lopez2017discovering}
David Lopez-Paz, Robert Nishihara, Soumith Chintala, Bernhard Scholkopf, and
  L{\'e}on Bottou.
\newblock Discovering causal signals in images.
\newblock In {\em Proceedings of the IEEE Conference on Computer Vision and
  Pattern Recognition}, 2017.

\bibitem{maaten2008visualizing}
Laurens van~der Maaten and Geoffrey Hinton.
\newblock Visualizing data using t-sne.
\newblock {\em Journal of Machine Learning Research}, 2008.

\bibitem{magliacane2018domain}
Sara Magliacane, Thijs van Ommen, Tom Claassen, Stephan Bongers, Philip
  Versteeg, and Joris~M Mooij.
\newblock Domain adaptation by using causal inference to predict invariant
  conditional distributions.
\newblock In {\em Advances in Neural Information Processing Systems}, 2018.

\bibitem{mingming2016domain}
Gong Mingming, Zhang Kun, Liu Tongliang, Tao Dacheng, Clark Glymour, and
  Bernhard Sch\"{o}lkopf.
\newblock Domain adaptation with conditional transferable components.
\newblock In {\em International Conference on Machine Learning}, 2016.

\bibitem{nichol2018reptile}
Alex Nichol and John Schulman.
\newblock Reptile: a scalable metalearning algorithm.
\newblock {\em arXiv preprint arXiv:1803.02999}, 2018.

\bibitem{oreshkin2018tadam}
Boris Oreshkin, Pau~Rodr{\'\i}guez L{\'o}pez, and Alexandre Lacoste.
\newblock Tadam: Task dependent adaptive metric for improved few-shot learning.
\newblock In {\em Advances in Neural Information Processing Systems}, 2018.

\bibitem{pan2009survey}
Sinno~Jialin Pan and Qiang Yang.
\newblock A survey on transfer learning.
\newblock {\em IEEE Transactions on Knowledge and Data Engineering}, 2009.

\bibitem{giambattista2018learning}
Giambattista Parascandolo, Niki Kilbertus, Mateo Rojas-Carulla, and Bernhard
  Sch\"{o}lkopf.
\newblock Learning independent causal mechanisms.
\newblock In {\em International Conference on Machine Learning}, 2018.

\bibitem{pearl2016causal}
J.~Pearl, M.~Glymour, and N.P. Jewell.
\newblock {\em Causal Inference in Statistics: A Primer}.
\newblock Wiley, 2016.

\bibitem{pearl1995causal}
Judea Pearl.
\newblock Causal diagrams for empirical research.
\newblock {\em Biometrika}, 1995.

\bibitem{pearl2009causality}
Judea Pearl.
\newblock {\em Causality: Models, Reasoning and Inference}.
\newblock Cambridge University Press, 2nd edition, 2009.

\bibitem{qiao2018few}
Siyuan Qiao, Chenxi Liu, Wei Shen, and Alan~L Yuille.
\newblock Few-shot image recognition by predicting parameters from activations.
\newblock In {\em Proceedings of the IEEE Conference on Computer Vision and
  Pattern Recognition}, 2018.

\bibitem{ramalho2019empirical}
Tiago Ramalho, Thierry Sousbie, and Stefano Peluchetti.
\newblock An empirical study of pretrained representations for few-shot
  classification.
\newblock {\em arXiv preprint arXiv:1910.01319}, 2019.

\bibitem{ravi2016optimization}
Sachin Ravi and Hugo Larochelle.
\newblock Optimization as a model for few-shot learning.
\newblock In {\em International Conference on Learning Representations}, 2017.

\bibitem{ren2018meta}
Mengye Ren, Eleni Triantafillou, Sachin Ravi, Jake Snell, Kevin Swersky,
  Joshua~B. Tenenbaum, Hugo Larochelle, and Richard~S. Zemel.
\newblock Meta-learning for semi-supervised few-shot classification.
\newblock In {\em International Conference on Learning Representations}, 2018.

\bibitem{roweis2000nonlinear}
Sam~T Roweis and Lawrence~K Saul.
\newblock Nonlinear dimensionality reduction by locally linear embedding.
\newblock {\em Science}, 2000.

\bibitem{rusu2018meta}
Andrei~A. Rusu, Dushyant Rao, Jakub Sygnowski, Oriol Vinyals, Razvan Pascanu,
  Simon Osindero, and Raia Hadsell.
\newblock Meta-learning with latent embedding optimization.
\newblock In {\em International Conference on Learning Representations}, 2019.

\bibitem{santoro2016meta}
Adam Santoro, Sergey Bartunov, Matthew Botvinick, Daan Wierstra, and Timothy
  Lillicrap.
\newblock Meta-learning with memory-augmented neural networks.
\newblock In {\em International Conference on Machine Learning}, 2016.

\bibitem{selvaraju2017grad}
Ramprasaath~R Selvaraju, Michael Cogswell, Abhishek Das, Ramakrishna Vedantam,
  Devi Parikh, and Dhruv Batra.
\newblock Grad-cam: Visual explanations from deep networks via gradient-based
  localization.
\newblock In {\em Proceedings of the IEEE International Conference on Computer
  Vision}, 2017.

\bibitem{snell2017prototypical}
Jake Snell, Kevin Swersky, and Richard Zemel.
\newblock Prototypical networks for few-shot learning.
\newblock In {\em Advances in Neural Information Processing Systems}, 2017.

\bibitem{sun2019meta}
Qianru Sun, Yaoyao Liu, Tat-Seng Chua, and Bernt Schiele.
\newblock Meta-transfer learning for few-shot learning.
\newblock In {\em Proceedings of the IEEE Conference on Computer Vision and
  Pattern Recognition}, 2019.

\bibitem{raphael2019robustly}
Raphael Suter, Đorđe Miladinović, Bernhard Sch\"{o}lkopf, and Stefan Bauer.
\newblock Robustly disentangled causal mechanisms: Validating deep
  representations for interventional robustness.
\newblock In {\em International Conference on Machine Learning}, 2019.

\bibitem{kaihua2020long}
Kaihua Tang, Jianqiang Huang, and Hanwang Zhang.
\newblock Long-tailed classification by keeping the good and removing the bad
  momentum causal effect.
\newblock In {\em Advances in Neural Information Processing Systems}, 2020.

\bibitem{tang2020unbiased}
Kaihua Tang, Yulei Niu, Jianqiang Huang, Jiaxin Shi, and Hanwang Zhang.
\newblock Unbiased scene graph generation from biased training.
\newblock In {\em Proceedings of the IEEE Conference on Computer Vision and
  Pattern Recognition}, 2020.

\bibitem{tanti2019transfer}
Marc Tanti, Albert Gatt, and Kenneth~P Camilleri.
\newblock Transfer learning from language models to image caption generators:
  Better models may not transfer better.
\newblock {\em arXiv preprint arXiv:1901.01216}, 2019.

\bibitem{tenenbaum2000global}
Joshua~B Tenenbaum, Vin De~Silva, and John~C Langford.
\newblock A global geometric framework for nonlinear dimensionality reduction.
\newblock {\em Science}, 2000.

\bibitem{turk1991face}
Matthew Turk and Alex Pentland.
\newblock Face recognition using eigenfaces.
\newblock In {\em Proceedings. 1991 IEEE Computer Society Conference on
  Computer Vision and Pattern Recognition}, 1991.

\bibitem{vaswani2017attention}
Ashish Vaswani, Noam Shazeer, Niki Parmar, Jakob Uszkoreit, Llion Jones,
  Aidan~N Gomez, {\L}ukasz Kaiser, and Illia Polosukhin.
\newblock Attention is all you need.
\newblock In {\em Advances in Neural Information Processing Systems}, 2017.

\bibitem{vinyals2016matching}
Oriol Vinyals, Charles Blundell, Timothy Lillicrap, Daan Wierstra, et~al.
\newblock Matching networks for one shot learning.
\newblock In {\em Advances in Neural Information Processing Systems}, 2016.

\bibitem{tan2019visual}
Tan Wang, Jianqiang Huang, Hanwang Zhang, and Qianru Sun.
\newblock Visual commonsense r-cnn.
\newblock In {\em Proceedings of the IEEE Conference on Computer Vision and
  Pattern Recognition}, 2019.

\bibitem{wang2019simpleshot}
Yan Wang, Wei-Lun Chao, Kilian~Q Weinberger, and Laurens van~der Maaten.
\newblock Simpleshot: Revisiting nearest-neighbor classification for few-shot
  learning.
\newblock {\em arXiv preprint arXiv:1911.04623}, 2019.

\bibitem{wang2019few}
Yaqing Wang and Quanming Yao.
\newblock Few-shot learning: A survey.
\newblock {\em arXiv preprint arXiv:1904.05046}, 2019.

\bibitem{WelinderEtal2010}
P.~Welinder, S.~Branson, T.~Mita, C.~Wah, F.~Schroff, S.~Belongie, and
  P.~Perona.
\newblock {Caltech-UCSD Birds 200}.
\newblock Technical report, California Institute of Technology, 2010.

\bibitem{xu2015show}
Kelvin Xu, Jimmy Ba, Ryan Kiros, Kyunghyun Cho, Aaron Courville, Ruslan
  Salakhudinov, Rich Zemel, and Yoshua Bengio.
\newblock Show, attend and tell: Neural image caption generation with visual
  attention.
\newblock In {\em International Conference on Machine Learning}, 2015.

\bibitem{xu2020deconfounded}
Xu~Yang, Hanwang Zhang, and Jianfei Cai.
\newblock Deconfounded image captioning: A causal retrospect.
\newblock {\em arXiv preprint arXiv:2003.03923}, 2020.

\bibitem{ye2020fewshot}
Han-Jia Ye, Hexiang Hu, De-Chuan Zhan, and Fei Sha.
\newblock Few-shot learning via embedding adaptation with set-to-set functions.
\newblock In {\em Proceedings of the IEEE Conference on Computer Vision and
  Pattern Recognition}, 2020.

\bibitem{yosinski2014transferable}
Jason Yosinski, Jeff Clune, Yoshua Bengio, and Hod Lipson.
\newblock How transferable are features in deep neural networks?
\newblock In {\em Advances in Neural Information Processing Systems}, 2014.

\bibitem{zagoruyko2016wide}
Sergey Zagoruyko and Nikos Komodakis.
\newblock Wide residual networks.
\newblock In {\em British Machine Vision Conference}, 2016.

\bibitem{zeiler2014visualizing}
Matthew~D Zeiler and Rob Fergus.
\newblock Visualizing and understanding convolutional networks.
\newblock In {\em European Conference on Computer Vision}, 2014.

\bibitem{zhang2020deepemd}
Chi Zhang, Yujun Cai, Guosheng Lin, and Chunhua Shen.
\newblock Deepemd: Few-shot image classification with differentiable earth
  mover's distance and structured classifiers.
\newblock 2020.

\bibitem{dong2020causal}
Dong Zhang, Hanwang Zhang, Jinhui Tang, Xian sheng Hua, and Qianru Sun.
\newblock Causal intervention for weakly-supervised semantic segmentation.
\newblock In {\em Advances in Neural Information Processing Systems}, 2020.

\bibitem{zhang2019few}
Hongguang Zhang, Jing Zhang, and Piotr Koniusz.
\newblock Few-shot learning via saliency-guided hallucination of samples.
\newblock In {\em Proceedings of the IEEE Conference on Computer Vision and
  Pattern Recognition}, 2019.

\bibitem{zhang2018metagan}
Ruixiang Zhang, Tong Che, Zoubin Ghahramani, Yoshua Bengio, and Yangqiu Song.
\newblock Metagan: An adversarial approach to few-shot learning.
\newblock In {\em Advances in Neural Information Processing Systems}, 2018.

\bibitem{zhang2020impact}
Youshan Zhang and Brian~D Davison.
\newblock Impact of imagenet model selection on domain adaptation.
\newblock In {\em Proceedings of the IEEE Winter Conference on Applications of
  Computer Vision Workshops}, 2020.

\bibitem{zhou2016learning}
Bolei Zhou, Aditya Khosla, Agata Lapedriza, Aude Oliva, and Antonio Torralba.
\newblock Learning deep features for discriminative localization.
\newblock In {\em Proceedings of the IEEE Conference on Computer Vision and
  Pattern Recognition}, 2016.

\end{thebibliography}
\newpage

\renewcommand{\thesection}{A.\arabic{section}}
\renewcommand*{\thesubsection}{A.\arabic{section}.\arabic{subsection}}
\renewcommand{\thetable}{A\arabic{table}}
\renewcommand{\thefigure}{A\arabic{figure}}
\renewcommand{\theequation}{A\arabic{equation}}

\setcounter{section}{0}
\setcounter{figure}{0}
\setcounter{table}{0}

\title{Supplementary Material for Interventional Few-Shot Learning}

\maketitle

This supplementary material is organized as follows:
\begin{itemize}
    \item Section~\ref{sec:a1} details our analysis in Section~\ref{sec:2.3} by showing many-shot learning converges to true causal effect through instrumental variable (IV);
    \item Section~\ref{sec:a2} gives the derivation for the backdoor adjustment formula in Eq.~\eqref{eq:backdoor};
    \item Section~\ref{sec:a3} presents the detailed derivation for the NWGM approximation used in Eq.~\eqref{eq:classwise_adjustment} and~\eqref{eq:combined_adjustment};
    \item Section~\ref{sec:a4} includes the algorithms for adding IFSL to fine-tuning and meta-learning;
    \item Section~\ref{sec:a5} shows the implementation details for pre-training (Section~\ref{sec:a5.1}), fine-tuning (Section~\ref{sec:a5.2}) and meta-learning (Section~\ref{sec:a5.3});
    \item Section~\ref{sec:a6} includes additional experimental results on Conventional Acc (Section~\ref{sec:a6.1}), Hardness-Specific Acc (Section~\ref{sec:a6.2}), CAM-Acc (Section~\ref{sec:a6.3}) and cross-domain evaluation (Section~\ref{sec:a6.4}).
\end{itemize}

\section{Instrumental Variable}
\label{sec:a1}
In this section, we will show that in our causal graph for many-shot learning, the sampling ID $I$ is essentially an instrumental variable for $X \rightarrow Y$ that achieves $P(Y|I)\approx P(Y|do(X))$. Before introducing instrumental variable, we first formally define \emph{d-separation}~\cite{pearl2016causal}, which gives a criterion to study the dependencies between nodes (data variables) in any structural causal model.

\noindent\textbf{d-separation}. A set of nodes $Z$ blocks a path $p$ if and only if 1) $p$ contains a \emph{chain} $A \rightarrow B \rightarrow C$ or a \emph{fork} $A \leftarrow B \rightarrow C$ and the middle node $B$ is in $Z$; 2) $p$ contains a \emph{collider} $A \rightarrow B \leftarrow C$ such that the middle node $B$ and its descendants are not in $Z$. If conditioning on $Z$ blocks every path between $X$ and $Y$, we say $X$ and $Y$ are \emph{d-separated} conditional on $Z$, \ie, $X$ and $Y$ are independent given $Z$ ($X \independent Y | Z$).

\noindent\textbf{Instrumental Variable}. For a structual causal model $\mathcal{G}$, a variable Z is an \emph{instrumental variable} (IV) to $X \rightarrow Y$ by satisfying the graphical criteria~\cite{pearl2009causality}: 1) $(Z \independent Y)_{\mathcal{G}_{\overline{X}}}~$; 2) $(Z \not\independent X)_{\mathcal{G}}~$, where $\mathcal{G}_{\overline{X}}$ is the manipulated graph where all incoming arrows to node $X$ are deleted. For the SCM of many-shot learning in Figure 4(a), it is easy to see that $I$ satisfies both criteria and therefore it is an IV for $X \rightarrow Y$. However, in the few-shot SCM in Figure 4(b), the paths $I\leftarrow X \leftarrow D \rightarrow C \rightarrow Y$ and $I\leftarrow X \rightarrow C \rightarrow Y$ are not blocked in $\mathcal{G}_{\overline{X}}$, which means the first criterion is not met $(I \not\independent Y)_{\mathcal{G}_{\overline{X}}}~$ and $I$ is not an instrumental variable in the few-shot learning case.

Instrumental variable can help find the true causal effect even in the presence of confounder. This is due to the collider junction that makes the IV and confounder independent ($I \independent D$ in Figure 4(a)).
To see this, we will first consider a simplified case of Figure 4(a) where each causal link represents a linear relationship and we aim to find the true causal effect from $X\rightarrow Y$ through linear regression. 
Without loss of generality, let $I, X, Y$ take the value of real number. Denote $r_{IX}, r_{XY},$ and $r_{IY}$ as the slope of regression line between $I$ and $X$, $X$ and $Y$, $I$ and $Y$ respectively. Notice that $r_{XY}$ is spurious as it is contaminated by the backdoor path $X \leftarrow D \rightarrow C \rightarrow Y$. However, since the path $I\rightarrow X\leftarrow D \rightarrow C \rightarrow Y$ is blocked due to collider at $X$, $r_{IY}$ is free from confounding bias. Therefore $r_{IY}/r_{IX}$ gives the true causal effect from $X \rightarrow Y$. Similarly, in the classification case of many-shot learning, a classifier is trained to maximize the conditional probability on the IV $P(Y|I)$. As the ID-sample matching $I \rightarrow X$ is deterministic, the classifier eventually learns to predict based on the true causal relationship $X \rightarrow Y$. Yet in the complex case of image classification, it is unreasonable to assume linear relationships between variables. In the nonlinear case, it is shown in \cite{balke1997bounds} that observations on IV provide a bound for the true causal effect. This means that learning based on $P(Y|I)$ provides an approximation to the true causal effect, \ie $P(Y|I)\approx P(Y|do(X))$.

\section{Derivation of Backdoor Adjustment for the Proposed Causal Graph}
\label{sec:a2}
We will show the derivation of the backdoor adjustment for the causal graph in Figure 3(a) using the three rules of \emph{do}-calculus~\cite{pearl1995causal}.

For a causal directed acyclic graph $\mathcal{G}$, let $X, Y, Z$ and $W$ be arbitrary disjoint sets of nodes. We use $\mathcal{G}_{\overline X}$ to denote the manipulated graph where all incoming arrows to node $X$ are deleted. Similarly $\mathcal{G}_{\underline X}$ represents the graph where outgoing arrows from node $X$ are deleted. We use lower case $x,y,z$ and $w$ for specific values taken by each set of nodes: $X=x, Y=y, Z=z$ and $W=w$. For any interventional distribution compatible with $\mathcal{G}$, we have the following three rules:

\noindent\textbf{Rule 1} Insertion/deletion of observations:
\begin{equation}
    P(y|do(x),z,w)=P(y|do(x),w), \mathrm{if} (Y \independent Z | X,W)_{\mathcal{G}_{\overline X}}
\end{equation}

\noindent\textbf{Rule 2} Action/observation exchange:
\begin{equation}
    P(y|do(x),do(z),w)=P(y|do(x),z, w), \mathrm{if} (Y \independent Z | X,W)_{\mathcal{G}_{\overline X \underline Z}}
\end{equation}

\noindent\textbf{Rule 3} Insertion/deletion of actions:
\begin{equation}
    P(y|do(x),do(z),w)=P(y|do(x),w), \mathrm{if} (Y \independent Z | X,W)_{\mathcal{G}_{\overline X \overline {Z(W)}}},
\end{equation}
where $Z(W)$ is the set of nodes in $Z$ that are not ancestors of any $W$-node in $\mathcal{G}_{\overline{X}}$.

In our causal graph, the desired interventional distribution $P(Y|do(X=\mathbf{x}))$ can be derived by:
\begin{align}
    P(Y|do(\mathbf{x})) &= \sum_{d} P(Y|do(X=\mathbf{x}),D=d) P(D=d|do(X=\mathbf{x})) \label{bd1}\\
                        &= \sum_{d} P(Y|do(X=\mathbf{x}),D=d) P(D=d) \label{bd2}\\
                        &= \sum_{d} P(Y|X=\mathbf{x},D=d) P(D=d) \label{bd3}\\
                        &= \sum_{d} \sum_{c} P(Y|X=\mathbf{x},D=d, C=c) P(C=c|X=\mathbf{x},D=d) P(D=d) \label{bd4}\\
                        &= \sum_{d} P(Y|X=\mathbf{x}, D=d, C=g(\mathbf{x},d)) P(D=d) \label{bd5},
\end{align}
where Eq.~\eqref{bd1} and Eq.~\eqref{bd4} follow the law of total probability; Eq.~\eqref{bd2} uses Rule 3 given $D \independent X$ in $\mathcal{G}_{\overline{X}}$; Eq.~\eqref{bd3} uses Rule 2 to change the intervention term to observation as $(Y \independent X | D)$ in $\mathcal{G}_{\underline{X}}$; Eq.~\eqref{bd5} is because in our causal graph, $C$ takes a deterministic value given by function $g(\mathbf{x}, d)$. This reduces summation over all values of $C$ in Eq.~\eqref{bd4} to a single probability measure in Eq.~\eqref{bd5}.

\section{Derivation of NWGM Approximation}
\label{sec:a3}
We will show the derivation of NWGM approximation used in Eq.~\eqref{eq:classwise_adjustment} and~\eqref{eq:combined_adjustment}. In a $K$-way FSL problem, let $f(\cdot)$ be a classifier function that calculates logits for $K$ classes and $\sigma$ be the softmax function over $K$ classes. The approximation effectively moves the outer expectation inside the classifier function: $\E \left[\sigma(f(\cdot))\right] \approx \sigma(f(\E[ \cdot]))$.

We will first show the derivation for moving the expectation inside softmax function, \ie, $\E [\sigma\left(f(\cdot)\right)] \approx \sigma \left(\E [f(\cdot)]\right)$. Without loss of generality, the backdoor adjustment formula in Eq. (3) and Eq. (4) can be written as:
\begin{equation}
    P(Y=y|do(X=\mathbf{x})) = \sum_{d \in D} \sigma(f_y(\mathbf{x} \oplus \mathbf{c})) P(d),
    \label{do_expanded}
\end{equation}
where $D$ represents the set of stratifications, $f_y$ is the classifier logit for class $y$, $\mathbf{c}=g(\mathbf{x},d)$ is the feature concatenated to $\mathbf{x}$ in Eq. (3) and (4) and $P(d)$ is the prior for each stratificaction.

It is shown in \cite{baldi2014dropout} that Eq.~\eqref{do_expanded} can be approximated by the Normalized Weighted Geometric Mean (NWGM) as:
\begin{align}
    \sum_{d \in D} \sigma(f_y(\mathbf{x} \oplus \mathbf{c})) P(d) &\approx NWGM_{d\in D}(\sigma(f_y(\mathbf{x} \oplus \mathbf{c}))) \label{eq:nwgmapprox}\\
                    &= \frac{\prod_d [exp(f_y(\mathbf{x} \oplus \mathbf{c}))]^{P(d)}}{\sum_{i=1}^{K} \prod_d [exp(f_i(\mathbf{x} \oplus \mathbf{c}))]^{P(d)}} \label{eq:nwgmdefinition}\\
                    &= \frac{exp(\sum_d f_y(\mathbf{x} \oplus \mathbf{c})P(d))}{\sum_{i=1}^K exp(\sum_d f_i(\mathbf{x} \oplus \mathbf{c})P(d))} \label{eq:expab}\\
                    &= \sigma\left(\E_d [ f_y(\mathbf{x} \oplus \mathbf{c}) ]\right),
\end{align}
where Eq.~\eqref{eq:nwgmapprox} follows~\cite{baldi2014dropout}, Eq.~\eqref{eq:nwgmdefinition} follows the definition of NWGM, Eq.~\eqref{eq:expab} is because $exp(a)^b=exp(ab)$.

Next we will show the derivation for linear, cosine and $k$-NN classifier to further move the expectation inside the classifier function, \ie, $\sigma(\E [f(\cdot)]) = \sigma(f(\E [\cdot]))$.

For the linear classifier, $f(\mathbf{x} \oplus \mathbf{c}) = \mathbf{W}_1 \mathbf{x} + \mathbf{W}_2 \mathbf{c}$, where $\mathbf{W}_1, \mathbf{W}_2 \in \mathbb{R}^{K\times N}$ denote the learnable weight, $N$ is the feature dimension, which is the same for $\mathbf{x}$ and $\mathbf{c}$ in Eq.~\eqref{eq:classwise_adjustment} and~\eqref{eq:combined_adjustment}. The bias term is dropped as it does not impact our analysis. Now the expectation can be further moved inside the classifier function through:
\begin{align}
    \sum_d f(\mathbf{x} \oplus \mathbf{c})) P(d) &= \sum_d (\mathbf{W}_1 \mathbf{x} + \mathbf{W}_2 \mathbf{c}) P(d) \\
                                                &= \mathbf{W}_1 \mathbf{x} + \sum_d \mathbf{W}_2 \mathbf{c} P(d) \label{eq:edx}\\
                                                &= f(\mathbf{x} \oplus \sum_d \mathbf{c} P(d)),
\end{align}
where Eq.~\eqref{eq:edx} is because the feature vector $\mathbf{x}$ is the same for all $d$ and $\E_d [\mathbf{x}] = \mathbf{x}$.

For the cosine classifier, $f(\mathbf{x} \oplus \mathbf{c}) = (\mathbf{W}_1 \mathbf{x} + \mathbf{W}_2 \mathbf{c}) / \norm{\mathbf{x} \oplus \mathbf{c}} \norm{\mathbf{W}}$, where $\mathbf{W} \in \mathbb{R}^{K \times 2N}$ is the concatenation of $\mathbf{W}_1$ and $\mathbf{W}_2$. In the special case where $\mathbf{x}$ and $\mathbf{c}$ are unit vector, $\norm{\mathbf{x} \oplus \mathbf{c}}$ is $\sqrt{2}$ and the cosine classifier function reduces to a linear combination of terms involving only $\mathbf{x}$ and only $\mathbf{c}$. From there, the analysis for linear classifier follows and we have $\sigma(\E f(\cdot)) = \sigma(f(\E \cdot))$ for cosine classifier. In the general case where $\mathbf{x}$ and $\mathbf{c}$ are not unit vector, moving the expectation inside cosine classifier function is an approximation $\sigma(\E [f(\cdot)]) \approx \sigma(f(\E [\cdot]))$.

For the $k$-NN classifier, our implementation calculates class centroids using the mean feature of the $K$ support sets and then uses the nearest centroid for prediction ($1$-NN). Specifically, let $\mathbf{x}$ be a feature vector and $\mathbf{x'}$ be the $i$th class centroid, $i\in \{1,\ldots,K\}$. The logit for class $i$ is given by $f_i(\mathbf{x})=-\norm{\mathbf{x}-\mathbf{x'}}^2$. It is shown in~\cite{snell2017prototypical} that $k$-NN classifier that uses squared Euclidean distance to generate logits is equivalent to a linear classifier with a particular parameterization. Therefore, our analysis on linear classifier follows for $k$-NN.

In summary, the derivation of $\E [\sigma(f(\cdot))] \approx \sigma(f(\E[ \cdot]))$ is a two-stage process where we first move the expectation inside the softmax function and then further move it inside the classifier function.

\section{Algorithms for Fine-tuning and Meta-Learning with IFSL}
\label{sec:a4}

In this section, we will briefly revisit the settings of fine-tuning and meta-learning and introduce how to integrate IFSL into them.

In fine-tuning, the goal is to train a classifier $\theta$ conditioned on the current support set $\mathcal{S}=\{(\mathbf{x}_i, y_i)\}_{i=1}^{n_s}$, where $\mathbf{x}_i$ is the feature generated by $\Omega$ for $i$th sample, $y_i$ is the ground-truth label for $i$th sample and $n_s$ is the support set size. This is achieved by first predicting the support label $\hat{y}$ using the classifier $P(y|\mathbf{x};\theta)$. Then with the predicted label $\hat{y}$ and ground-truth label $y$, one can calculate a loss $\mathcal{L}(\hat{y}, y)$ (usually cross-entropy loss) to update the classifier parameter, \eg through stochastic gradient descent. Adding IFSL to fine-tuning is simple: 1) Pick an adjustment strategy introduced in Section 3. Each implementation defines the set of pre-trained knowledge stratifications $D$, function form of $g(X,D)$, function form of $P(Y|X,D,C)$ and the prior $P(D)$; 2) The classifier prediction is now based on $P(Y|do(X);\theta)$. The process of fine-tuning with IFSL is summarized in Algorithm~\ref{alg:1}. Note that for the non-parametric $k$-NN classifier, the fine-tuning process is not applicable. When adding IFSL to $k$-NN, each sample is represented by the \emph{adjusted feature} instead of original feature $\mathbf{x}$. Please refer to the classifier inputs in Eq.~\eqref{eq:featurewise_adjustment},~\eqref{eq:classwise_adjustment} and~\eqref{eq:combined_adjustment} for the exact form of adjusted feature.

In meta-learning, the goal is to learn the additional ``learning behavior'' parameterized by $\phi$ using training episodes $\{(\mathcal{S}_i, \mathcal{Q}_i)\}$ sampled from training dataset $\mathcal{D}$. The classifier in meta-learning makes predictions by additionally conditioning on the learning behavior, written as $P_{\phi}(y|\mathbf{x};\theta)$. Within each episode, $\theta$ is first fine-tuned on the support set $\mathcal{S}_i$. Then the fine-tuned model is tested on the query set $\mathcal{Q}_i$ to obtain the loss $\mathcal{L}_\phi(\mathcal{S}_i,\mathcal{Q}_i)$ (\eg using cross-entropy loss). Finally the loss is used to update $\phi$ using an optimizer. It is also easy to integrate IFSL into meta-learning by only changing the classifier from $P_{\phi}(y|\mathbf{x};\theta)$ to $P_{\phi}(y|do(\mathbf{x});\theta)$. The flow of meta-learning with IFSL is presented in Algorithm~\ref{alg:2}. Firstly notice that the initialization of $\theta$ in each task may depend on $\phi$ or $\mathcal{S}_i$. For example, in MAML~\cite{finn2017model} $\phi$ essentially defines an initialization of model parameters, and in LEO~\cite{rusu2018meta} the initial classifier parameter is generated conditioned on $\phi$ and $\mathcal{S}_i$. Secondly, although the fine-tuning of $\theta$ largely follows Algorithm~\ref{alg:1}, some meta-learning methods additionally utilize meta-knowledge $\phi$. For example, in SIB the gradients for updating $\theta$ are predicted by $\phi$ using unlabelled query features instead of calculated from $\mathcal{L}(\hat{y}, y)$ as in Algorithm~\ref{alg:1}.

\begin{figure}[h!]
    \begin{minipage}{.48\linewidth}
        \begin{algorithm}[H]
        \label{alg:1}
        \SetAlgoLined
        \SetKwInOut{KwIn}{Input}
        \SetKwInOut{KwOut}{Output}
        \KwIn{$D$, Support set $\mathcal{S}=\{(\mathbf{x}_i,y_i)\}_{i=1}^{n_s}$}
        \KwOut{Fine-tuned classifier parameters $\theta$}
         Initialize $\theta$\;
         \While{not converged}{
          \For{$i=1,\ldots,n_s$} {
            \For{$d \in D$} {
                Calculate $c=g(\mathbf{x}, d)$\;
                Obtain $P(Y|\mathbf{x}_i, c, d;\theta), P(d)$
            }
            Prediction $\hat{y}_i = P(y|do(\mathbf{x});\theta)$\;
            Update $\theta$ using $\mathcal{L}(\hat{y}_i,y_i)$
          }
         }
         \KwRet{$\theta$}
         \caption{Fine-tuning + IFSL}
        \end{algorithm}
    \end{minipage}
    \hfill
    \begin{minipage}{.48\linewidth}
        \begin{algorithm}[H]
        \label{alg:2}
        \SetAlgoLined
        \SetKwInOut{KwIn}{Input}
        \SetKwInOut{KwOut}{Output}
        \KwIn{$D$, training dataset $\mathcal{D}$}
        \KwOut{Optimized meta-parameters $\phi$}
         Initialize $\phi$\;
         \While{not converged}{
            Sample $(\mathcal{S}_i,\mathcal{Q}_i)$ from $\mathcal{D}$ \;
            Initialize classifier $\theta$ with $\phi, \mathcal{S}_i$\;
            Fine-tune $\theta$ using \textbf{Algorithm 1} conditioned on $\phi$ \;
            Predict query based on $P_{\phi}(y|do(\mathbf{x});\theta)$\;
            Update $\phi$ using $\mathcal{L}_{\phi}(\mathcal{S}_i,\mathcal{Q}_i;\theta)$
         }
         \KwRet{$\phi$}
         \caption{Meta-Learning + IFSL}
        \end{algorithm}
    \end{minipage}
\end{figure}

\section{Implementation Details}
\label{sec:a5}

\subsection{Pre-training}
\label{sec:a5.1}

Prior to fine-tuning or meta-learning, we pre-trained a deep neural network (DNN) as feature extractor on the train split of a dataset. We use ResNet-10\cite{he2016deep} or WRN-28-10\cite{zagoruyko2016wide} as feature extractor backbone. This section will present the architecture and exact training procedure for our backbones.

\begin{wrapfigure}{r}{0.4\textwidth}
\centering
\captionsetup{font=footnotesize,labelfont=footnotesize}
\begin{subfigure}[t]{.2\textwidth}
  \centering
  \captionsetup{width=\linewidth}
  \includegraphics[width=0.93\linewidth]{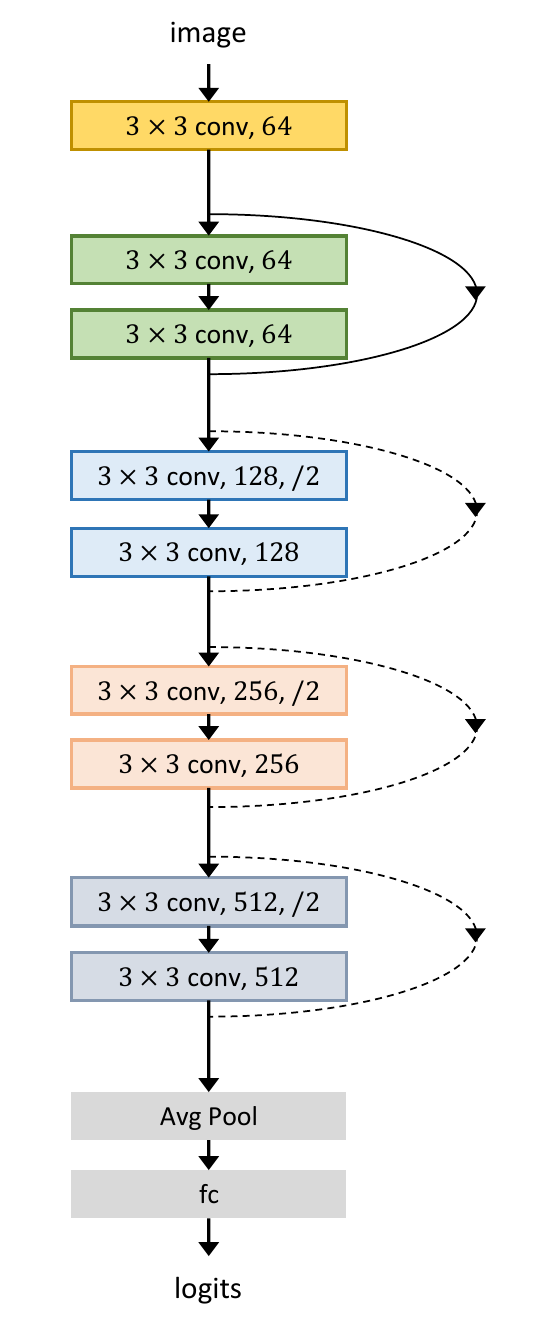}
  \caption{}
  \label{fig:A1a}
\end{subfigure}%
\begin{subfigure}[t]{.2\textwidth}
  \centering
  \captionsetup{width=\linewidth}
  \includegraphics[width=0.95\linewidth]{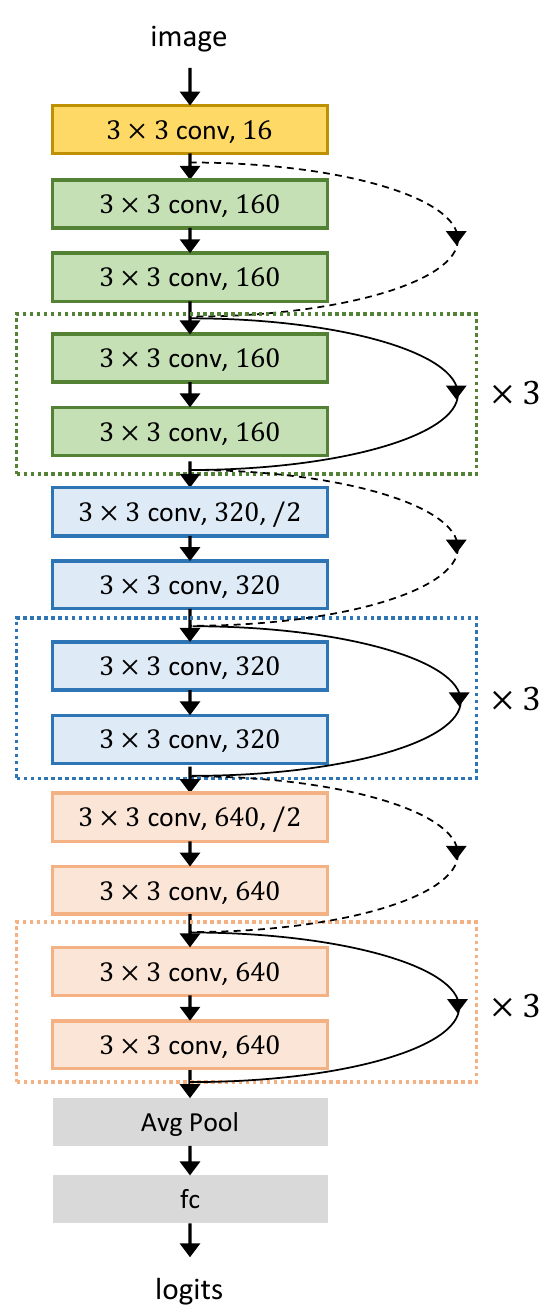}
  \caption{}
  \label{fig:A1b}
\end{subfigure}
\caption{The architecture of our backbones: (a) ResNet-10~\cite{he2016deep}; (b) WRN-28-10~\cite{zagoruyko2016wide}.}
\label{fig:A1}
\vspace{-7mm}
\end{wrapfigure}

\noindent\textbf{Network Architecture.} The architecture of our ResNet-10 and WRN-28-10 backbone is shown in Figure~\ref{fig:A1}. Specifically, each convolutional layer is described as ``$n\times n$ conv, $p$'', where $n$ is the kernel size and $p$ is the number of output channels. Convolutional layers with ``$/2$'' have a stride of 2 and are used to perform downsampling. The solid curved lines represent identity shortcuts, and the dotted lines are projection shortcuts implemented by $1\times 1$ convolutions. The batch normalization and ReLU layers are omitted in Figure~\ref{fig:A1} to highlight the key structure of the two backbones.

\noindent\textbf{Pre-training Procedure.} The networks are trained from scratch with stochastic gradient descent in a fully-supervised manner, \ie, minimizing cross-entropy loss on the train split of a dataset. Specifically the training is conducted on 90 epochs with early stopping using validation accuracy. We used batch size of 256 and image size of $84 \times 84$. For data augmentation, a random patch is sampled from an image, resized to $84 \times 84$ and randomly flipped along horizontal axis before used for training. The initial learning rate is set to 0.1 and it is scaled down by factor of 10 every 30 epochs.

\subsection{Fine-Tuning}
\label{sec:a5.2}
We consider linear, cosine and $k$-NN classifier for our fine-tuning experiments. In a $K$-way FSL problem, the detailed implementations for the classifier function $f(\mathbf{x})$ are:

\noindent\textbf{Linear}. $f(\mathbf{x})=\mathbf{W}\mathbf{x} + \mathbf{b}$, where $\mathbf{x}$ is the input feature, $\mathbf{W}\in \mathbb{R}^{K\times N}$ is the learnable weight parameter, $N$ is the feature dimension and $\mathbf{b} \in \mathbb{R}^{K}$ is the learnable bias parameter.

\noindent\textbf{Cosine}. $f(\mathbf{x})=\mathbf{W}\mathbf{x}/\norm{\mathbf{W}}\norm{\mathbf{x}}$, where $\mathbf{W}\in \mathbb{R}^{K\times N}$ is the learnable weight parameter. We implemented cosine classifier without using the bias term.

\noindent\textbf{$k$-NN}. Our implementation of $k$-NN is similar to~\cite{snell2017prototypical,wang2019simpleshot}. For each of the $K$ classes, we first calculated the average support set feature (centroid) denoted as $\mathbf{x}_i, i\in \{1,\ldots,K\}$. The classifier output for class $i$ is then given by $f_i(\mathbf{x})=-\norm{\mathbf{x}-\mathbf{x}_i}^2$. Notice that the prediction given by this classifier will be the nearest centroid.

We froze the backbone and used the average pooling layer output of $\Omega$ to learn the classifier. The output logits from classifier functions are normalized using softmax to generate probability output $P(y|\mathbf{x})$. For linear and cosine classifier, we followed~\cite{chen2019closer} and trained the classifier for 100 iteration with a batch size of 4. For fine-tuning baseline, we set the learning rate as $1\times 10^{-2}$ and weight decay as $1\times 10^{-3}$. For IFSL, we set the learning rate as $5\times 10^{-3}$ and weight decay as $1\times 10^{-3}$. $k$-NN classifier is non-parametric and can be initialized directly from support set.

\subsection{Meta-Learning}
\label{sec:a5.3}

\noindent\textbf{MAML.}
MAML~\cite{finn2017model} aims to learn an initialization of network parameters such that it can be fine-tuned within a few steps to solve a variety of few-shot classification tasks. 
When using pre-trained network with MAML, it has been shown that learning initialization of the backbone can lead to unsatisfactory performance~\cite{chen2019closer,sun2019meta}. Therefore in our experiment, we froze the backbone and appended a 2-layer MLP with ReLU activation in between the hidden layers and a linear classifier after the average pooling layer of $\Omega$. The hidden dimension of the layers in MLP is the same as output dimension of $\Omega$ (512 for ResNet-10 and 640 for WRN-28-10). The initialization of MLP and the linear classifier is meta-learnt using MAML. For hyper-parameters, we set the inner loop learning rate $\alpha=0.01$, the outer loop learning rate $\beta=0.01$ and the number of adaptation steps as $20$. For IFSL, we adopted the same hyper-parameter setting and set $n$=8 for feature-wise and combined adjustment.
Implementation-wise, we adopted the released code\footnote{https://github.com/wyharveychen/CloserLookFewShot} from~\cite{chen2019closer} and performed experiments on MAML without using first-order approximation. Following the implementation in~\cite{chen2019closer}, the model was trained on 10,000 randomly sampled tasks with model selection using validation accuracy. We used 2,000 randomly sampled tasks for validation and testing.

\noindent\textbf{MTL.} MTL~\cite{sun2019meta} learns scaling and shifting parameters at each convolutional layer of the backbone. We used the MTL implementation released by the author\footnote{https://github.com/yaoyao-liu/meta-transfer-learning} which adopts linear classifier. We integrated our ResNet-10 and WRN-28-10 backbones into the released code. The learning rate for scaling and shifting weights $\phi_{SS}$ and initial classifier parameters was set to $1\times 10^{-4}$ uniformly. We set the inner loop learning rate for classifier as $1\times 10^{-2}$ and the inner loop update step as 100. For IFSL, we adopted the same hyper-parameter setting and set $n$=8 for feature-wise and combined adjustment.
We trained the MTL model on 10,000 randomly sampled tasks with model selection using validation accuracy and used 2,000 randomly sampled tasks for validation and testing.
We used 3 RTX 2080 Ti for MTL experiments on WRN-28-10 backbone.

\noindent\textbf{LEO.} LEO~\cite{rusu2018meta} learns to generate classifier parameters conditioned on support set and the generated parameters are further fine-tuned within each FSL task. Our experiments were conducted on the released code of LEO\footnote{https://github.com/deepmind/leo} using linear classifier. Following author's implementation, we saved the center cropped features from our pre-trained backbones and used the saved features to train LEO. For baseline, we used the hyper-parameter settings released by the author. For IFSL, we set $n$=8 for feature-wise and combined adjustment and halved the outer loop learning rate compared to baseline. The model was trained up to 100,000 randomly sampled tasks from training split with early stopping using validation accuracy. We used 2,000 randomly sampled tasks for validation and testing.

\noindent\textbf{Matching Net.} Matching Net (MN)~\cite{vinyals2016matching} is a metric-based method that learns a distance kernel function for $k$-NN. We used the Matching Net implementation in~\cite{chen2019closer}. The implementation follows the setup in~\cite{vinyals2016matching} and uses LSTM-based fully conditional embedding. We set the learning rate as 0.01 uniformly. For IFSL, we used $n$=16 for feature-wise and combined adjustment. The model was trained using 10,000 randomly sampled tasks with model selection using validation accuracy. We used 2,000 randomly sampled tasks for validation and testing.

\noindent\textbf{SIB.} SIB~\cite{hu1empirical} initializes classifier from support set and generates gradients conditioned on unlabelled query set features to update classifier parameters. We followed the SIB implementation released by the author\footnote{https://github.com/hushell/sib\_meta\_learn} which uses cosine classifier.
In the transductive setting, the query set size is set to 15.
In the inductive setting, we used only 1 query sample randomly selected from the $K$ classes in each episode.
In terms of hyper-parameter settings, we took 3 synthetic gradient steps ($K=3$) for all our experiments.
For baseline, the learning rate for SIB network and classifier was set to $1\times 10 ^{-3}$ following author's implementation. For IFSL, we set the learning rate to $5\times 10^{-4}$ and used $n$=4 for feature-wise and combined adjustment.
In both transductive and inductive settings, we meta-trained SIB using 50,000 randomly sampled tasks with model selection using validation accuracy. We used 2,000 randomly sampled tasks for validation and testing.


\section{Additional Results}
\label{sec:a6}

In this section, we include additional results on 1) \textbf{Conventional Acc} in Table~\ref{tab:a1} supplementary to Table 1; 2) \textbf{Hardness-Specific Acc} in Figure~\ref{fig:a2} for \textit{mini}ImageNet and Figure~\ref{fig:a3} for \textit{tiered}ImageNet, supplementary to Figure 5; 3) \textbf{CAM-Acc} in Table~\ref{tab:a2} supplementary to Figure 6; 4) \textbf{Cross-Domain Evaluation} in Table~\ref{tab:a3} supplementary to Table 3.

\newpage
\subsection{Conventional Acc}
\label{sec:a6.1}
\begin{figure}[!htbp]
    \centering
    \fontsize{6}{7.2}\selectfont
    \renewcommand\arraystretch{1.2}
    \captionsetup{width=\linewidth}
    \captionof{table}{Supplementary to Table 1. Acc (\%) and 95\% confidence intervals averaged over 2000 5-way FSL tasks before and after applying three proposed implementations of adjustment. Specifically, ``\emph{ft}'' refers to feature-wise adjustment, ``\emph{cl}'' refers to class-wise adjustment and ``\emph{ft+cl}'' refers to combined adjustment.}
    \setlength\tabcolsep{4pt}
    \begin{tabular}{@{\hskip0pt}c | c c c c c c c c c c}
    \hhline{-|-|-|-|-|-|-|-|-|-|-|}
    \hhline{-|-|-|-|-|-|-|-|-|-|-|}
    \multicolumn{3}{c}{} & \multicolumn{4}{c}{\textbf{ResNet-10}} & \multicolumn{4}{c}{\textbf{WRN-28-10}} \\
    
    \multicolumn{3}{c}{} & \multicolumn{2}{c}{\textit{mini}ImageNet} & \multicolumn{2}{c}{\textit{tiered}ImageNet} & \multicolumn{2}{c}{\textit{mini}ImageNet} & \multicolumn{2}{c}{\textit{tiered}ImageNet} \\ \cmidrule(lr){4-5} \cmidrule(lr){6-7} \cmidrule(lr){8-9} \cmidrule(lr){10-11}
    
    \multicolumn{3}{c}{\multirow{-3}{*}[0.3em]{\textbf{Method}}} & $5$-shot & $1$-shot & $5$-shot & $1$-shot & $5$-shot & $1$-shot & $5$-shot & $1$-shot\\
    \hline
    
     & \multicolumn{1}{c}{\multirow{4}{*}{\shortstack{Linear}}} & & 76.38 $\pm$ 0.36 & 56.26 $\pm$ 0.47 & 81.01 $\pm$ 0.38 & 61.39 $\pm$ 0.47 & 79.79 $\pm$ 0.33 & 60.69 $\pm$ 0.45 & 85.37 $\pm$ 0.34 & 67.27 $\pm$ 0.49\\
     
     & \multicolumn{1}{c}{} & \textit{ft} & 76.84 $\pm$ 0.36 & 57.37 $\pm$ 0.43 & 81.45 $\pm$ 0.38 & 61.88 $\pm$ 0.47 & 80.22 $\pm$ 0.31 & 60.84 $\pm$ 0.45 & 85.70 $\pm$ 0.33 & 67.94 $\pm$ 0.48\\
     
     & \multicolumn{1}{c}{} & \textit{cl} & 77.23 $\pm$ 0.34 & 59.45 $\pm$ 0.45 & 81.33 $\pm$ 0.38 & 62.60 $\pm$ 0.48 & 80.27 $\pm$ 0.32 & 62.15 $\pm$ 0.44 & 85.54 $\pm$ 0.33 & 68.11 $\pm$ 0.48\\
     
    & \multicolumn{1}{c}{} & \textit{ft+cl} & 77.97 $\pm$ 0.34 & 60.13 $\pm$ 0.45 & 82.08 $\pm$ 0.37 & 64.29 $\pm$ 0.48 & 80.97 $\pm$ 0.31 & 64.12 $\pm$ 0.44 & 86.19 $\pm$ 0.34 & 69.96 $\pm$ 0.46\\ \hhline{~|-;-|-|-|-|-|-|-|-|-|}
    
    &\multicolumn{1}{c}{\multirow{4}{*}{\shortstack{Cosine}}} & & 76.68 $\pm$ 0.36 & 56.40 $\pm$ 0.46 & 81.13 $\pm$ 0.39 & 62.08 $\pm$ 0.47 & 79.72 $\pm$ 0.33 & 60.83 $\pm$ 0.46 & 85.41 $\pm$ 0.34 & 67.30 $\pm$ 0.50 \\
    
    & \multicolumn{1}{c}{} & \textit{ft} & 76.83 $\pm$ 0.35 & 56.86 $\pm$ 0.44 & 81.34 $\pm$ 0.37 & 62.45 $\pm$ 0.47 & 79.80 $\pm$ 0.32 & 61.25 $\pm$ 0.44 & 85.74 $\pm$ 0.33 & 67.86 $\pm$ 0.46\\
     
     & \multicolumn{1}{c}{} & \textit{cl} & 76.99 $\pm$ 0.35 & 57.65 $\pm$ 0.45 & 81.42 $\pm$ 0.38 & 63.37 $\pm$ 0.48 & 79.96 $\pm$ 0.32 & 62.04 $\pm$ 0.45 & 85.77 $\pm$ 0.33 & 68.45 $\pm$ 0.46\\
     
    & \multicolumn{1}{c}{} & \textit{ft+cl} & 77.63 $\pm$ 0.34 & 59.84 $\pm$ 0.46 & 81.75 $\pm$ 0.38 & 64.47 $\pm$ 0.48 & 80.74 $\pm$ 0.32 & 63.76 $\pm$ 0.45 & 86.13 $\pm$ 0.33 & 69.36 $\pm$ 0.47\\ \hhline{~|-;-|-|-|-|-|-|-|-|-|}
    
    &\multicolumn{1}{c}{\multirow{4}{*}{\shortstack{$k$-NN}}} & & 76.63 $\pm$ 0.36 & 55.92 $\pm$ 0.46 & 80.85 $\pm$ 0.39 & 61.16 $\pm$ 0.48 & 79.60 $\pm$ 0.32 & 60.34 $\pm$ 0.45 & 84.67 $\pm$ 0.34 & 67.25 $\pm$ 0.52 \\
    
    & \multicolumn{1}{c}{} & \textit{ft} & 77.98 $\pm$ 0.34 & 60.71 $\pm$ 0.44 & 81.95 $\pm$ 0.36 & 65.66 $\pm$ 0.48 & 81.17 $\pm$ 0.31 & 64.87 $\pm$ 0.44 & 85.76 $\pm$ 0.34 & 71.00 $\pm$ 0.47\\
     
     & \multicolumn{1}{c}{} & \textit{cl} & 78.36 $\pm$ 0.35 & 61.32 $\pm$ 0.45 & 81.93 $\pm$ 0.37 & 65.71 $\pm$ 0.48 & 80.61 $\pm$ 0.31 & 64.43 $\pm$ 0.45 & 85.90 $\pm$ 0.33 & 70.08 $\pm$ 0.48\\
     
    \multirow{-12}{*}{\STAB{\rotatebox[origin=c]{90}{\textbf{Fine-Tuning}}}} & \multicolumn{1}{c}{} & \textit{ft+cl} & 78.42 $\pm$ 0.34 & 62.31 $\pm$ 0.44 & 81.98 $\pm$ 0.38 & 65.71 $\pm$ 0.47 & 81.08 $\pm$ 0.32 & 64.98 $\pm$ 0.43 & 86.06 $\pm$ 0.32 & 70.94 $\pm$ 0.49\\
    
    \hline
    
    &\multicolumn{1}{c}{\multirow{4}{*}{\shortstack{MAML~\cite{finn2017model}}}} & & 70.85 $\pm$ 0.38 & 56.59 $\pm$ 0.48 & 74.02 $\pm$ 0.41 & 59.17 $\pm$ 0.52 & 73.92 $\pm$ 0.36 & 58.02 $\pm$ 0.47 & 77.20 $\pm$ 0.38 & 61.40 $\pm$ 0.54 \\
    
    & \multicolumn{1}{c}{} & \textit{ft} & 73.84 $\pm$ 0.37 & 57.63 $\pm$ 0.47 & 80.19 $\pm$ 0.40 & 60.03 $\pm$ 0.51 & 78.82 $\pm$ 0.36 & 58.55 $\pm$ 0.48 & 84.74 $\pm$ 0.37 & 66.74 $\pm$ 0.52\\
     
    & \multicolumn{1}{c}{} & \textit{cl} & 73.01 $\pm$ 0.36 & 56.69 $\pm$ 0.48 & 78.41 $\pm$ 0.40 & 61.16 $\pm$ 0.53 & 76.22 $\pm$ 0.35 & 58.32 $\pm$ 0.46 & 81.74 $\pm$ 0.38 & 63.61 $\pm$ 0.51\\
     
    & \multicolumn{1}{c}{} & \textit{ft+cl} & 76.37 $\pm$ 0.37 & 59.36 $\pm$ 0.48 & 81.04 $\pm$ 0.39 & 63.88 $\pm$ 0.50 & 79.25 $\pm$ 0.34 & 62.84 $\pm$ 0.46 & 85.10 $\pm$ 0.39 & 67.70 $\pm$ 0.53\\ \hhline{~|-;-|-|-|-|-|-|-|-|-|}
    
    &\multicolumn{1}{c}{\multirow{4}{*}{\shortstack{LEO~\cite{rusu2018meta}}}} & & 74.49 $\pm$ 0.36 & 58.48 $\pm$ 0.48 & 80.25 $\pm$ 0.38 & 65.25 $\pm$ 0.51 & 75.86 $\pm$ 0.35 & 59.77 $\pm$ 0.47 & 82.15 $\pm$ 0.37 & 68.90 $\pm$ 0.49 \\
    
    & \multicolumn{1}{c}{} & \textit{ft} & 76.77 $\pm$ 0.35 & 60.52 $\pm$ 0.47 & 80.97 $\pm$ 0.36 & 65.44 $\pm$ 0.49 & 77.81 $\pm$ 0.34 & 61.81 $\pm$ 0.46 & 84.95 $\pm$ 0.36 & 69.59 $\pm$ 0.47\\
     
    & \multicolumn{1}{c}{} & \textit{cl} & 74.66 $\pm$ 0.36 & 58.62 $\pm$ 0.46 & 80.74 $\pm$ 0.37 & 65.37 $\pm$ 0.50 & 76.13 $\pm$ 0.35 & 60.22 $\pm$ 0.47 & 82.31 $\pm$ 0.37 & 69.23 $\pm$ 0.48\\
     
    & \multicolumn{1}{c}{} & \textit{ft+cl} & 71.91 $\pm$ 0.35 & 61.09 $\pm$ 0.47 & 81.43 $\pm$ 0.36 & 66.03 $\pm$ 0.48 & 77.72 $\pm$ 0.34 & 62.19 $\pm$ 0.45 & 85.04 $\pm$ 0.36 & 70.28 $\pm$ 0.47\\ \hhline{~|-;-|-|-|-|-|-|-|-|-|}
    
    &\multicolumn{1}{c}{\multirow{4}{*}{\shortstack{MTL~\cite{sun2019meta}}}} & & 75.65 $\pm$ 0.35 & 58.49 $\pm$ 0.46 & 81.14 $\pm$ 0.36 & 64.29 $\pm$ 0.50 & 77.30 $\pm$ 0.34 & 62.99 $\pm$ 0.46 & 83.23 $\pm$ 0.37 & 70.08 $\pm$ 0.52 \\
    
    & \multicolumn{1}{c}{} & \textit{ft} & 77.17 $\pm$ 0.35 & 58.85 $\pm$ 0.44 & 82.01 $\pm$ 0.36 & 64.67 $\pm$ 0.47 & 79.40 $\pm$ 0.34 & 63.65 $\pm$ 0.45 & 84.76 $\pm$ 0.36 & 70.25 $\pm$ 0.49\\
     
    & \multicolumn{1}{c}{} & \textit{cl} & 77.10 $\pm$ 0.34 & 58.86 $\pm$ 0.45 & 82.34 $\pm$ 0.36 & 66.70 $\pm$ 0.51 & 79.29 $\pm$ 0.35 & 63.14 $\pm$ 0.46 & 86.21 $\pm$ 0.37 & 70.16 $\pm$ 0.50\\
     
    & \multicolumn{1}{c}{} & \textit{ft+cl} & 78.03 $\pm$ 0.33 & 61.17 $\pm$ 0.45 & 82.35 $\pm$ 0.35 & 65.72 $\pm$ 0.48 & 80.20 $\pm$ 0.33 & 64.40 $\pm$ 0.45 & 86.02 $\pm$ 0.35 & 71.45 $\pm$ 0.48\\ \hhline{~|-;-|-|-|-|-|-|-|-|-|}
    
    &\multicolumn{1}{c}{\multirow{4}{*}{\shortstack{MN~\cite{vinyals2016matching}}}} & & 75.21 $\pm$ 0.35 & 61.05 $\pm$ 0.46 & 79.92 $\pm$ 0.37 & 66.01 $\pm$ 0.50 & 77.15 $\pm$ 0.36 & 63.45 $\pm$ 0.45 & 82.43 $\pm$ 0.37 & 70.38 $\pm$ 0.49 \\
    
    & \multicolumn{1}{c}{} & \textit{ft} & 75.52 $\pm$ 0.35 & 61.23 $\pm$ 0.45 & 80.18 $\pm$ 0.36 & 66.33 $\pm$ 0.49 & 77.80 $\pm$ 0.35 & 64.42 $\pm$ 0.46 & 83.82 $\pm$ 0.36 & 70.90 $\pm$ 0.50\\
     
    & \multicolumn{1}{c}{} & \textit{cl} & 75.40 $\pm$ 0.34 & 61.14 $\pm$ 0.44 & 80.04 $\pm$ 0.35 & 66.26 $\pm$ 0.50 & 77.23 $\pm$ 0.35 & 64.21 $\pm$ 0.47 & 82.77 $\pm$ 0.35 & 70.61 $\pm$ 0.51\\
     
    & \multicolumn{1}{c}{} & \textit{ft+cl} & 76.73 $\pm$ 0.34 & 62.64 $\pm$ 0.46 & 80.79 $\pm$ 0.35 & 67.30 $\pm$ 0.48 & 78.55 $\pm$ 0.36 & 64.89 $\pm$ 0.44 & 84.03 $\pm$ 0.36 & 71.41 $\pm$ 0.49\\ \hhline{~|-;-|-|-|-|-|-|-|-|-|}
    
    &\multicolumn{1}{c}{\multirow{4}{*}{\shortstack{SIB~\cite{hu1empirical}\\(transductive)}}} & & 78.88 $\pm$ 0.35 & 67.10 $\pm$ 0.56 & 85.09 $\pm$ 0.35 & 77.64 $\pm$ 0.58 & 81.73 $\pm$ 0.34 & 71.31 $\pm$ 0.56 & 88.19 $\pm$ 0.34 & 81.97 $\pm$ 0.56 \\
    
    & \multicolumn{1}{c}{} & \textit{ft} & 79.58 $\pm$ 0.35 & 67.94 $\pm$ 0.55 & 85.12 $\pm$ 0.35 & 77.68 $\pm$ 0.57 & 82.00 $\pm$ 0.34 & 71.95 $\pm$ 0.56 & 88.20 $\pm$ 0.34 & 82.01 $\pm$ 0.56\\
     
    & \multicolumn{1}{c}{} & \textit{cl} & 79.04 $\pm$ 0.33 & 67.77 $\pm$ 0.55 & 85.22 $\pm$ 0.35 & 77.72 $\pm$ 0.56 & 81.93 $\pm$ 0.35 & 71.66 $\pm$ 0.56 & 88.21 $\pm$ 0.33 & 82.01 $\pm$ 0.54\\
     
    & \multicolumn{1}{c}{} & \textit{ft+cl} & 80.32 $\pm$ 0.35 & 68.85 $\pm$ 0.56 & 85.43 $\pm$ 0.35 & 78.03 $\pm$ 0.57 & 83.21 $\pm$ 0.33 & 73.51 $\pm$ 0.56 & 88.69 $\pm$ 0.33 & 83.07 $\pm$ 0.52\\ \hhline{~|-;-|-|-|-|-|-|-|-|-|}
    
    &\multicolumn{1}{c}{\multirow{4}{*}{\shortstack{SIB~\cite{hu1empirical}\\(inductive)}}} & & 75.64 $\pm$ 0.36 & 57.20 $\pm$ 0.57 & 81.69 $\pm$ 0.34 & 65.51 $\pm$ 0.56 & 78.17 $\pm$ 0.35 & 60.12 $\pm$ 0.56 & 84.96 $\pm$ 0.36 & 69.20 $\pm$ 0.58 \\
    
    & \multicolumn{1}{c}{} & \textit{ft} & 76.23 $\pm$ 0.35 & 58.67 $\pm$ 0.56 & 82.04 $\pm$ 0.35 & 66.69 $\pm$ 0.57 & 79.34 $\pm$ 0.35 & 61.77 $\pm$ 0.56 & 85.24 $\pm$ 0.36 & 70.05 $\pm$ 0.57\\
     
    & \multicolumn{1}{c}{} & \textit{cl} & 76.61 $\pm$ 0.35 & 58.12 $\pm$ 0.55 & 82.21 $\pm$ 0.35 & 66.28 $\pm$ 0.56 & 79.11 $\pm$ 0.35 & 61.25 $\pm$ 0.55 & 85.63 $\pm$ 0.34 & 69.90 $\pm$ 0.57\\
     
    \parbox[t]{2mm}{\multirow{-24}{*}{\rotatebox[origin=c]{90}{\textbf{Meta-Learning}}}} & \multicolumn{1}{c}{} & \textit{ft+cl} & 77.68 $\pm$ 0.34 & 60.33 $\pm$ 0.54 & 82.75 $\pm$ 0.35 & 67.34 $\pm$ 0.55 & 80.05 $\pm$ 0.34 & 63.14 $\pm$ 0.54 & 86.14 $\pm$ 0.34 & 71.45 $\pm$ 0.55\\
    \hhline{-|-|-|-|-|-|-|-|-|-|-|}
    \hhline{-|-|-|-|-|-|-|-|-|-|-|}
\end{tabular}
    \label{tab:a1}
\end{figure}

\newpage
\subsection{Hardness-Specific Acc}
\label{sec:a6.2}
\begin{figure}[hbt!]
    \centering
    \begin{subfigure}[t]{0.3\linewidth}
        \centering
        \includegraphics[height=1.2in]{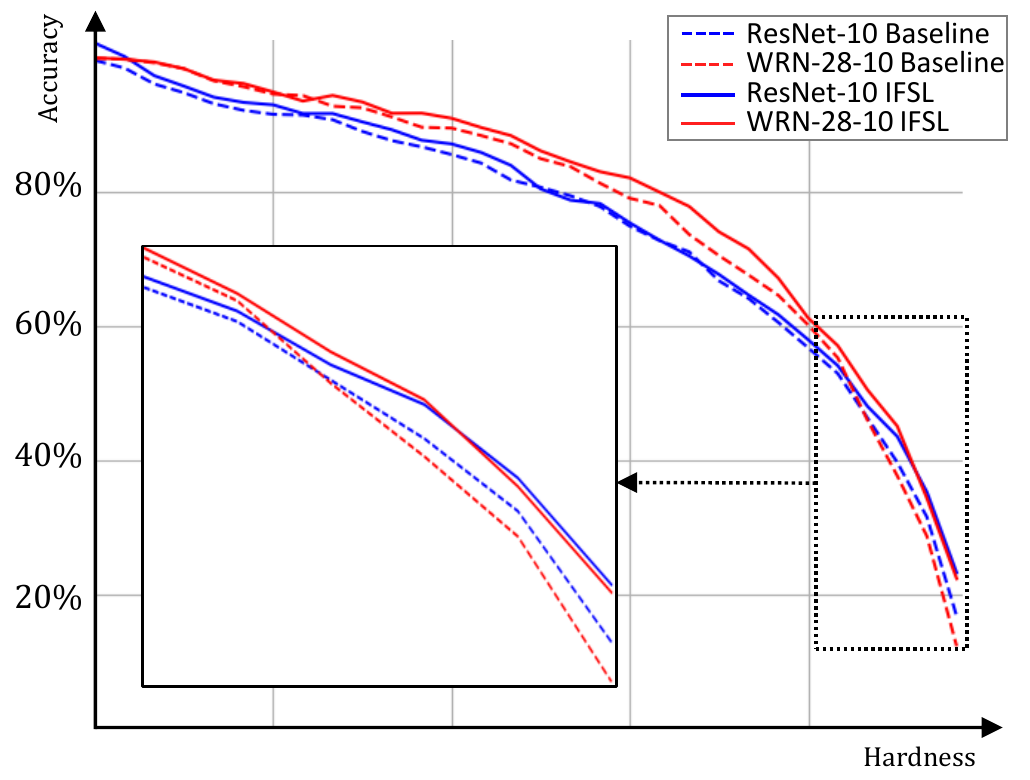}
        \caption{Linear}
    \end{subfigure}%
    ~ 
    \begin{subfigure}[t]{0.3\linewidth}
        \centering
        \includegraphics[height=1.2in]{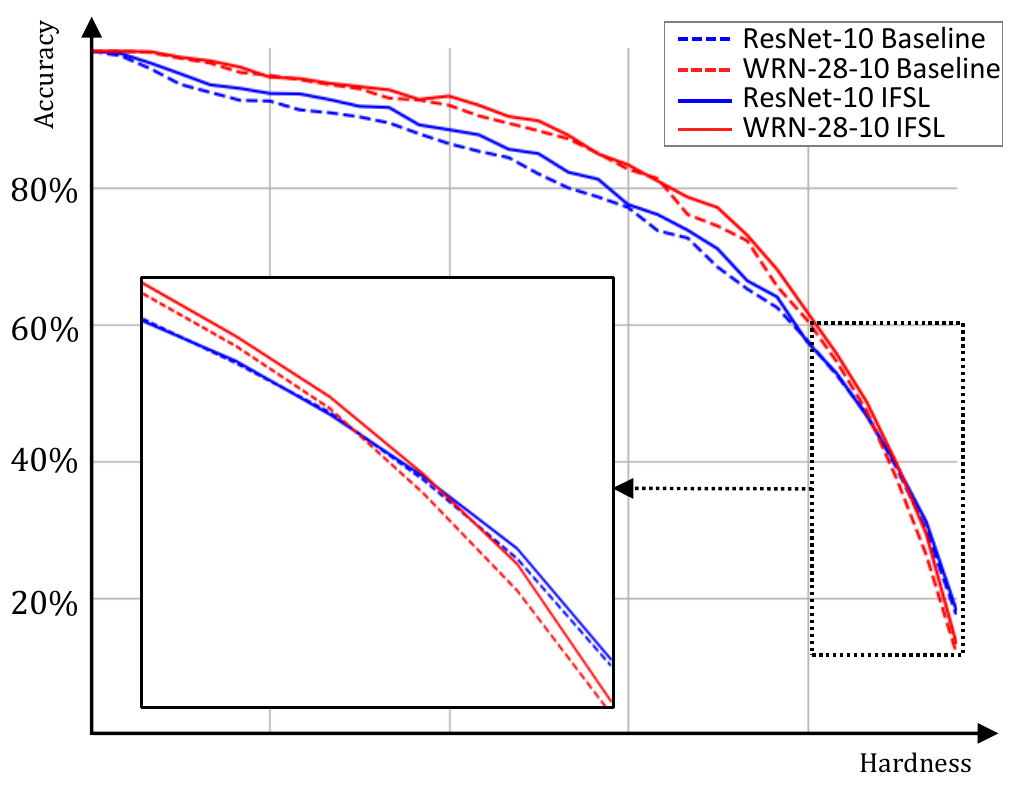}
        \caption{Cosine}
    \end{subfigure}%
    \begin{subfigure}[t]{0.3\linewidth}
        \centering
        \includegraphics[height=1.2in]{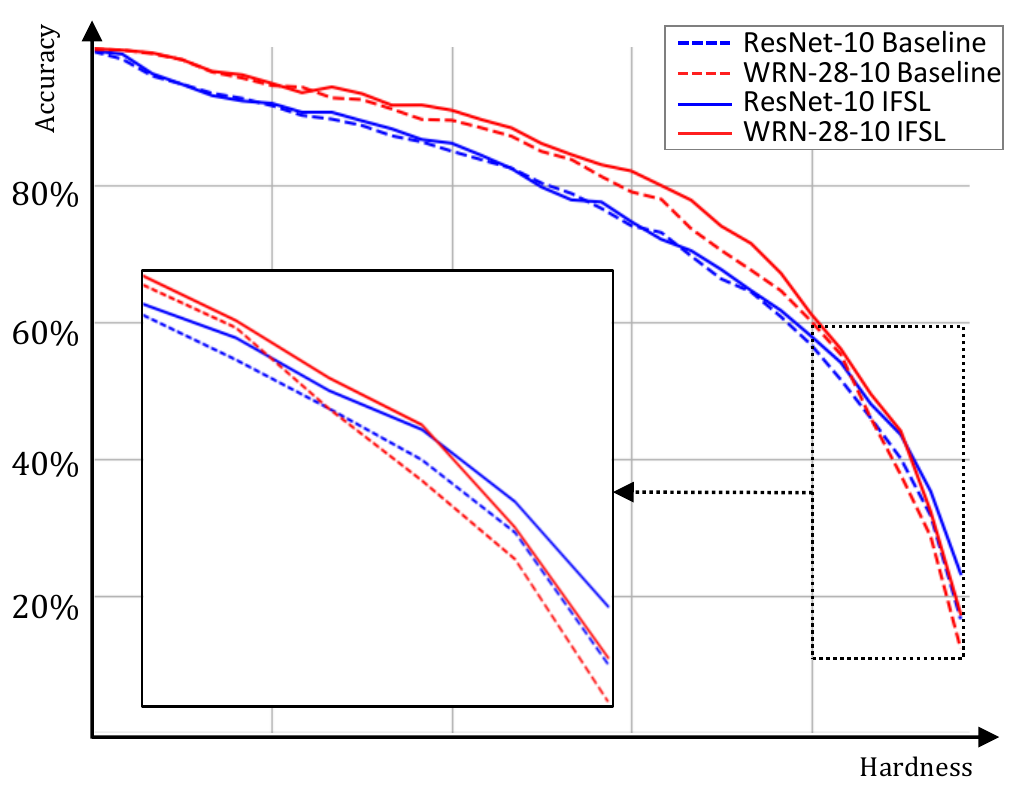}
        \caption{$k$-NN}
    \end{subfigure}
    \begin{subfigure}[t]{0.3\linewidth}
        \centering
        \includegraphics[height=1.2in]{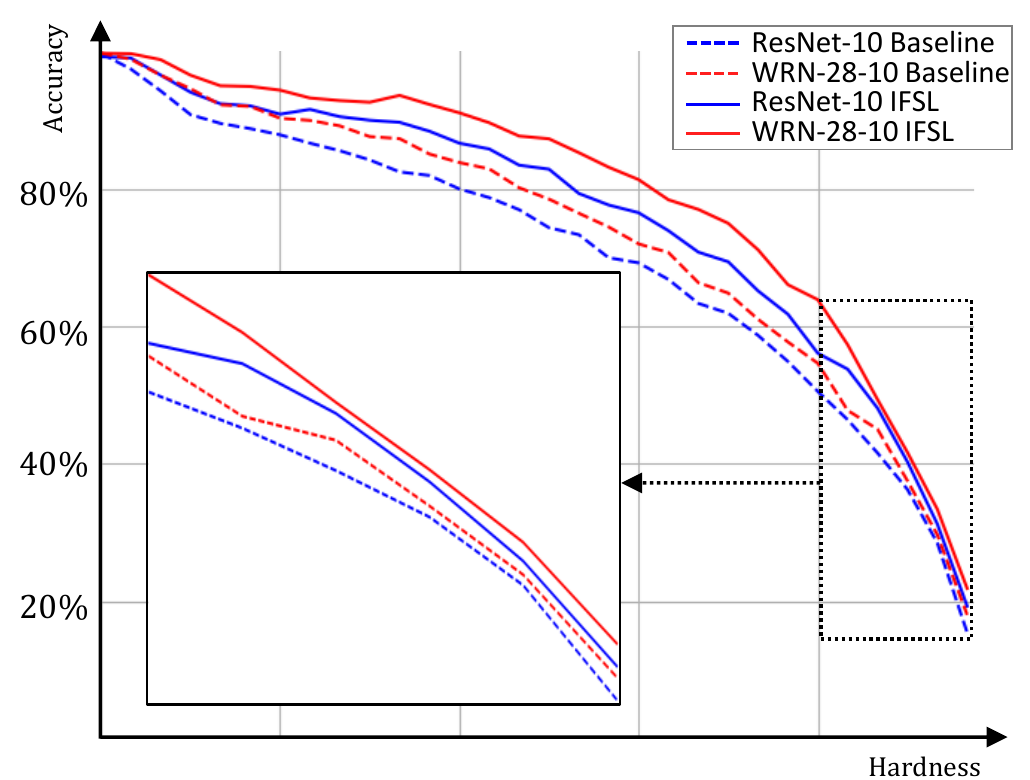}
        \caption{MAML~\cite{finn2017model}}
    \end{subfigure}%
    \begin{subfigure}[t]{0.3\linewidth}
        \centering
        \includegraphics[height=1.2in]{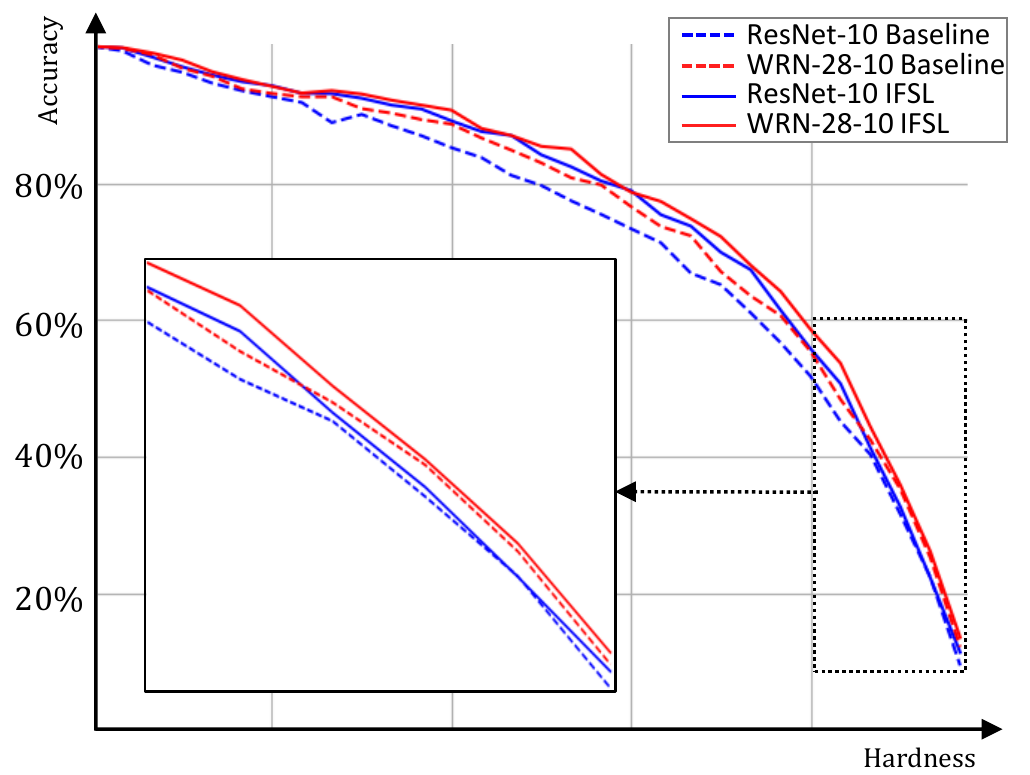}
        \caption{LEO~\cite{rusu2018meta}}
    \end{subfigure}%
    \begin{subfigure}[t]{0.3\linewidth}
        \centering
        \includegraphics[height=1.2in]{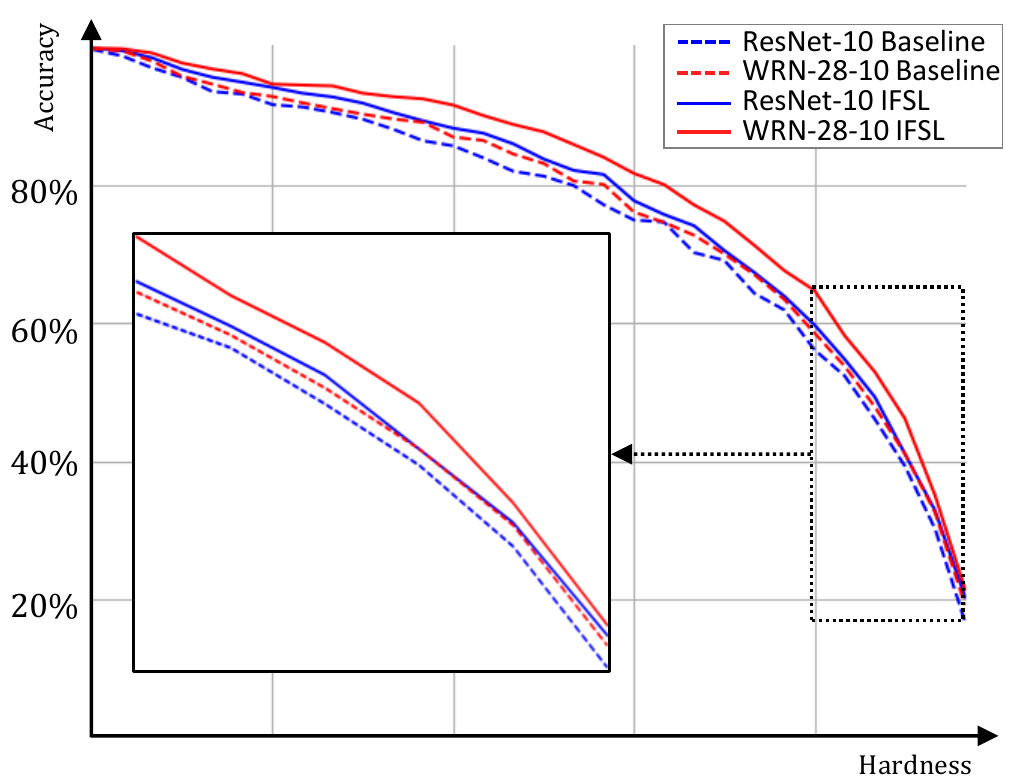}
        \caption{MTL~\cite{sun2019meta}}
    \end{subfigure}
    \begin{subfigure}[t]{0.3\linewidth}
        \centering
        \includegraphics[height=1.2in]{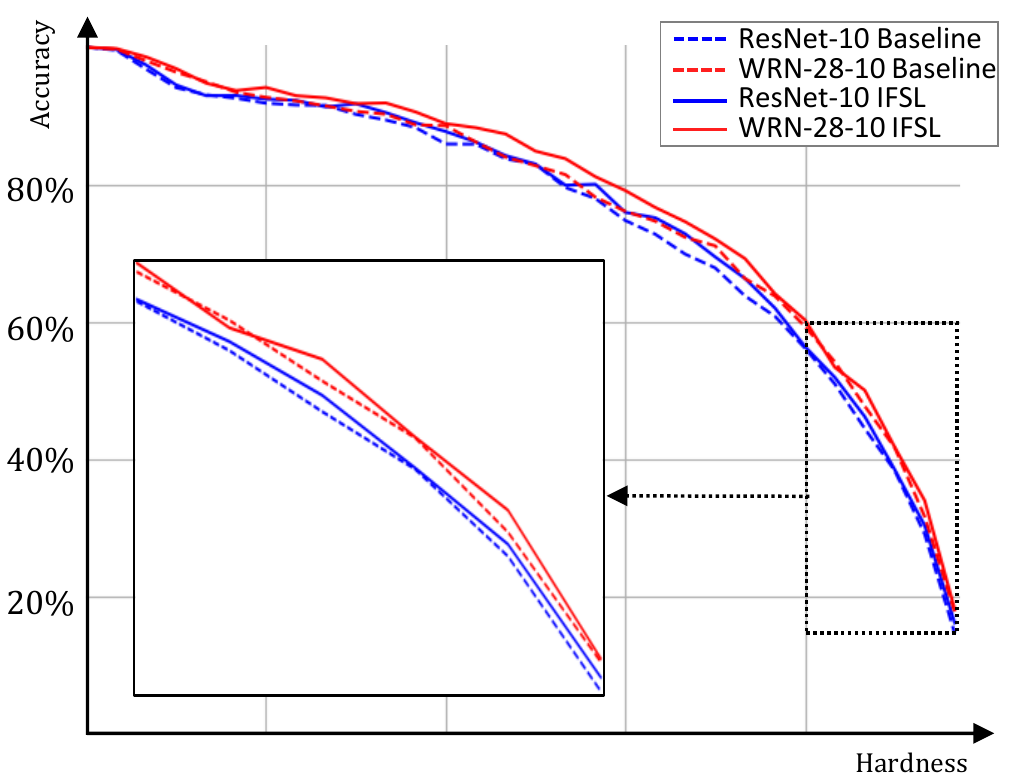}
        \caption{Matching Net~\cite{vinyals2016matching}}
    \end{subfigure}%
    \begin{subfigure}[t]{0.3\linewidth}
        \centering
        \includegraphics[height=1.2in]{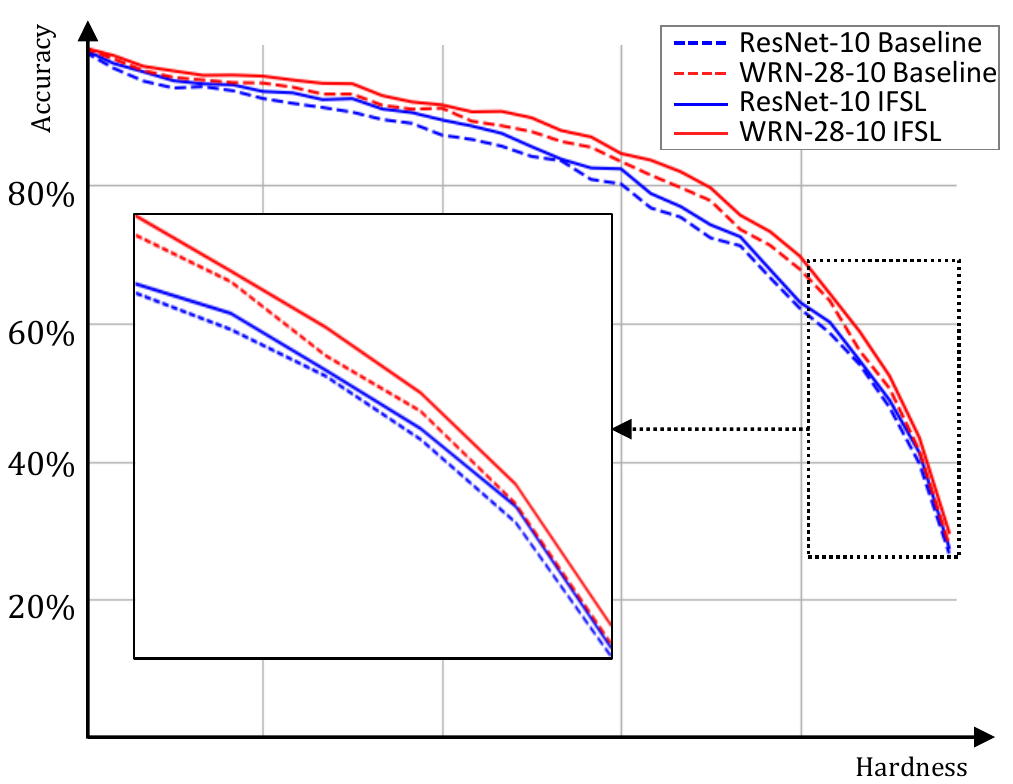}
        \caption{SIB(transductive)~\cite{hu1empirical}}
    \end{subfigure}%
    \begin{subfigure}[t]{0.3\linewidth}
        \centering
        \includegraphics[height=1.2in]{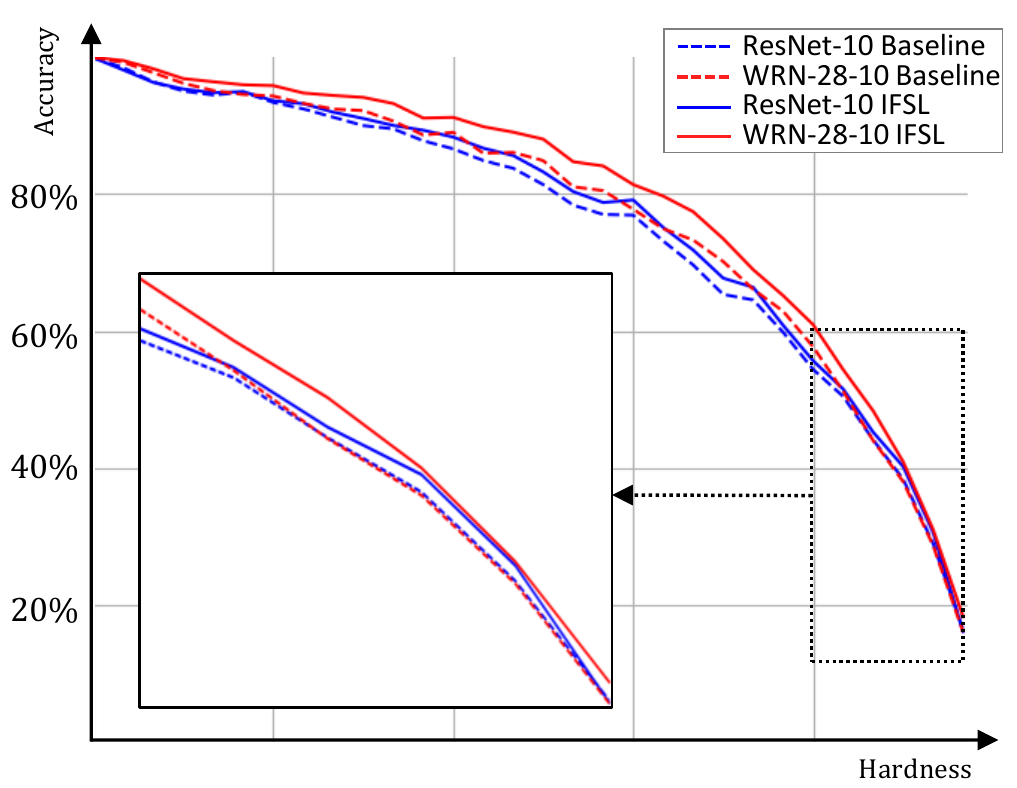}
        \caption{SIB(inductive)~\cite{hu1empirical}}
    \end{subfigure}
    \caption{Supplementary to Figure 5. Hardness-specific Acc of 5-shot fine-tuning and meta-learning on \textit{mini}ImageNet.}
    \label{fig:a2}
\end{figure}
\begin{figure}[!htbp]
    \centering
    \begin{subfigure}[t]{0.3\linewidth}
        \centering
        \includegraphics[height=1.2in]{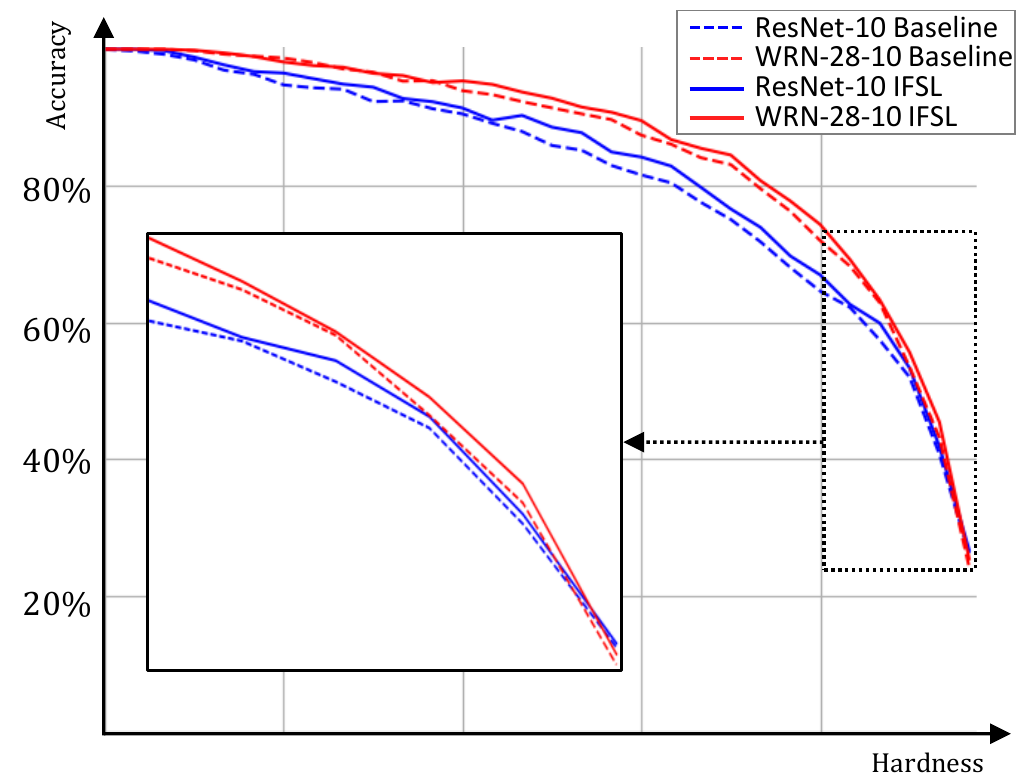}
        \caption{Linear}
    \end{subfigure}%
    ~ 
    \begin{subfigure}[t]{0.3\linewidth}
        \centering
        \includegraphics[height=1.2in]{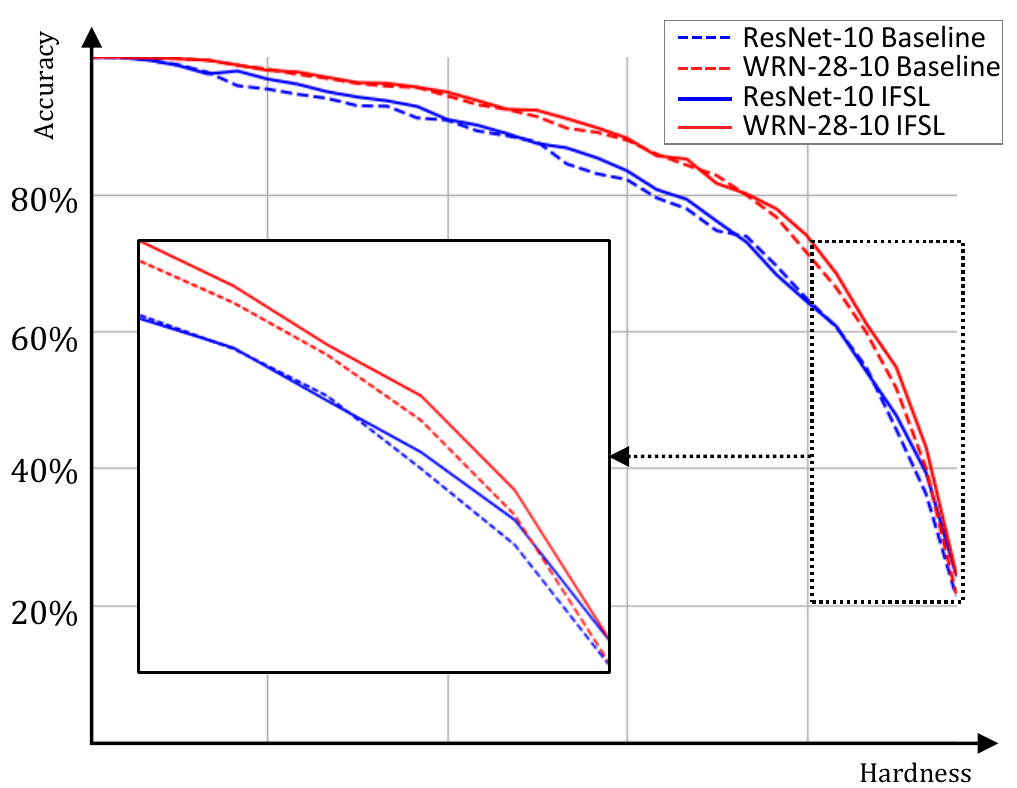}
        \caption{Cosine}
    \end{subfigure}%
    \begin{subfigure}[t]{0.3\linewidth}
        \centering
        \includegraphics[height=1.2in]{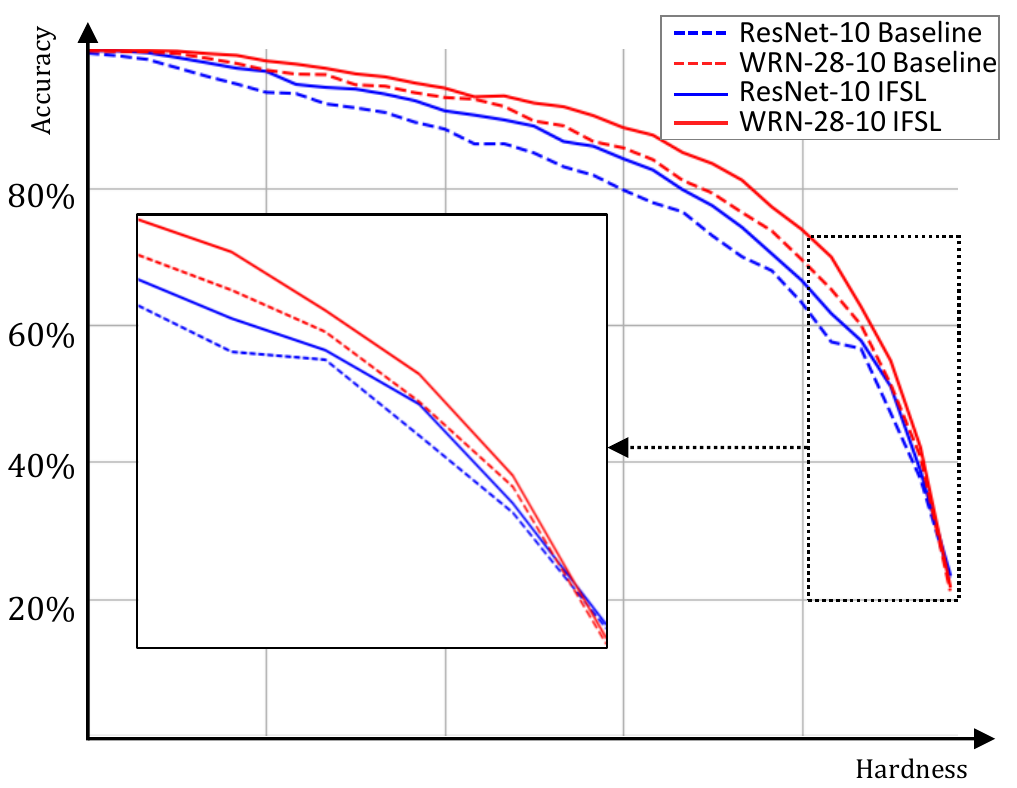}
        \caption{$k$-NN}
    \end{subfigure}
    \begin{subfigure}[t]{0.3\linewidth}
        \centering
        \includegraphics[height=1.2in]{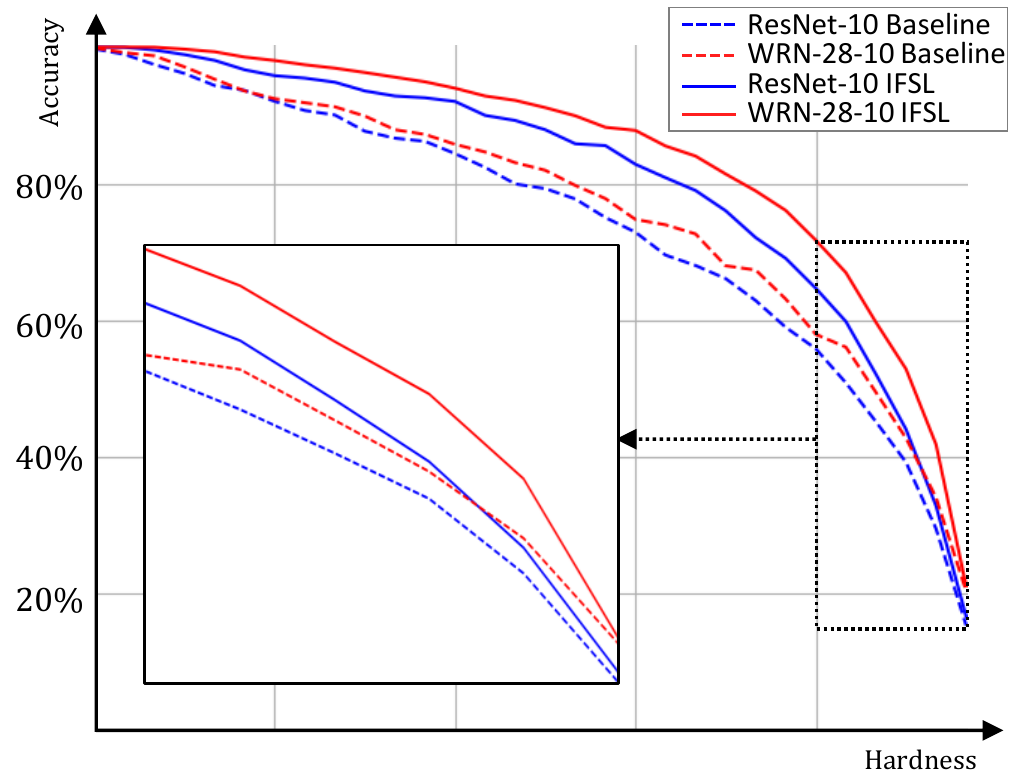}
        \caption{MAML~\cite{finn2017model}}
    \end{subfigure}%
    \begin{subfigure}[t]{0.3\linewidth}
        \centering
        \includegraphics[height=1.2in]{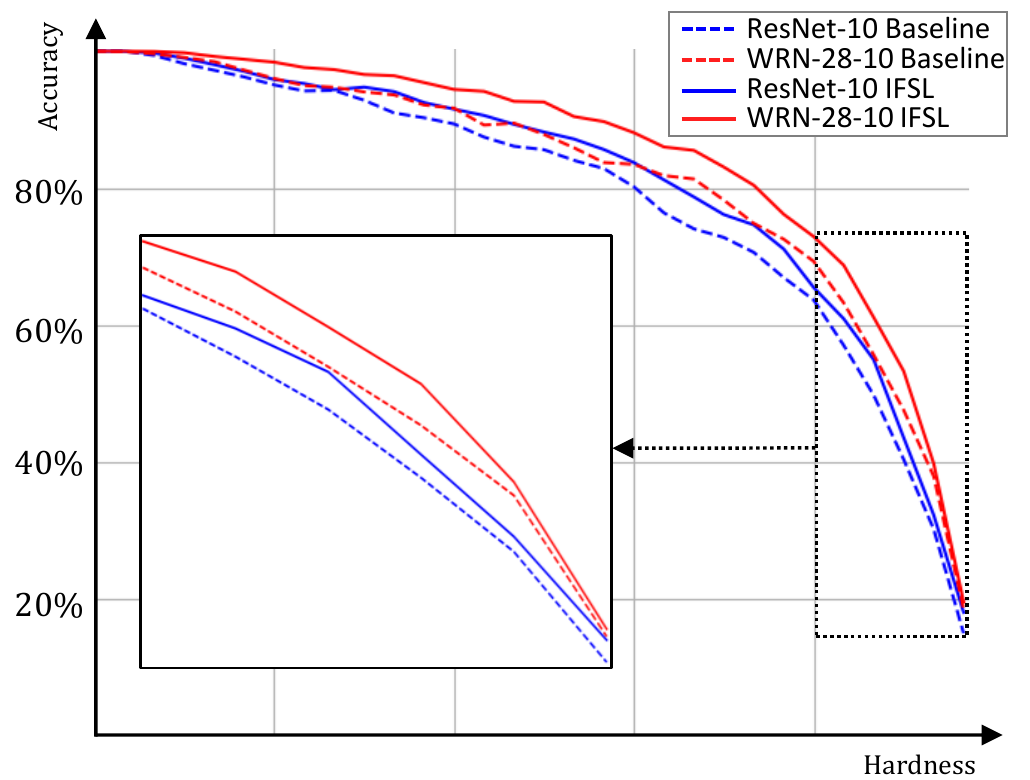}
        \caption{LEO~\cite{rusu2018meta}}
    \end{subfigure}%
    \begin{subfigure}[t]{0.3\linewidth}
        \centering
        \includegraphics[height=1.2in]{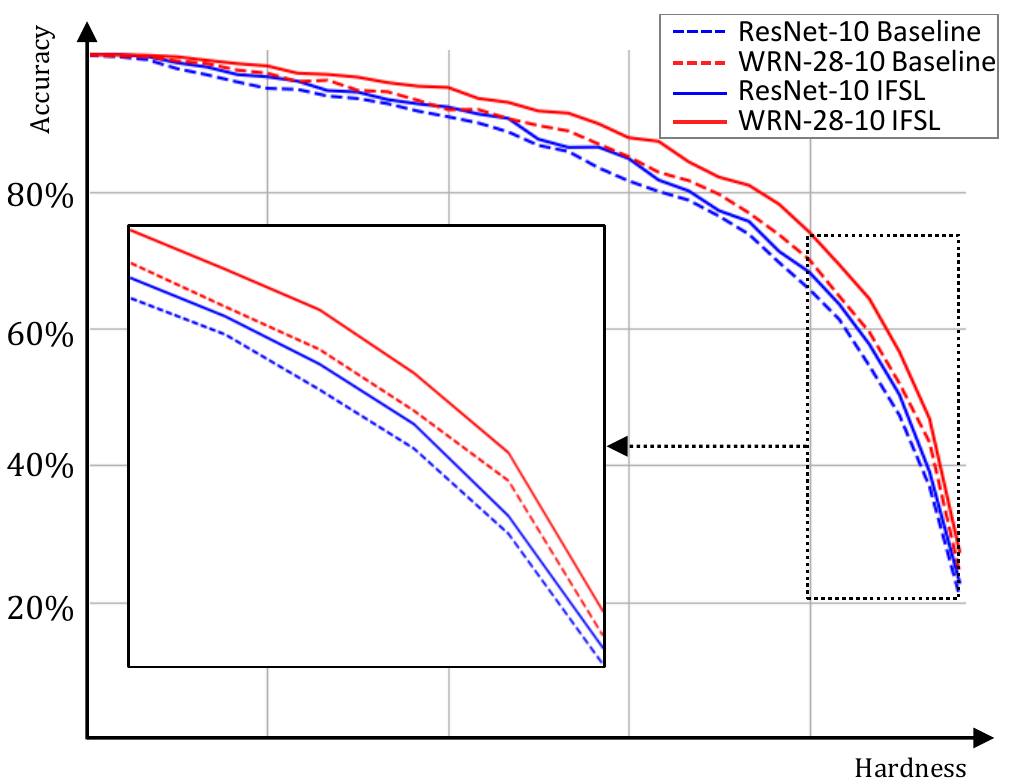}
        \caption{MTL~\cite{sun2019meta}}
    \end{subfigure}
    \begin{subfigure}[t]{0.3\linewidth}
        \centering
        \includegraphics[height=1.2in]{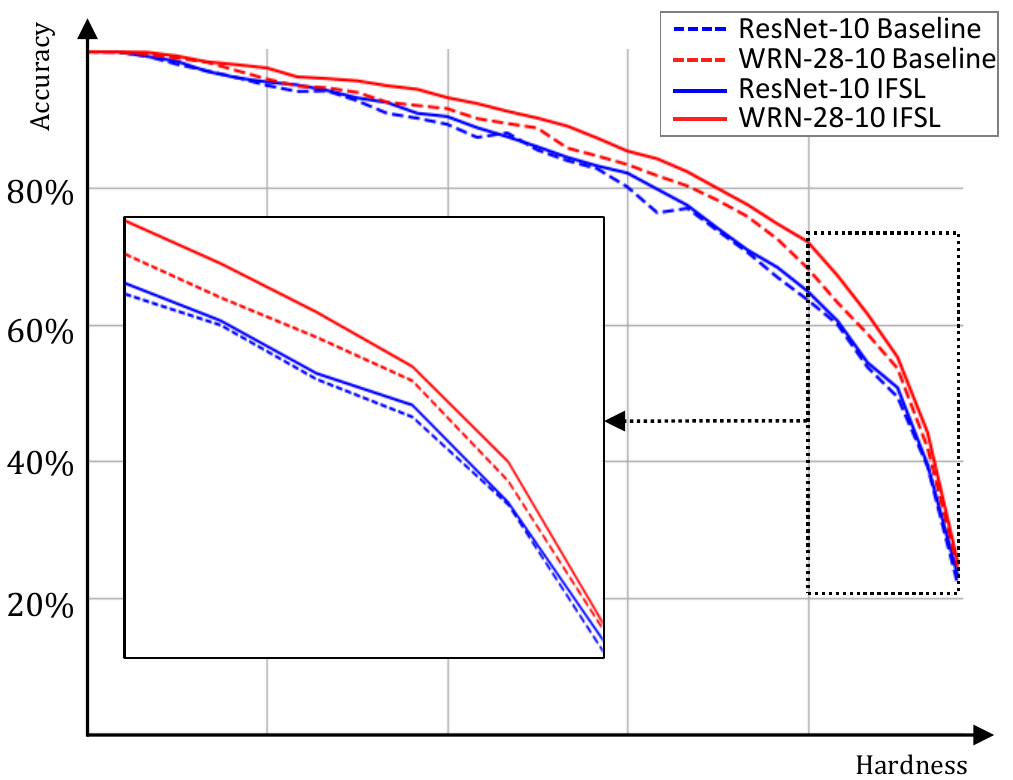}
        \caption{Matching Net~\cite{vinyals2016matching}}
    \end{subfigure}%
    \begin{subfigure}[t]{0.3\linewidth}
        \centering
        \includegraphics[height=1.2in]{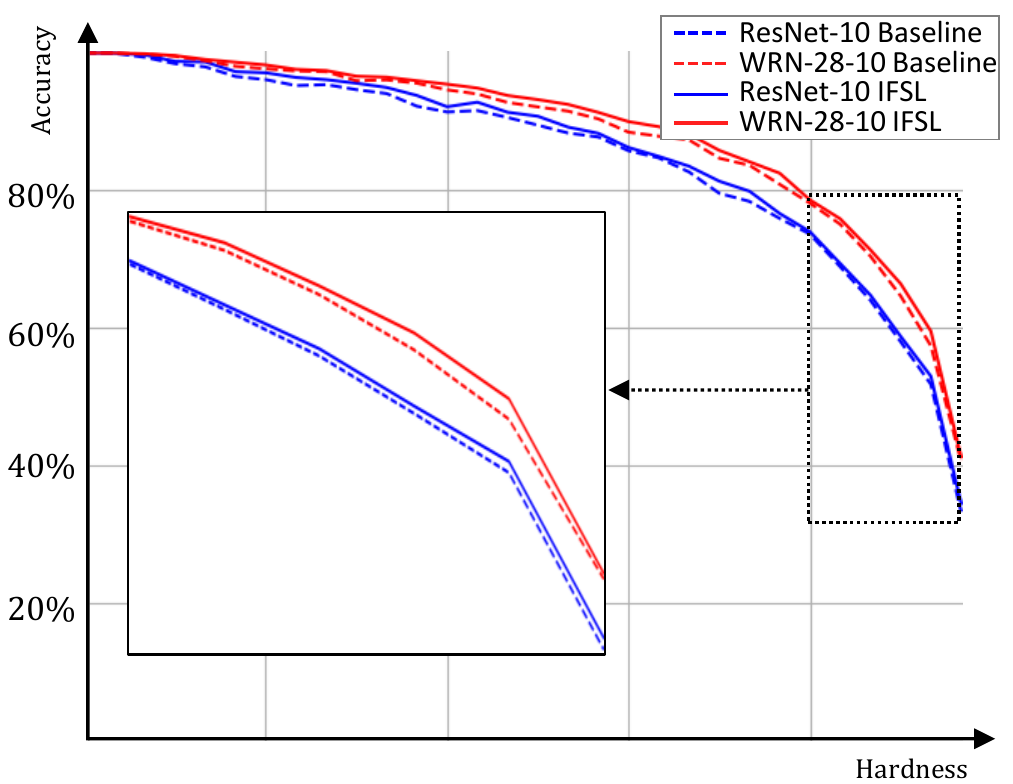}
        \caption{SIB(transductive)~\cite{hu1empirical}}
    \end{subfigure}%
    \begin{subfigure}[t]{0.3\linewidth}
        \centering
        \includegraphics[height=1.2in]{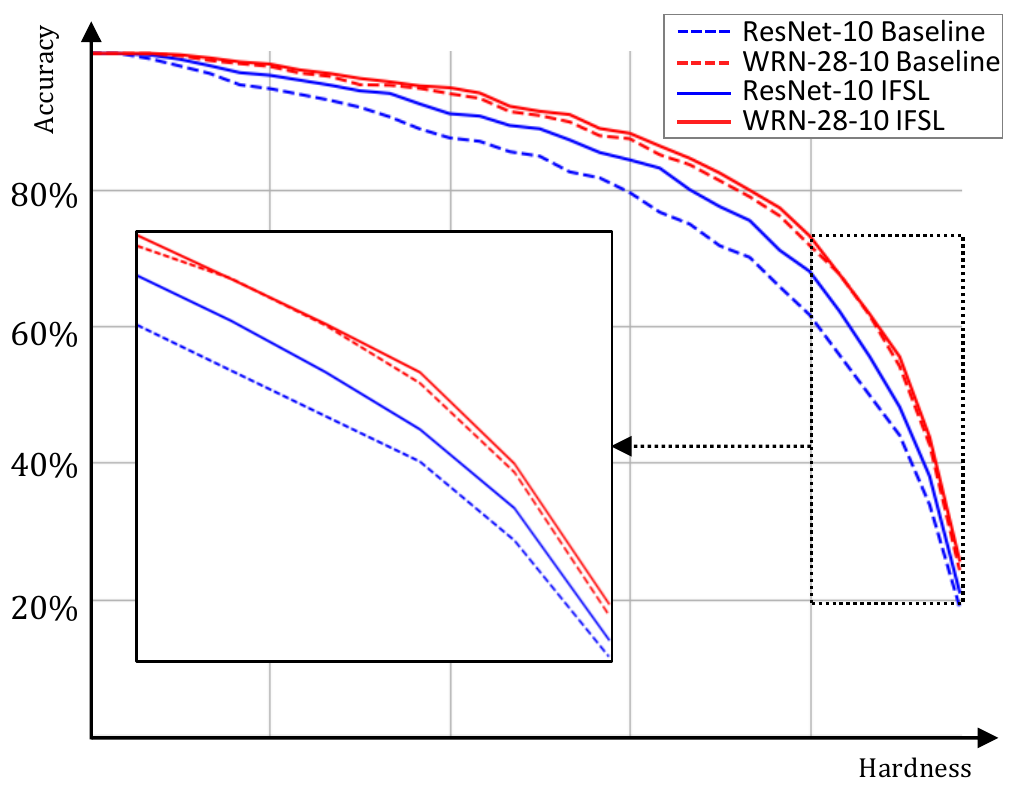}
        \caption{SIB(inductive)~\cite{hu1empirical}}
    \end{subfigure}
    \caption{Supplementary to Figure 5. Hardness-specific Acc of 5-shot fine-tuning and meta-learning on \textit{tiered}ImageNet.}
    \label{fig:a3}
\end{figure}

\newpage
\subsection{CAM-Acc}
\label{sec:a6.3}
\begin{figure}[!htbp]
    \centering
    \fontsize{6}{7.2}\selectfont
    \renewcommand\arraystretch{1.2}
    \captionsetup{width=\linewidth}
    \captionof{table}{Supplementary to Figure 6. CAM-Acc (\%) on fine-tuning and meta-learning. We used combined adjustment for IFSL.}
    \setlength\tabcolsep{4pt}
    \begin{tabular}{@{\hskip0pt}c | c c c c c c c c c c}
    \hhline{-|-|-|-|-|-|-|-|-|-|-|}
    \hhline{-|-|-|-|-|-|-|-|-|-|-|}
    \multicolumn{3}{c}{} & \multicolumn{4}{c}{\textbf{ResNet-10}} & \multicolumn{4}{c}{\textbf{WRN-28-10}} \\
    
    \multicolumn{3}{c}{} & \multicolumn{2}{c}{\textit{mini}ImageNet} & \multicolumn{2}{c}{\textit{tiered}ImageNet} & \multicolumn{2}{c}{\textit{mini}ImageNet} & \multicolumn{2}{c}{\textit{tiered}ImageNet} \\ \cmidrule(lr){4-5} \cmidrule(lr){6-7} \cmidrule(lr){8-9} \cmidrule(lr){10-11}
    
    \multicolumn{3}{c}{\multirow{-3}{*}[0.3em]{\textbf{Method}}} & $5$-shot & $1$-shot & $5$-shot & $1$-shot & $5$-shot & $1$-shot & $5$-shot & $1$-shot\\
    \hline
    
     & \multicolumn{1}{c}{\multirow{2}{*}{\shortstack{Linear}}} & & 29.02 $\pm$ 0.38 & 25.22 $\pm$ 0.38 & 31.62 $\pm$ 0.38 & 31.05 $\pm$ 0.39 & 25.99 $\pm$ 0.35 & 24.74 $\pm$ 0.34 & 30.17 $\pm$ 0.36 & 29.76 $\pm$ 0.37\\
     
    & \multicolumn{1}{c}{} & +IFSL & 29.85 $\pm$ 0.37 & 26.67 $\pm$ 0.38 & 31.75 $\pm$ 0.38 & 31.43 $\pm$ 0.37 & 26.02 $\pm$ 0.37 & 24.96 $\pm$ 0.36 & 32.57 $\pm$ 0.36 & 30.64 $\pm$ 0.38\\ \hhline{~|-;-|-|-|-|-|-|-|-|-|}
    
    &\multicolumn{1}{c}{\multirow{2}{*}{\shortstack{Cosine}}} & & 28.10 $\pm$ 0.37 & 27.12 $\pm$ 0.38 & 29.82 $\pm$ 0.37 & 28.54 $\pm$ 0.38 & 27.54 $\pm$ 0.38 & 25.73 $\pm$ 0.37 & 32.60 $\pm$ 0.35 & 31.21 $\pm$ 0.36 \\
     
    & \multicolumn{1}{c}{} & +IFSL & 28.18 $\pm$ 0.37 & 27.26 $\pm$ 0.38 & 31.38 $\pm$ 0.37 & 28.70 $\pm$ 0.40 & 27.82 $\pm$ 0.38 & 25.85 $\pm$ 0.38 & 33.65 $\pm$ 0.37 & 31.66 $\pm$ 0.35\\ \hhline{~|-;-|-|-|-|-|-|-|-|-|}
    
    &\multicolumn{1}{c}{\multirow{2}{*}{\shortstack{$k$-NN}}} & & 27.96 $\pm$ 0.37 & 26.65 $\pm$ 0.37 & 32.25 $\pm$ 0.39 & 30.36 $\pm$ 0.39 & 24.15 $\pm$ 0.34 & 23.30 $\pm$ 0.33 & 23.91 $\pm$ 0.34 & 21.99 $\pm$ 0.33 \\
     
    \multirow{-6}{*}{\STAB{\rotatebox[origin=c]{90}{\textbf{Fine-Tuning}}}} & \multicolumn{1}{c}{} & +IFSL & 28.15 $\pm$ 0.37 & 26.81 $\pm$ 0.37 & 32.75 $\pm$ 0.39 & 30.84 $\pm$ 0.39 & 25.23 $\pm$ 0.36 & 24.14 $\pm$ 0.35 & 28.04 $\pm$ 0.37 & 26.46 $\pm$ 0.37\\
    
    \hline
    
    &\multicolumn{1}{c}{\multirow{2}{*}{\shortstack{MAML~\cite{finn2017model}}}} & & 29.43 $\pm$ 0.37 & 27.39 $\pm$ 0.38 & 32.72 $\pm$ 0.40 & 32.14 $\pm$ 0.40 & 27.56 $\pm$ 0.36 & 26.46 $\pm$ 0.36 & 34.39 $\pm$ 0.40 & 31.07 $\pm$ 0.39 \\

    & \multicolumn{1}{c}{} & +IFSL & 30.06 $\pm$ 0.38 & 28.42 $\pm$ 0.38 & 32.93 $\pm$ 0.40 & 32.24 $\pm$ 0.39 & 27.61 $\pm$ 0.36 & 26.91 $\pm$ 0.38 & 34.57 $\pm$ 0.41 & 31.22 $\pm$ 0.40\\ \hhline{~|-;-|-|-|-|-|-|-|-|-|}
    
    &\multicolumn{1}{c}{\multirow{2}{*}{\shortstack{LEO~\cite{rusu2018meta}}}} & & 30.24 $\pm$ 0.38 & 28.56 $\pm$ 0.37 & 31.64 $\pm$ 0.38 & 29.88 $\pm$ 0.37 & 29.15 $\pm$ 0.38 & 27.86 $\pm$ 0.38 & 31.27 $\pm$ 0.37 & 29.73 $\pm$ 0.38 \\
     
    & \multicolumn{1}{c}{} & +IFSL & 30.67 $\pm$ 0.37 & 28.76 $\pm$ 0.37 & 32.01 $\pm$ 0.38 & 30.65 $\pm$ 0.37 & 29.20 $\pm$ 0.37 & 28.45 $\pm$ 0.38 & 31.98 $\pm$ 0.39 & 30.32 $\pm$ 0.38\\ \hhline{~|-;-|-|-|-|-|-|-|-|-|}
    
    &\multicolumn{1}{c}{\multirow{2}{*}{\shortstack{MTL~\cite{sun2019meta}}}} & & 31.45 $\pm$ 0.39 & 30.13 $\pm$ 0.39 & 33.52 $\pm$ 0.39 & 33.11 $\pm$ 0.39 & 30.56 $\pm$ 0.39 & 29.78 $\pm$ 0.40 & 33.13 $\pm$ 0.39 & 32.35 $\pm$ 0.39 \\

    & \multicolumn{1}{c}{} & +IFSL & 34.21 $\pm$ 0.39 & 31.59 $\pm$ 0.40 & 33.67 $\pm$ 0.38 & 33.50 $\pm$ 0.39 & 31.78 $\pm$ 0.39 & 30.12 $\pm$ 0.39 & 33.30 $\pm$ 0.39 & 32.64 $\pm$ 0.39\\ \hhline{~|-;-|-|-|-|-|-|-|-|-|}
    
    &\multicolumn{1}{c}{\multirow{2}{*}{\shortstack{MN~\cite{vinyals2016matching}}}} & & 28.50 $\pm$ 0.38 & 28.42 $\pm$ 0.39 & 32.55 $\pm$ 0.40 & 31.88 $\pm$ 0.39 & 24.93 $\pm$ 0.38 & 25.34 $\pm$ 0.39 & 34.87 $\pm$ 0.37 & 29.10 $\pm$ 0.38 \\

    & \multicolumn{1}{c}{} & +IFSL & 28.68 $\pm$ 0.38 & 28.77 $\pm$ 0.38 & 32.67 $\pm$ 0.40 & 32.10 $\pm$ 0.40 & 27.93 $\pm$ 0.37 & 25.81 $\pm$ 0.37 & 35.47 $\pm$ 0.41 & 30.71 $\pm$ 0.39\\ \hhline{~|-;-|-|-|-|-|-|-|-|-|}
    
    &\multicolumn{1}{c}{\multirow{2}{*}{\shortstack{SIB~\cite{hu1empirical}\\(transductive)}}} & & 32.10 $\pm$ 0.39 & 31.19 $\pm$ 0.39 & 32.16 $\pm$ 0.39 & 30.49 $\pm$ 0.39 & 28.32 $\pm$ 0.37 & 26.76 $\pm$ 0.38 & 31.02 $\pm$ 0.36 & 28.43 $\pm$ 0.38 \\
     
    & \multicolumn{1}{c}{} & +IFSL & 32.14 $\pm$ 0.39 & 31.34 $\pm$ 0.39 & 34.31 $\pm$ 0.40 & 32.59 $\pm$ 0.40 & 31.54 $\pm$ 0.38 & 29.82 $\pm$ 0.36 & 32.33 $\pm$ 0.37 & 30.26 $\pm$ 0.38\\ \hhline{~|-;-|-|-|-|-|-|-|-|-|}
    
    &\multicolumn{1}{c}{\multirow{2}{*}{\shortstack{SIB~\cite{hu1empirical}\\(inductive)}}} & & 31.26 $\pm$ 0.38 & 30.56 $\pm$ 0.39 & 31.35 $\pm$ 0.40 & 30.48 $\pm$ 0.39 & 29.76 $\pm$ 0.38 & 28.02 $\pm$ 0.37 & 29.45 $\pm$ 0.39 & 27.98 $\pm$ 0.39 \\

    \parbox[t]{2mm}{\multirow{-12}{*}{\rotatebox[origin=c]{90}{\textbf{Meta-Learning}}}} & \multicolumn{1}{c}{} & +IFSL & 31.46 $\pm$ 0.39 & 30.78 $\pm$ 0.40 & 31.56 $\pm$ 0.39 & 30.89 $\pm$ 0.40 & 30.23 $\pm$ 0.37 & 28.75 $\pm$ 0.39 & 30.07 $\pm$ 0.40 & 28.57 $\pm$ 0.39\\
    
    \hhline{-|-|-|-|-|-|-|-|-|-|-|}
    \hhline{-|-|-|-|-|-|-|-|-|-|-|}
\end{tabular}
    \label{tab:a2}
\end{figure}

\newpage
\subsection{Cross-Domain Evaluation}
\label{sec:a6.4}
\begin{figure}[!htbp]
    \centering
    \fontsize{7}{8.4}\selectfont
    \renewcommand\arraystretch{1}
    \captionsetup{width=\linewidth}
    \captionof{table}{Supplementary to Table 3. Acc (\%) and 95\% confidence interval averaged over 2000 5-way FSL tasks on cross-domain evaluation. Specifically, ``\emph{ft}'' refers to feature-wise adjustment, ``\emph{cl}'' refers to class-wise adjustment and ``\emph{ft+cl}'' refers to combined adjustment.}
    \setlength\tabcolsep{12pt}
    \begin{tabular}{@{\hskip0pt}c | c c c c c c}
    \hhline{-|-|-|-|-|-|-|}
    \hhline{-|-|-|-|-|-|-|}
    \multicolumn{3}{c}{} & \multicolumn{2}{c}{\textbf{ResNet-10}} & \multicolumn{2}{c}{\textbf{WRN-28-10}}\\ \cmidrule(lr){4-5} \cmidrule(lr){6-7}
    
    \multicolumn{3}{c}{\multirow{-2}{*}[0.3em]{\textbf{Method}}} & $5$-shot & $1$-shot & $5$-shot & $1$-shot\\
    \hline
    
     & \multicolumn{1}{c}{\multirow{4}{*}{\shortstack{Linear}}} & & 58.84 $\pm$ 0.41 & 42.25 $\pm$ 0.42 & 62.12 $\pm$ 0.40 & 42.89 $\pm$ 0.41 \\
     
     & \multicolumn{1}{c}{} & \textit{ft} & 60.12 $\pm$ 0.39 & 42.30 $\pm$ 0.41 & 63.13 $\pm$ 0.39 & 43.39 $\pm$ 0.40\\
     
     & \multicolumn{1}{c}{} & \textit{cl} & 60.51 $\pm$ 0.40 & 42.43 $\pm$ 0.42 & 62.95 $\pm$ 0.39 & 44.21 $\pm$ 0.40\\
     
    & \multicolumn{1}{c}{} & \textit{ft+cl} & 60.65 $\pm$ 0.39 & 45.14 $\pm$ 0.40 & 64.15 $\pm$ 0.38 & 45.64 $\pm$ 0.39\\ \hhline{~|-;-|-|-|-|-|}
    
    &\multicolumn{1}{c}{\multirow{4}{*}{\shortstack{Cosine}}} & & 58.30 $\pm$ 0.39 & 40.47 $\pm$ 0.40 & 60.21 $\pm$ 0.39 & 42.12 $\pm$ 0.39\\
    
    & \multicolumn{1}{c}{} & \textit{ft} & 58.32 $\pm$ 0.39 & 41.01 $\pm$ 0.40 & 61.16 $\pm$ 0.38 & 42.35 $\pm$ 0.41\\
     
     & \multicolumn{1}{c}{} & \textit{cl} & 58.68 $\pm$ 0.39 & 40.67 $\pm$ 0.41 & 61.87 $\pm$ 0.40 & 43.23 $\pm$ 0.40\\
     
    & \multicolumn{1}{c}{} & \textit{ft+cl} & 60.23 $\pm$ 0.38 & 42.78 $\pm$ 0.40 & 62.49 $\pm$ 0.38 & 45.12 $\pm$ 0.39\\ \hhline{~|-;-|-|-|-|-|}
    
    &\multicolumn{1}{c}{\multirow{4}{*}{\shortstack{$k$-NN}}} & & 57.18 $\pm$ 0.40 & 38.44 $\pm$ 0.37 & 59.31 $\pm$ 0.41 & 40.53 $\pm$ 0.42\\
    
    & \multicolumn{1}{c}{} & \textit{ft} & 59.44 $\pm$ 0.39 & 43.49 $\pm$ 0.40 & 62.48 $\pm$ 0.39 & 45.68 $\pm$ 0.43\\
     
     & \multicolumn{1}{c}{} & \textit{cl} & 58.37 $\pm$ 0.39 & 43.20 $\pm$ 0.41 & 62.04 $\pm$ 0.39 & 45.36 $\pm$ 0.40\\
     
    \multirow{-12}{*}{\STAB{\rotatebox[origin=c]{90}{\textbf{Fine-Tuning}}}} & \multicolumn{1}{c}{} & \textit{ft+cl} & 59.59 $\pm$ 0.40 & 43.45 $\pm$ 0.40 & 62.45 $\pm$ 0.40 & 45.72 $\pm$ 0.40\\
    
    \hline
    
    &\multicolumn{1}{c}{\multirow{4}{*}{\shortstack{MAML~\cite{finn2017model}}}} & & 51.09 $\pm$ 0.43 & 37.20 $\pm$ 0.46 & 55.04 $\pm$ 0.42 & 39.06 $\pm$ 0.47 \\
    
    & \multicolumn{1}{c}{} & \textit{ft} & 54.95 $\pm$ 0.44 & 37.34 $\pm$ 0.47 & 59.57 $\pm$ 0.44 & 39.25 $\pm$ 0.46\\
     
    & \multicolumn{1}{c}{} & \textit{cl} & 53.62 $\pm$ 0.43 & 38.13 $\pm$ 0.47 & 56.80 $\pm$ 0.45 & 40.32 $\pm$ 0.48\\
     
    & \multicolumn{1}{c}{} & \textit{ft+cl} & 56.71 $\pm$ 0.46 & 40.36 $\pm$ 0.46 & 60.89 $\pm$ 0.45 & 42.16 $\pm$ 0.47\\ \hhline{~|-;-|-|-|-|-|}
    
    &\multicolumn{1}{c}{\multirow{4}{*}{\shortstack{LEO~\cite{rusu2018meta}}}} & & 56.52 $\pm$ 0.46 & 39.21 $\pm$ 0.53 & 56.66 $\pm$ 0.48 & 41.45 $\pm$ 0.54\\
    
    & \multicolumn{1}{c}{} & \textit{ft} & 56.77 $\pm$ 0.48 & 39.72 $\pm$ 0.54 & 62.95 $\pm$ 0.47 & 45.46 $\pm$ 0.55\\
     
    & \multicolumn{1}{c}{} & \textit{cl} & 56.73 $\pm$ 0.47 & 40.12 $\pm$ 0.55 & 56.90 $\pm$ 0.47 & 41.93 $\pm$ 0.56\\
     
    & \multicolumn{1}{c}{} & \textit{ft+cl} & 61.27 $\pm$ 0.46 & 42.79 $\pm$ 0.52 & 63.30 $\pm$ 0.47 & 43.81 $\pm$ 0.56\\ \hhline{~|-;-|-|-|-|-|}
    
    &\multicolumn{1}{c}{\multirow{4}{*}{\shortstack{MTL~\cite{sun2019meta}}}} & & 56.61 $\pm$ 0.42 & 41.56 $\pm$ 0.43 & 56.89 $\pm$ 0.41 & 43.15 $\pm$ 0.44\\
    
    & \multicolumn{1}{c}{} & \textit{ft} & 61.34 $\pm$ 0.41 & 42.90 $\pm$ 0.43 & 63.49 $\pm$ 0.40 & 45.28 $\pm$ 0.44\\
     
    & \multicolumn{1}{c}{} & \textit{cl} & 60.62 $\pm$ 0.41 & 42.87 $\pm$ 0.42 & 62.94 $\pm$ 0.40 & 45.57 $\pm$ 0.43\\
     
    & \multicolumn{1}{c}{} & \textit{ft+cl} & 62.39 $\pm$ 0.40 & 44.51 $\pm$ 0.43 & 65.00 $\pm$ 0.40 & 46.67 $\pm$ 0.43\\ \hhline{~|-;-|-|-|-|-|}
    
    &\multicolumn{1}{c}{\multirow{4}{*}{\shortstack{MN~\cite{vinyals2016matching}}}} & & 53.39 $\pm$ 0.46 & 40.34 $\pm$ 0.56 & 53.08 $\pm$ 0.45 & 42.04 $\pm$ 0.57\\
    
    & \multicolumn{1}{c}{} & \textit{ft} & 54.22 $\pm$ 0.46 & 40.62 $\pm$ 0.57 & 54.97 $\pm$ 0.47 & 42.52 $\pm$ 0.58\\
     
    & \multicolumn{1}{c}{} & \textit{cl} & 53.72 $\pm$ 0.47 & 40.42 $\pm$ 0.56 & 53.43 $\pm$ 0.45 & 42.19 $\pm$ 0.56\\
     
    & \multicolumn{1}{c}{} & \textit{ft+cl} & 56.03 $\pm$ 0.45 & 41.68 $\pm$ 0.54 & 58.69 $\pm$ 0.44 & 43.58 $\pm$ 0.56\\ \hhline{~|-;-|-|-|-|-|}
    
    &\multicolumn{1}{c}{\multirow{4}{*}{\shortstack{SIB~\cite{hu1empirical}\\(transductive)}}} & & 60.60 $\pm$ 0.46 & 45.87 $\pm$ 0.55 & 62.60 $\pm$ 0.49 & 49.16 $\pm$ 0.58\\
    
    & \multicolumn{1}{c}{} & \textit{ft} & 61.12 $\pm$ 0.45 & 46.64 $\pm$ 0.55 & 63.15 $\pm$ 0.47 & 49.78 $\pm$ 0.56\\
     
    & \multicolumn{1}{c}{} & \textit{cl} & 60.70 $\pm$ 0.46 & 46.14 $\pm$ 0.56 & 63.02 $\pm$ 0.48 & 49.43 $\pm$ 0.57\\
     
    & \multicolumn{1}{c}{} & \textit{ft+cl} & 62.07 $\pm$ 0.44 & 47.07 $\pm$ 0.53 & 64.07 $\pm$ 0.49 & 50.71 $\pm$ 0.54\\ \hhline{~|-;-|-|-|-|-|}
    
    &\multicolumn{1}{c}{\multirow{4}{*}{\shortstack{SIB~\cite{hu1empirical}\\(inductive)}}} & & 59.06 $\pm$ 0.42 & 41.48 $\pm$ 0.43 & 59.94 $\pm$ 0.42 & 43.27 $\pm$ 0.44\\
    
    & \multicolumn{1}{c}{} & \textit{ft} & 59.45 $\pm$ 0.41 & 41.98 $\pm$ 0.44 & 60.33 $\pm$ 0.44 & 43.61 $\pm$ 0.45\\
     
    & \multicolumn{1}{c}{} & \textit{cl} & 59.32 $\pm$ 0.42 & 41.67 $\pm$ 0.43 & 60.46 $\pm$ 0.43 & 43.52 $\pm$ 0.45\\
     
    \parbox[t]{2mm}{\multirow{-24}{*}{\rotatebox[origin=c]{90}{\textbf{Meta-Learning}}}} & \multicolumn{1}{c}{} & \textit{ft+cl} & 59.89 $\pm$ 0.41 & 43.20 $\pm$ 0.43 & 61.45 $\pm$ 0.43 & 44.27 $\pm$ 0.44\\
    \hhline{-|-|-|-|-|-|-|}
    \hhline{-|-|-|-|-|-|-|}
\end{tabular}
    \label{tab:a3}
\end{figure}

\end{document}